\documentclass[11pt]{article}

\usepackage[margin=1in]{geometry}  
\usepackage{graphics,graphicx,color}
\usepackage{amssymb,amsfonts,amsthm,amsmath,mathrsfs}
\usepackage{multirow, booktabs}
\usepackage{subcaption}
\usepackage{algorithm}
\usepackage{algpseudocode}
\usepackage[toc,page]{appendix}
\usepackage{authblk}
\usepackage{natbib}
\usepackage[colorlinks=true, linkcolor=blue, citecolor=blue]{hyperref}
\usepackage{url}
\usepackage{bm}
\usepackage{bbm}
\usepackage[super]{nth}
\usepackage{placeins}
\usepackage{xr}

\newcommand{\rowname}[1]
{\rotatebox{90}{\makebox[\tempdima][c]{\textbf{#1}}}}

\theoremstyle{remark}

\theoremstyle{plain}

\newcommand\dd{\mathrm{d}}
\newcommand{\x}{{\bm x}}

\newcommand{\y}{{\bm y}}

\newcommand{\ch}{\mathcal{H}}
\newcommand{\cx}{\mathcal{X}}
\newcommand{\ip}[3][{}]{\ensuremath{\left \langle #2, #3 \right \rangle_{#1}}}
\DeclareMathOperator*{\argmax}{arg\,max}
\DeclareMathOperator*{\argmin}{arg\,min}

\graphicspath{ {./Figures/} }
\title{A Deterministic Sampling Method via Maximum Mean Discrepancy Flow with Adaptive Kernel}

\author{Yindong Chen$^{1}$, Yiwei Wang$^{2}$, Chun Liu$^{1}$, Lulu Kang$^{3}$\footnote{Lulu Kang is the corresponding author.}}
\date{}

\begin{document}

\maketitle

\begin{center}
$^1$Department of Applied Mathematics, Illinois Institute of Technology, Chicago, IL, U.S.A.\\
Email: \url{ychen296@hawk.iit.edu} and \url{cliu124@illinoistech.edu}\\
$^2$Department of Mathematics, University of California, Riverside, CA, U.S.A.\\
Email: \url{yiweiw@ucr.edu}\\
$^3$Department of Mathematics and Statistics, University of Massachusetts, Amherst, MA, U.S.A.\\
Email: \url{lulukang@umass.edu}
\end{center}

\begin{abstract}
We propose a novel deterministic sampling method, EVI-MMD, to approximate a target distribution $\rho^*$ by minimizing the kernel discrepancy, also known as the Maximum Mean Discrepancy (MMD). 
Leveraging the \emph{energetic variational inference} framework \citep{wang2021particle}, we transform the MMD minimization problem into solving a dynamic system of Ordinary Differential Equations (ODEs) for particles. The implicit Euler scheme is employed to solve the ODE system, leading to a proximal minimization problem at each iteration, which is efficiently addressed using optimization algorithms such as L-BFGS. 
A key innovation of our method is a dynamic bandwidth selection strategy for the Gaussian kernel, which, although heuristic at this stage, represents a meaningful step toward addressing a long-standing challenge in kernel-based methods.
Comprehensive numerical experiments demonstrate that this adaptive bandwidth significantly enhances the performance of EVI-MMD. 
We apply the EVI-MMD algorithm to two types of sampling problems: (1) when the target distribution is fully specified by a density function, and (2) the ``two-sample problem,'' where only training data are available. 
In the latter case, EVI-MMD serves as a generative model, producing new samples that faithfully replicate the distribution of the training data. 
With carefully tuned parameters, EVI-MMD outperforms several existing methods in both scenarios. 

\end{abstract}

\newpage 

\section{Introduction}\label{sec:intro}

Many methods in statistics, machine learning, and applied mathematics require sampling from a certain target distribution.
For example, in numerical integration, the multidimensional integration $I=\mathbb{E}_{\bm x\sim \mu}[f(\bm X)]=\int_{\Omega} f(\bm x) \mu(\dd \bm x)$ is approximated by the sample mean $\hat{I}_n=\frac{1}{N}\sum_{i=1}^N f(\bm x_i)$, where the $f(\bm x)$ is the integral function, $\mu$ is the probability measure with the support region $\Omega$, and $\bm x_i$'s are the i.i.d. samples following the distribution of $\mu$.
Statistical design of experiments is also related to this area.
One such instance is the uniform space-filling design \citep{fang2000uniform}, in which the design points should approximate the uniform distribution.
Contrary to the two cases, where the target distribution is fully specified, in the ``two-sample problem'' the target distribution is completely unknown and only the training data are given. 
So the target distribution is the empirical distribution of the training data. 
Generative learning models can be used to generate new samples from the empirical distribution of training data in a parametric or nonparametric fashion.
They have gained a lot of attention and popularity due to the wide application of generative adversarial networks (or GANs) \citep{creswell2018generative,goodfellow2014generative} and variational autoencoders (or VAE) \citep{kingma2013auto}, which are built on parametric deep neural networks. 

In recent decades, variational inference (VI) has become an important and popular tool in machine learning, statistics, applied mathematics \citep{jordan1999introduction, WibisonoE7351,blei2017variational,mnih2016variational, gorbach2018scalable}, etc.
In short, the main goal of a VI method is to generate samples to approximate a target distribution.
Naturally, VI is strongly tied to these aforementioned research areas.
Its computational advantage has propelled the development of many VI-based supervised and unsupervised learning methods, such as Bayesian neural networks \citep{grave2011practical,welling2017multiplicative, wu2019deterministic,shridhar2019comprehensive}, Gaussian process model \citep{king2006fast, nguyen2013efficient, nguyen2014automated, shetha2015sparse, damianou2016variational, cheng2017variational}, and generative models \citep{kingma2014semi}.

In this paper, we propose a new variational inference approach by minimizing the kernel discrepancy via the energetic variational approach \citep{wang2021particle}.
Essentially, we generate samples, or \emph{particles}, to approximate various target distributions that are fully specified or empirically available from training data.

\subsection{Related Works}\label{sub:related}
The core idea of VI is to minimize a user-specified dissimilarity functional that measures the difference between two distributions.
Many dissimilarity functionals, such as Kullback-Leibler (KL-)divergence and the more general $f$-divergence \citep{csiszar2004information,zhang2019variational}, Wasserstein distance \citep{villani2021topics}, kernel stein discrepancy (KSD) \citep{liu2016kernelized, chen2018stein}, and kernel discrepancy, have been used in the literature.

If the target distribution is known up to the intractable normalizing constant, KL-divergence is commonly used \citep{liu2016stein, blei2017variational, ma2019sampling, heng2021gibbs}.
For example, the Langevin Monte Carlo (LMC) \citep{welling2011bayesian,cheng2018underdamped,bernton2018langevin} and the Stein Variational Gradient Descent (SVGD) \citep{liu2016stein} can be considered as a discretization of the Wasserstein gradient flow \citep{jordan1998variational} of the KL-divergence.
However, KL-divergence is only suitable for the target distribution whose density function takes the form $\frac{1}{Z} \exp(-V(\x))$.
Moreover, the KL-divergence-based algorithms require repeated evaluation of the gradient of the target distribution, which can be computationally costly if the target distribution is complicated to compute.

Kernel discrepancy is another popular dissimilarity functional.
In machine learning, kernel discrepancy is better known as \emph{Maximum Mean Discrepancy} or \emph{MMD}.
It is suitable for the case where the target distribution is compactly supported. 
Unlike KL-divergence, MMD does not have the ``$\log 0$'' issue which can occur when the density values are really small at certain particles.
What is more, it is easy to compute the MMD for the two-sample problems \citep{li2015generative,chwialkowski16kernel,li2017mmd}, in which the target distribution is the empirical distribution of training data. 
Recently, \cite{chen2026stationarymmdpoints} introduced stationary MMD points and applied it to numerical integration.
For these reasons, we choose kernel discrepancy or MMD as the objective functional.
We defer the detailed review of kernel discrepancy/MMD and its related literature in Section \ref{sec:background}.

Another important aspect of a VI approach is the minimization method.
As reviewed by \cite{blei2017variational}, the complexity of the minimization is largely decided by the distribution family $\mathcal{Q}$, i.e., the set of feasible distributions to approximate the target distribution.
It can be a family of parametric distributions.
But sometimes, the parametric distribution is too restrictive, and thus flow-based VI methods have been created, in which $\mathcal{Q}$ consists of distributions obtained by a series of smooth transformations from a tractable initial reference distribution.
Examples include normalizing flow VI methods \citep{rezende2015variational, kingma2016improved} and particle-based VI methods (ParVIs) \citep{liu2016stein, liu2017stein, liu2018riemannian, chen2018unified, liu2019understanding, chen2019projected,wang2021particle, arbel2019maximum, korba2021kernel, mroueh2019sobolev}.
Our proposed approach belongs to the ParVIs category.
Among all the ParVIs, SVGD \citep{liu2017stein} is one of the most popular early works. 
We compare the proposed approach with some other ParVI methods, including SVGD.

\subsection{Our Contributions}\label{sub:contribution}

In this paper, we propose a deterministic sampling method by minimizing the kernel discrepancy or MMD via the general energetic variational inference (EVI) framework \citep{wang2021particle}.
The EVI transforms the minimization problem into a dynamic system, which can be solved by many numerical schemes, including the implicit Euler method as demonstrated in this paper. 
We name it \emph{EVI-MMD} algorithm. 
The primary contributions and merits are summarized as follows:
\begin{itemize}
\item Numerical Stability via Implicit Schemes: Unlike most existing MMD gradient flows that utilize explicit Euler schemes (e.g., \cite{arbel2019maximum}), EVI-MMD employs an implicit Euler discretization. This transforms the particle update into a proximal minimization problem at each iteration, which is inherently more stable and allows for larger time steps ($\tau$) without the risk of particle collapse or divergence.
\item A Solution to the Bandwidth Challenge: A major contribution of this work is the introduction of a dynamic bandwidth selection strategy ($h_n = a/n^c + b$). Bandwidth selection has long been a ``bottleneck'' for kernel-based methods. Our strategy provides a meaningful step toward resolving this by balancing ``exploration'' (moving particles to high-density regions via large kernels) and ``exploitation'' (fine-tuning particle alignment via smaller kernels).
\item Versatility across Sampling Scenarios: While many MMD-based approaches \citep{gretton2012kernel,li2015generative,liu2016kernelized,li2017mmd,mak2018support,cheng2021neural,hofert2021quasi}  are limited to the ``two-sample'' problem, EVI-MMD is designed to be versatile. We demonstrate its effectiveness in two distinct scenarios: (1) when the target distribution is known via a density function (approximate inference), and (2) when only training data are available (generative modeling).
\item Gradient-Free Target Information: Unlike KL-divergence-based methods like SVGD or LMC, which require the gradient of the log-target density ($\nabla \log \rho^*$), EVI-MMD relies only on the density values (or samples). This makes it particularly useful for target distributions where the gradient is computationally expensive or intractable to compute.
\end{itemize}

The rest of the paper is organized as follows.
Section \ref{sec:background} gives the necessary background on the kernel discrepancy and the general deterministic sampling method by the EVI framework.
In Section \ref{sec:method}, we apply the general EVI framework to minimize the kernel discrepancy and discuss the practical issues of EVI-MMD, including the specification of adaptive bandwidth and other tuning parameters. 
In Section \ref{sec:num}, three groups of examples are used to compare the EVI-MMD algorithm with some alternative methods.
We conclude the paper in Section \ref{sec:con}.
The codes for all examples in both main body of the paper and supplement are available from the Github repository \url{https://github.com/lulukang/AdaptiveKernelMMD}.

\section{Background}\label{sec:background}
We first review the concept of kernel discrepancy, which is better known as MMD in the machine learning literature, and then explain the EVI framework.
The two combined are the foundation of the proposed EVI-MMD algorithm. 

\subsection{Kernel Discrepancy or MMD}\label{sub:mmd}

Before its wide recognition in machine learning as MMD, kernel discrepancy has been an important concept in QMC literature and was promoted as a goodness-of-fit statistic and a quality measure for statistical experimental design \citep{hickernell1998generalized,hickernell1999goodness,fang2000uniform,fang2000miscellanea,fang2002centered,hickernell2002uniform}.
Kernel discrepancy can be interpreted in different ways.
\cite{hickernell2016trio} and \cite{li2020transformed} explained \emph{three identities} of kernel discrepancy.
First, it can be considered as a norm on a Hilbert space of measures, which has to include the Dirac measure.
Second, it is commonly used as a deterministic cubature error bound for Monte Carlo methods.
Third, it is the root mean squared cubature error, where the kernel function is also the covariance function for a stochastic process.
Here we review it using the second identity and then generalize it and connect it with MMD.

Let $\Omega\subset \mathbb{R}^d$ be the domain of a probability measure $\mu$, which has density $\rho(\bm x)$ and cumulative distribution function $F(\bm x)$.
The three concepts, measure, density, and CDF, are used interchangeably in the rest of the paper to refer to distribution.
Let $(\ch, \ip[\ch]{\cdot}{\cdot})$ be a reproducing kernel Hilbert space (RKHS) of functions $f:\Omega \rightarrow \mathbb{R}$.
By definition, the reproducing kernel, $K$, is the unique function defined on $\Omega \times \Omega$ with the properties that $K(\cdot, \bm x)\in \ch$ for any $\bm x \in \Omega$ and $f(\bm x)=\ip[\ch]{K(\cdot,\bm x)}{f}$.
The second property implies that $K$ reproduces function values via the inner product.
It can be verified that $K$ is symmetric in its arguments and positive definite.

A cubature method approximates the integral $I=\int_{\Omega} f(\bm x)\rho(\bm x)\dd \bm x=\mathbb{E}_{\bm x \sim \mu}[f(\bm X)]$ of an $f\in \ch$ by the sample mean
\[\hat{I}_N=\frac{1}{N}\sum_{i=1}^N f(\bm x_i),\quad \text{where } \bm x_i\sim ^{iid} F(\bm x).\]
Let $\cx=\{\bm x_i\}_{i=1}^N$ be the set of the i.i.d. samples following $F(\bm x)$ distribution.
To measure the quality of the approximation, define the cubature error as
\[
\text{err}(f,\mathcal{X})=I-\hat{I}_N=\int_{\Omega} f(\bm x)\rho(\bm x)\dd \bm x-\frac{1}{N}\sum_{i=1}^N f(\bm x_i)=\int_{\Omega} f(\bm x)\dd [F(\bm x)-F_{\cx}(\bm x)],
\]
where $F_{\cx}$ is the empirical CDF based on the sample $\cx$.
Under modest assumptions of the reproducing kernel, based on Cauchy-Schwarz inequality, the tight error bound is
\[
|\text{err}(f, \cx)|\leq \|f\|_{\ch}D(\cx, F, K),
\]
where $\|f\|_{\ch}$ is the norm of the function $f$ based on the inner product of the RKHS $\ch$ and $D(\cx, F, K)$ is the kernel discrepancy whose square is equal to
\begin{align}\nonumber
D^2(\mathcal{X}, F, K)&=\int_{\Omega \times \Omega} K(\bm x,\bm y) \dd [F(\bm x)-F_{\cx}(\bm x)] \dd [F(\bm y)-F_{\cx}(\bm y)]\\\label{eq:mmdv1}
&=\int_{\Omega\times \Omega} K(\bm x,\bm y)\dd F(\bm x)\dd F(\bm y)-\frac{2}{N}\sum_{i=1}^N\int_{\Omega} K(\bm x_i, \bm y)\dd F(\bm y)+\frac{1}{N^2} \sum_{i,j=1}^NK(\bm x_i, \bm x_j).
\end{align}

Recall that the kernel discrepancy is also the norm on a Hilbert space of measures, i.e., the first identity mentioned earlier.
More specifically, this Hilbert space of measures, denoted by $\mathcal{M}$, is the closure of the pre-Hilbert space and its inner product is defined as
\[
\ip[\mathcal{M}]{\nu_1}{\nu_2}=\int_{\Omega\times \Omega} K(\bm x,\bm y)\nu_1(\dd \bm x) \nu_2(\dd \bm y).
\]
For the given kernel $K$, the Hilbert space contains all measures such that $\|\nu\|_{\mathcal{M}}$ is bounded.
Please see \cite{hickernell2016trio} or \cite{li2020transformed} for the detailed definitions of the RKHS $\ch$, $\mathcal{M}$, and the derivation of \eqref{eq:mmdv1}.
The kernel discrepancy can be more generally defined by
\begin{equation}\label{eq:mmdv2}
D^2(\nu_1, \nu_2, K)=\int_{\Omega\times \Omega}K(\bm x, \bm y)[\nu_1(\dd \bm x)-\nu_2(\dd \bm x)][\nu_1(\dd \bm y)-\nu_2(\dd \bm y)],
\end{equation}
measuring the difference between any $\nu_1, \nu_2\in \mathcal{M}$.
\cite{gretton2012kernel} defined the maximum mean discrepancy (MMD) as
\[
\text{MMD}(\ch, \nu_1,\nu_2)=\sup_{f\in \ch}(\mathbb{E}_{\bm x\sim \nu_1}[f(\bm x)]-\mathbb{E}_{\bm y\sim \nu_2}[f(\bm y)]),
\]
and under the same definition of $\ch$ and $\mathcal{M}$, the square of MMD is
\[
\text{MMD}^2(\ch, \nu_1,\nu_2)=\mathbb{E}_{\bm x,\bm x' \sim \nu_1}[K(\bm x, \bm x')]-2\mathbb{E}_{\bm x\sim \nu_1,\bm y\sim \nu_2}[K(\bm x,\bm y)]+\mathbb{E}_{\bm y\sim \nu_2, \bm y'\sim \nu_2}[K(\bm y, \bm y')],
\]
which is equivalent to $D^2(\nu_1,\nu_2,K)$ in \eqref{eq:mmdv2}.
Therefore, in the rest of the paper, we use kernel discrepancy and MMD interchangeably.

Kernel discrepancy has many desirable properties, one of which is measuring the difference between distributions.
In fact, $\text{MMD}(\ch, \nu_1,\nu_2)=0$ if and only if $\nu_1=\nu_2$, provided that $\Omega$ is a compact metric space and more importantly, $K$ is a universal kernel and thus $\ch$ is a universal RKHS \citep{gretton2012kernel}.
Simply put, universal kernel \citep{micchelli2006universal} means that $K$ has to be complex enough such that $\ch$ and $\mathcal{M}$ are sufficiently big.
Lower-order polynomial kernels, such as linear and second-order polynomials are not universal.
MMD induced by the second-order polynomial kernel can distinguish two distributions in terms of mean and variance, and the linear kernel can only do so in terms of the mean.
On the other hand, the Gaussian kernel is universal and thus the MMD based on it can be used as a metric for measures \citep{micchelli2006universal,fukumizu2007kernel}.
Therefore, with a proper kernel, if $D^2(\cx_N,F,K)\rightarrow 0$ as $N\rightarrow \infty$, then $F_{\cx_N}\rightarrow F$.
For fixed $N$, if $D^2(\cx,F,K)\rightarrow 0$ as $n \rightarrow \infty$ ($n$ is the notation for iteration of algorithm), then $F_{\cx}\rightarrow F$.
Kernel discrepancy is also related to energy distance \citep{szekely2013energy} and support points \citep{mak2018support}.
If set $K(\bm x,\bm y)=-\|\bm x-\bm y\|_2^2$, then the kernel discrepancy becomes energy distance. 
For the two-sample problem, the energy distance is given by
\begin{equation}\label{eq:energy}
E(F_n, F)=\frac{2}{N\cdot M}\sum_{i=1}^N \sum_{l=1}^M \|\bm x_i-\bm y_l\|_2-\frac{1}{N^2}\sum_{i,j}^N \|\bm x_i-\bm x_j\|_2-\frac{1}{M^2}\sum_{l,k=1}^N \|\bm y_l-\bm y_k\|_2.
\end{equation}

\subsection{Deterministic Sampling through EVI}\label{sub:evi}

Motivated by the energetic variational approaches for modeling the
dynamics of non-equilibrium thermodynamical systems \citep{Giga2017}, the energetic variational inference (EVI) framework uses a continuous energy-dissipation law to specify the dynamics of minimizing the objective function in machine learning problems.
Under the EVI framework, a practical algorithm can be obtained by introducing a suitable discretization to the continuous energy-dissipation law.
This idea was introduced and applied to variational inference by \cite{wang2021particle}.
It can also be applied to other machine learning problems similar to \cite{trillos2018bayesian} and \cite{weinan2020machine}.

We first introduce the EVI using the continuous formulation.
Let $\bm \phi_t$ be the dynamic flow map $\bm \phi_t: \mathbb{R}^d \rightarrow \mathbb{R}^d$ that continuously transforms the $d$-dimensional distribution from an initial distribution toward the target one and we require the map $\bm \phi_t$ to be smooth and one-to-one.
Let $\rho_{[\bm \phi_t]}$ denote the probability density that is transformed by ${\bm \phi}_t$ from an initial distribution. 
For a given target distribution $\rho^*$, one can define a functional $\mathcal{F}(\bm \phi_t) = D(\rho_{[\bm \phi]_t} || \rho^*)$, where $D$ is the user-specified dissimilarity functional, such as the KL-divergence in \cite{wang2021particle}.
Taking the analogy of a thermodynamics system, $\mathcal{F}(\bm \phi_t)$ can be viewed as the Helmholtz free energy.
Following the First and Second Law of thermodynamics \citep{Giga2017} (kinetic energy is set to be zero), one can impose the following energy-dissipation
\begin{equation}\label{eq:energydiss}
\frac{\dd}{\dd t} \mathcal{F}(\bm \phi_t) = - \triangle(\bm \phi_t, \dot{\bm \phi}_t),
\end{equation}
to describe a dynamics of minimizing $\mathcal{F}(\bm \phi_t)$. 
Here $\triangle(\bm \phi_t, \dot{\bm \phi}_t)$ is a user-specified functional representing the rate of energy dissipation, and $\dot{\bm \phi}_t$ is the derivative of $\bm \phi_t$ with time $t$.
So $\dot{\bm \phi}_t$ can be interpreted as the ``velocity'' of the transformation.
Each variational formulation gives a natural path of decreasing the objective functional $\mathcal{F}(\bm \phi_t)$ toward an equilibrium \citep{trillos2018bayesian}.

The dissipation functional should satisfy $\triangle(\bm \phi_t, \dot{\bm \phi}_t)\geq 0$ so that $\mathcal{F}(\bm \phi_t)$ decreases with time.
As discussed in \cite{wang2021particle}, there are many ways to specify $\triangle(\bm \phi_t, \dot{\bm \phi}_t)$ and the simplest among them is a quadratic functional in terms of $\dot{\bm \phi}_t$,
\begin{equation}\label{quad_D}
\triangle(\bm \phi_t,\dot{\bm \phi}_t)=\int_{\Omega_t} \eta(\rho_{[\bm \phi_t]}) \|\dot{\bm \phi}_t\|_2^2 \dd \bm x,
\end{equation}
where $\rho_{[\bm \phi_t]}$ denotes the pdf of the current distribution which is the initial distribution transformed by $\bm \phi_t$, $\eta(\cdot)$ is a user-specified positive function of $\rho_{[\bm \phi_t]}$, $\Omega_t$ is the current support, and $\| {\bm a} \|_2 = \sqrt{{\bm a}^\top {\bm a}}$ for $\forall {\bm a} \in \mathbb{R}^d$.

With the specified energy-dissipation law \eqref{eq:energydiss}, the energy variational approach derives the dynamics of the systems through two variational procedures, the Least Action Principle (LAP) and the Maximum Dissipation Principle (MDP), which leads to
\begin{equation*}
\frac{\delta \frac{1}{2}\triangle}{\delta \dot{\bm \phi}_t} = - \frac{\delta \mathcal{F}}{\delta \bm \phi_t}.
\end{equation*}
The approach is motivated by the seminal works of Raleigh \citep{strutt1871some} and Onsager \citep{onsager1931reciprocal,onsager1931reciprocal2}.
Using the quadratic $\triangle(\bm \phi_t, \dot{\bm \phi}_t)$ (\ref{quad_D}), the dynamics of decreasing $\mathcal{F}$ is
\begin{equation}\label{eq:EVIv1}
\eta(\rho_{[\bm \phi_t]}) \dot{\bm \phi}_t = - \frac{\delta \mathcal{F}}{\delta \bm \phi_t}.
\end{equation}

In general, this continuous formulation is difficult to solve, since the manifold of $\bm \phi_t$ is of infinite dimension.
Naturally, there are different approaches to approximating an infinite-dimensional manifold by a finite-dimensional manifold.
One such approach, as used in \cite{wang2021particle}, is to use particles (or samples) to approximate the continuous distribution $\rho_{[\bm \phi_t]}$ with kernel regularization.
If this approximation applies to \eqref{eq:EVIv1}, after the LAP and MDP variational steps, we call it the ``variation-then-approximation'' approach.
If this approximation is applied to \eqref{eq:energydiss} directly, before any variational steps, we call it the ``approximation-then-variation'' approach.
The latter leads to a discrete version of the energy-dissipation law, i.e.,
\begin{equation}\label{eq:EVIv2}
\frac{\dd}{\dd t} \mathcal{F}_h(\{\bm x_i(t)\}_{i=1}^N)=-\triangle_h(\{\bm x_i(t)\}_{i=1}^N, \{\dot{\bm x}_i(t)\}_{i=1}^N).
\end{equation}
Here $\{\bm x_i(t)\}_{i=1}^N$ are the locations of $N$ particles at time $t$ and $\dot{\bm x}_i(t)$ is the derivative of $\bm x_i$ with $t$, and thus is the velocity of the $i$th particle as it moves toward the target distribution.
The functional $\mathcal{F}_{h}$ and $\triangle_h$ are the discretized free energy and dissipation by the $N$ particles. 
The subscript $h$ of $\mathcal{F}_h$ and $\triangle$ denotes the bandwidth parameter of the kernel function used in the kernelization operation.
Applying the variational steps to \eqref{eq:EVIv2}, we obtain the dynamics of decreasing $\mathcal{F}$ at the particle level,
\begin{equation}\label{eq:EVIv3}
\frac{\delta \frac{1}{2}\triangle_h}{\delta \dot{\bm x}_i(t)} = - \frac{\delta \mathcal{F}_h}{\delta \bm x_i}, \quad \text{for }i=1, \ldots, N.
\end{equation}
This leads to an ODE system of $\{\bm x_i(t)\}_{i=1}^N$ that can be solved by different numerical schemes.
The solution of the ODE system is the particles approximating the target distribution. 
Using the dissipation $\triangle(\bm \phi_t,\dot{\bm \phi}_t)=G\int_{\Omega_t}  \rho_{[\bm \phi_t]} \|\dot{\bm \phi}_t\|_2^2 \dd \bm x$, the discretized dissipation is $\triangle_h = -\frac{G}{N}\sum_{i=1}^N\|\dot{\x}_i(t)\|_2^2,$ where $G$ is a positive constant. 
Then \eqref{eq:EVIv3} becomes 
\begin{equation}\label{eq:EVIv4}
\frac{G}{N} \dot{{\bm x}_i} = - \frac{\delta \mathcal{F}}{\delta \x_i} (\{ \x_i \}_{i=1}^N), \quad \text{for }i=1, \ldots, N.
\end{equation}

The most straightforward way to solve \eqref{eq:EVIv3} is the explicit Euler method, 
\begin{equation*}
\text{Explicit Euler:}\qquad \frac{G}{N}\cdot \frac{\bm x_i^{(n+1)} - \bm x_i^{(n)}}{\tau} = - \frac{\delta \mathcal{F}}{\delta \x_i} ( \{ \x_i^{n} \}_{i=1}^N),
\end{equation*}
which is equivalent to minimizing $\mathcal{F}_h$ using the gradient descent method, which is used in \cite{arbel2019maximum}. 
Another approach is to adopt the implicit Euler scheme to solve the ODE system.
This is done by discretizing the left-hand side of \eqref{eq:EVIv3} in time $t$ and replacing the $\{\bm x_i\}_{i=1}^N$ by $\{\bm x_i^{(n+1)}\}_{i=1}^N$ for $i=1,\ldots, N$ in the right-hand side, i.e., 
\begin{equation}\label{eq:EVI-MMD-Implicit}
\text{Implicit Euler:}\qquad \frac{G}{N}\cdot \frac{\bm x_i^{(n+1)} - \bm x_i^{(n)}}{\tau} = - \frac{\delta \mathcal{F}}{\delta \x_i} ( \{ \x_i^{n+1} \}_{i=1}^N).
\end{equation}
It is easy to show that a solution of the nonlinear system \eqref{eq:EVI-MMD-Implicit} can be obtained by solving an optimization problem
\begin{equation*}
\{\bm x_i^{(n+1)}\}_{i=1}^N = \argmin_{\{\bm x_i\}_{i=1}^N} (J_n(\{\bm x_i\}_{i=1}^N)),
\end{equation*}
where
\begin{equation}\label{eq:J}
J_n(\{\bm x_i\}_{i=1}^N):= \frac{G}{2\tau N} \sum_{i=1}^N \|\bm x_i - \bm x_i^{(n)} \|_2^2 + \mathcal{F}_h(\{\bm x_i\}_{i=1}^N). 
\end{equation}
We can therefore define the general EVI Algorithm \ref{alg:EVI-MMD-Ordinary}. 
It shares some resemblance in structure with the proximal point algorithm \citep{rockafellar1976monotone}. 
Compared to the explicit Euler method, the implicit method is more stable even with a relatively large step size $\tau$.
Indeed, it can be shown that \citep{wang2021particle}
\begin{equation*}
\mathcal{F}_h (\{ \x_i^{n+1} \}) \leq 	J_n(\ {\bm x}_i^{n+1} \}_{i=1}^N) \leq J_n(\ {\bm x}_i^{n} \}_{i=1}^N) = \mathcal{F} (\{ \x_i^{n} \}).
\end{equation*}
So the set of particles always reduces $\mathcal{F}_h(\{ \x_i  \}_{i=1}^N)$ in each iteration. 
Many other novel approaches can be proposed by applying different numerical algorithms to solve \eqref{eq:EVIv3} and/or by transforming the original optimization problem into a differential equation system. 

\begin{algorithm}[htb]
\caption{The Implicit EVI Algorithm} \label{alg:EVI-MMD-Ordinary}
\begin{algorithmic}[1]
\Require 
\begin{itemize}
\item [] $\rho_0$: the distribution of the initial particles
\item [] $N$: total number of particles
\item [] \texttt{maxIter}: the total number of iterations
\item [] $\tau$: step size of implicit Euler
\item [] $G$: user-specified positive constant for proper scaling
\item [] $h$: bandwidth parameter of the kernel function
\end{itemize}
\State Generate initial particles $\{\bm x_i^0\}_{i=1}^N$ from a distribution $\rho_0$.
\For{$n=0:\texttt{maxIter}$}
\State $\{\bm x_i^{(n+1)}\}_{i=1}^N= \argmax_{\{\bm x_i\}^N} \frac{G}{2\tau N}\sum_{i=1}^{N}\|\bm x_i^{(n)} - \bm x_i \|^2 + \mathcal{F}_h(  \{ \x_i \}_{i=1}^N )$
\EndFor
\end{algorithmic}
\end{algorithm}

\section{Practical EVI-MMD Algorithm}\label{sec:method}

Given the target probability measure $\mu$ whose CDF is $F$, and the proper reproducing kernel $K$, we choose the squared kernel discrepancy $D^2(\cx_N, F, K)$ as the discrete free energy $\mathcal{F}_h$, i.e.,
\begin{equation}\label{eq:evi-mmd1}
\mathcal{F}_h(\{\x_i(t)\}_{i=1}^N)=D^2(\{\x_i(t)\}_{i=1}^N, F, K), 
\end{equation}
which measures the difference between the empirical distribution of the particles and the target distribution.
We call Algorithm \ref{alg:EVI-MMD-Ordinary} with \eqref{eq:evi-mmd1} as the \emph{EVI-MMD} algorithm. 
To make sure the EVI-MMD performs well in general, there are still some challenges. 
In this section, we address each challenge and propose a practical EVI-MMD algorithm.

\subsection{Estimation of the Free Energy}\label{section:evaluation of energy}

The first challenge is how to estimate the free energy $D^2(\{\x_i(t)\}_{i=1}^N, F, K)$ efficiently. 
There are three terms in $D^2(\{\x_i(t)\}_{i=1}^N, F, K)$ in \eqref{eq:mmdv1}. 
The first term $\int_{\Omega\times \Omega} K(\bm x,\bm y)\dd F(\bm x)\dd F(\bm y)$ only depends on the target distribution and does not affect the minimization with respect to the particles. 
So we do not need to compute it in the optimization procedure. 
The third term $\frac{1}{N^2} \sum_{i,j=1}^NK(\bm x_i, \bm x_j)$ is easily calculated based on the particles $\{\bm x_i(t)\}_{i=1}^N$, which we call ``square-term''. 
Assume $\rho^*(\bm y)$ is the probability density function associated with the target CDF $F(\bm x)$. 
The main challenge is how to compute the ``cross-term'' in \eqref{eq:mmdv1}, i.e.,
\begin{equation}
\sum_{i=1}^N \int_{\Omega} K(\x_i, \y) \dd F(\y) =  \sum_{i=1}^N \int_{\Omega} K(\x_i, \y) \rho^* (\y) \dd \y.
\end{equation}
Since it is difficult to sample from $\rho^*(\bm y)$ directly, one cannot use any standard Monte Carlo integration. 
Fortunately, using the Gaussian kernel $K(\bm x_i, \bm x_j) = \exp\left(-\frac{\|\bm x_i - \bm x_j \|_2^2}{2h^2}\right)$, one can estimate this integration by generating samples from a Gaussian distribution. 
Indeed, for the Gaussian kernel, the cross-term can be estimated by
\begin{equation}\label{Cross_est}
\sum_{i=1}^N \int_{\Omega} \exp\left(-\frac{\|\bm y - \bm x_i \|_2^2}{2h^2} \right) \rho^* (\y) \dd \y
= \sum_{i=1}^N  C_h  \mathbb{E}_{\y \sim \mathcal{N}(\x_i,  h^2\bm I_d)} [ \rho^*(\y) ]
\approx \sum_{i=1}^N \frac{C_h}{L} \sum_{l=1}^L \rho^*(\x_i + h\bm \xi_l), 
\end{equation}
where $\{\bm \xi_l\}_{l=1}^L$ are sampled from the $d$-dimensional standard normal $\mathcal{N}({\bm 0}, \bm I_d)$ and $C_h = (2 \pi)^{d/2} h^d$ is the normalizing constant. 
The gradient of the cross-term with respect to $\x_i$ is also easy to compute based on the approximation (\ref{Cross_est}), i.e.,
\begin{equation}\label{eq:cross1}
\nabla_{\x_i} \left( \sum_{i=1}^N \int_{\Omega} \exp\left(-\frac{\|\bm y - \bm x_i \|_2^2}{2h^2} \right) \rho^* (\y) \dd \y \right)
    \approx \sum_{i=1}^N \frac{C_h}{L} \sum_{l=1}^L \nabla_{\x_i} \rho^*(\x_i + h{\bm \xi}_l).
\end{equation}
Other than the Gaussian kernel, any translation-invariant kernel that corresponds to a probability distribution and can be efficiently sampled from, can be used here.
In this paper, we use the Gaussian kernel when the target distribution is fully specified. 
Theoretically, it is unclear if the error in estimating the cross-term will significantly affect the final performance of EVI-MMD. 
In the numerical examples in Section \ref{sec:num} we set $L =100$ or $500$ depending on the scale of the problem and the computational cost. 
It is shown to achieve a good numerical performance. 

If the target distribution is an empirical one and only available from the training data $\{\bm y_i\}_{i=1}^M$ with sample size $M$, i.e., the two-sample problem, it is easy to estimate the cross-term by
\begin{equation}\label{eq:cross2}
\sum_{i=1}^N \int_{\Omega} K(\x_i, \y) \dd F(\y)= \frac{1}{M}\sum_{i=1}^N \sum_{j=1}^M K(\x_i, \y_j). 
\end{equation}
Due to the fact that $F(\bm y)$ represents the empirical CDF of the training data, the numerical integration is exactly as in \eqref{eq:cross2} and no approximation is involved.  
To reduce computation for large data, we can use the mini-batch procedure, which means randomly drawing a subset of samples $\{\bm y_i\}_{i=1}^L$ from $\{\bm y_i\}_{i=1}^M$ to compute \eqref{eq:cross2} in each iteration. 
For the two-sample problem, we can use kernels other than the Gaussian kernel. 
In Section \ref{sec:num}, we also use the kernel $K(\bm x, \bm y)=-\|\bm x-\bm y\|_2$ which leads to the energy distance and compare it with the Gaussian kernel using the EVI-MMD algorithm.

\subsection{Choice of the Ratio $G/\tau$}

The parameter $\tau$ is interpreted as the time step size in the implicit Euler in \eqref{eq:EVI-MMD-Implicit}. 
The positive constant $G$ is to scale the dissipation law to suit the chosen free energy. 
It is easy to see from Algorithm \ref{alg:EVI-MMD-Ordinary} that only the ratio $\tau^* = \tau / G$ affects the implicit Euler procedure, so we consider how to choose $\tau^*$. 
Notice that
\begin{equation*}
\begin{aligned}
J_n(\{\bm x_i^{(n+1)}\}_{i=1}^N) &= \frac{1}{2\tau^* N} \sum_{i=1}^N \|\bm x_i^{(n+1)} - \bm x_i^{(n)} \|_2^2 + \mathcal{F}_h(\{\bm x_i^{(n+1)}\}_{i=1}^N) \\
& \leq J_n(\{\bm x_i^{(n)}\}_{i=1}^N) = \mathcal{F}_h(\{\bm x_i^{(n)}\}_{i=1}^N),
\end{aligned}
\end{equation*}
which indicates
\begin{equation}
\sum_{i=1}^N \|\bm x_i^{(n+1)} - \bm x_i^{(n)} \|_2^2 \leq 2 \tau^* N | \mathcal{F}_h(\{\bm x_i^{(n)}\}_{i=1}^N) - \mathcal{F}_h(\{\bm x_i^{(n+1)}\}_{i=1}^N) |.
\end{equation}

The above inequality shows that, in each iteration, the displacement of the particles, i.e., the left-hand side of the inequality, is bounded above by the change of $\mathcal{F}_h$. 
For Gaussian kernel, the scale of $|\mathcal{F}_h(\{\bm x_i^{(n)}\}_{i=1}^N) - \mathcal{F}_h(\{\bm x_i^{(n+1)}\}_{i=1}^N) |$ is almost independent with the dimension. 
As a result, the convergence of the algorithm can be extremely slow for high-dimensional problems because $\sum_{i=1}^N \|\bm x_i^{(n+1)} - \bm x_i^{(n)} \|_2^2$ can be small if $\tau^*$ is small. 
This observation motivates us to pick a relatively large $\tau^*$ to balance the scale of the first and second terms in $J_n(\{ \x_i^{n+1} \}_{i=1}^N)$. 
We have done extensive numerical experiments and they suggest taking $\tau^*\approx d$, which is the dimension of $\bm x$. 
More detailed discussions on tuning parameters are in Section \ref{sub:finalalg}.

\subsection{Adaptive Bandwidth Selection for Gaussian Kernel}\label{sub:bandwidth}

Kernel selection is an important component in any MMD-based algorithm. 
The key question is how to choose the bandwidth since it is much easier to select the proper kernel function based on the problem. 
Bandwidth parameter significantly affects the efficiency and robustness of the algorithm \citep{briol2019statistical}. 
Although some approaches have been proposed in the literature \citep{binkowski2018demystifying, briol2019statistical, li2017mmd}, a satisfactory and universal solution is still not achieved yet.

In this paper, we propose a new adaptive bandwidth selection for the Gaussian kernel $K(\x, y) = \exp\left(- \| x - y \|^2 / 2 h^2\right)$, which was also considered in \cite{JMLR:v26:24-1574}.
We have done extensive numerical studies and observed the following trend. 
If the bandwidth $h$ is too small, there will be outliers that converge too slowly toward the target distribution, whereas if $h$ is too large, all the particles will eventually collapse to the same location as the algorithm iterates. 

\begin{figure}
\centering
\begin{subfigure}{\linewidth}
\includegraphics[width= 0.47 \linewidth]{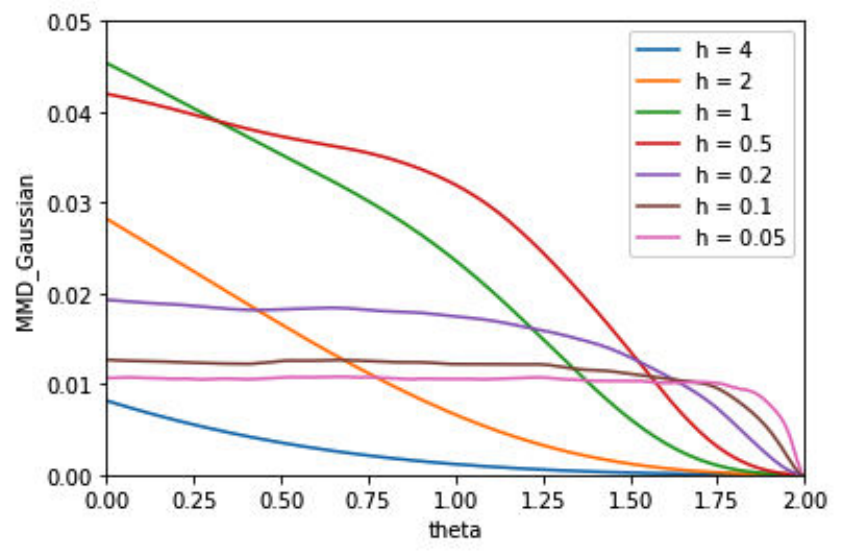}
\hfill
\includegraphics[width = 0.5 \linewidth]{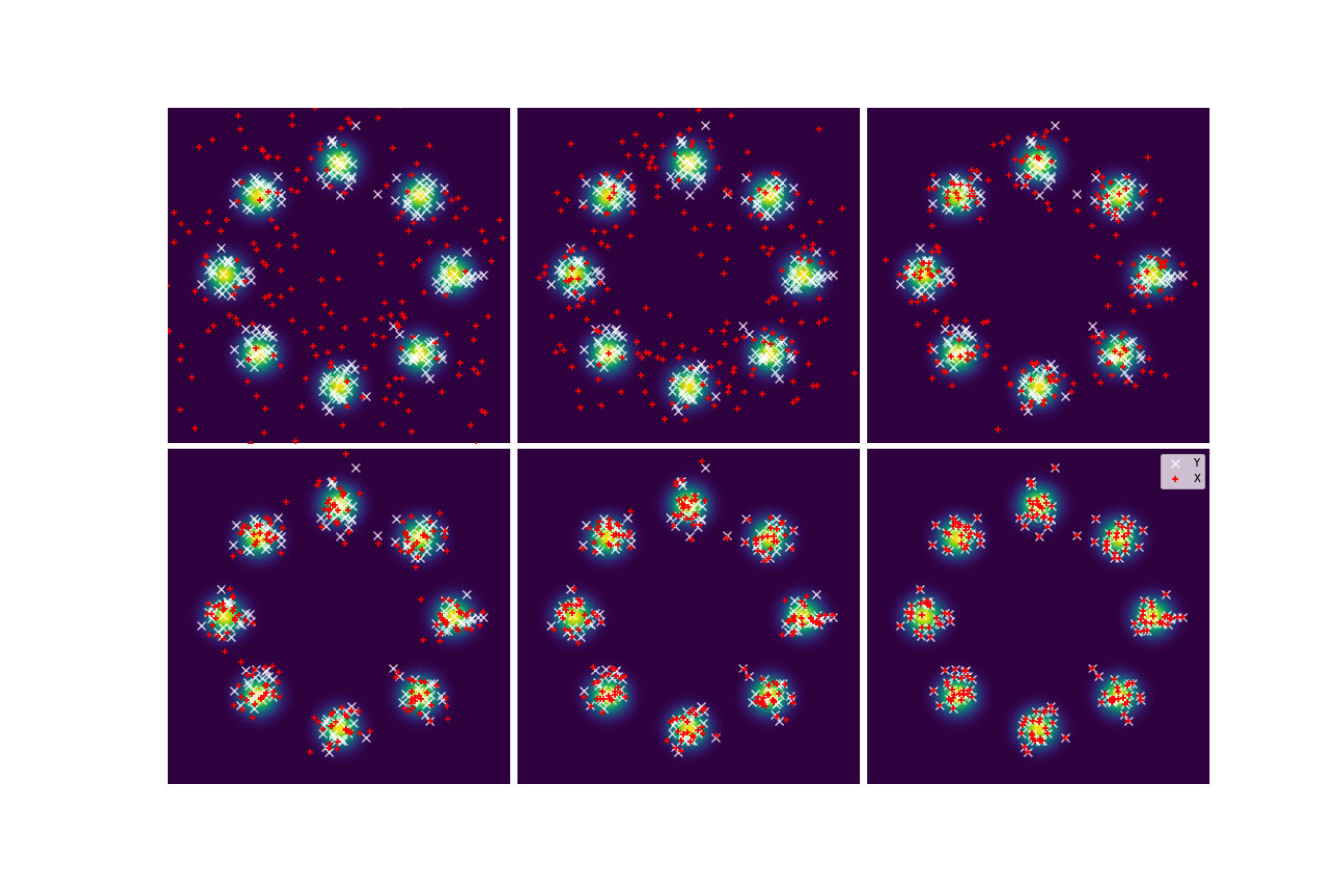}
\end{subfigure}
\caption{Left: decreasing MMD$^2$ curves with respect to $\theta\in [0,2]$ using different $h$. Right: from the first to the second row and from left to right, for $\theta=0, 1, 1.5, 1.75, 1.9, 2$, the red dots are samples $\{\bm x_{i}^{\theta}\}_{i=1}^N$ and are plotted on the contour plot of the target distribution.}\label{fig:mmd for different h}
\end{figure}

To illustrate this point, we construct a toy example to demonstrate different patterns of the decreasing ${\rm MMD}^2$ curves with different bandwidth settings. 
For a given $\theta \in [0, 2]$, we generate samples $\{\bm x_{i}^{\theta}\}_{i=1}^N$ from the model $\bm x_{i}^{\theta} = \bm y_i + (2-\theta)\bm z_i$, where $\bm z_i$ is sampled from a two-dimensional standard Gaussian distribution and $\bm y_i$ is sampled from the fully specified target distribution, an eight-component Gaussian mixture distribution (the second toy example in Section \ref{sub:toy}). 
It is obvious that as $\theta\rightarrow 2$, $\{\bm x_{i}^{\theta}\}_{i=1}^N$ converges to the target distribution. 
Indeed, in the right panel of Figure \ref{fig:mmd for different h}, the samples $\{\bm x_i^{\theta}\}^N$ (red dots) are closer to the Gaussian mixture distribution (green contour plot) as $\theta \rightarrow 2$. 
Using the Gaussian kernel, we compute ${\rm MMD}^2 ( \{\bm x_{i}^{\theta}\}_{i=1}^N, \{\bm y_i \}_{i=1}^N)$ for different $\theta$ and $h$ values. 
It is expected that all MMD$^2$ curves decrease to zero as $\theta\rightarrow 2$ as shown in the left panel of Figure \ref{fig:mmd for different h}. 
But they have very different decreasing patterns. 
For $h=0.5,1, 2, 4$, the ${\rm MMD}^2$ decreases very fast initially when $\theta$ is small but much slower when $\theta$ is close to $2$.
For $h=0.05, 0.1, 0.2$, the trend of ${\rm MMD}^2$ is completely opposite.
The curves are flat initially for $\theta$ in the range of $(0,1.75)$ and drop to zero very fast near the end.

In this toy example, samples $\{\bm x_{i}^{\theta}\}_{i=1}^N$ of different $\theta$ values mimic the particles that are evolving toward the target distribution in the EVI-MMD algorithm. 
Generalizing this observation to any MMD-based algorithm, the convergence of the particles is not at the optimal speed if $h$ is fixed throughout the algorithmic iteration. 
Therefore, we should choose a relatively large $h$ at the beginning of the iteration and gradually decrease $h$. 
In this way, we can take advantage of the fast decent of MMD$^2$ both at the early and end stages of the algorithm and avoid the plateau. 
The idea behind such an operation is ``exploration v.s. exploitation''. 
In the early stage, with a large $h$, the particles would explore a large neighborhood to find the region with a higher density of the target distribution. 
As the algorithm proceeds, when most particles are already at the region with a high probability density, the particles only need to exploit their close neighborhood and adjust their positions relative to other particles. 
As a result, the particles appear to align more regularly as shown in the toy examples in Section \ref{sub:toy}

Based on this idea, we propose to specify the bandwidth as follows,
\begin{equation}\label{eq:bandwidth}
h_n = \frac{a}{n^c} + b,
\end{equation}
where $n$ is the iteration index and $a$, $b$, and $c$ are three user-specified parameters. 
Among them, $a=h_{0}-b$ and we set it to be the median of the pairwise distance between the initial particles. 
In our examples, the initial distribution is a uniform distribution with a proper domain. 
The domain is obtained from the fully specified target distribution or from the training data in two-sample problems. 
The parameter $b>0$ is a regulatory parameter, which serves as a lower bound of $h_n$. 
It is not very influential to the algorithm since it is only to stop $h_n$ from becoming zero. 
For lower-dimensional problems, such as the examples in Section \ref{sub:toy} and \ref{sub:highdim}, we set $b=0.01$. 
For high-dimensional problems, such as the image data in Section \ref{sub:gen}, we set $b=1$. 
In fact, in all our examples, when $n=\texttt{maxIter}$ (in Algorithm \ref{alg:EVI-MMD}), $h_n$ is still significantly larger than $b$, and the largest \texttt{maxIter} we have used is 5000 (in this case $a$ is also very large). 
The parameter $c>0$ decides the rate of decrease of the bandwidth with respect to $n$. 
More discussion about the tuning parameters is in Section \ref{sub:finalalg}

\subsection{EVI-MMD with Adaptive Bandwidth}\label{sub:finalalg}

We summarize the EIV-MMD method using the Gaussian kernel and the adaptive bandwidth selection in Algorithm \ref{alg:EVI-MMD}. 
For the problem with a fully specified target distribution, the free energy is 
\begin{align*}
\mathcal{F}_{h_n}^*(\{\bm x_i\}_{i=1}^N)&=\int_{\Omega\times \Omega} \exp\left(-\frac{\|\bm x-\bm y\|_2^2}{2h_n^2}\right)\rho^*(\bm x)\rho^*(\bm y)\dd x\dd y-\frac{2}{N}\frac{C_{h_n}}{L}\sum_{i=1}^N\sum_{l=1}^L\rho^*(\bm x_i+h_n\bm \xi_l)\\
&+\frac{1}{N^2} \sum_{i,j=1}^N\exp\left(-\frac{\|\bm x_i-\bm x_j\|_2^2}{2h_n^2}\right).
\end{align*}
For the two-sample problem (training data size is $M$), the free energy is 
\begin{align*}
\mathcal{F}_{h_n}^*(\{\bm x_i\}_{i=1}^N)&=\frac{1}{M^2}\sum_{l,k=1}^M\exp\left(-\frac{\|\bm y_l-\bm y_k\|_2^2}{2h_n^2}\right)-\frac{2}{N\cdot M}\sum_{i=1}^N\sum_{j=1}^M\exp\left(-\frac{\|\bm x_i-\bm y_j\|_2^2}{2h_n^2}\right)\\
&+\frac{1}{N^2} \sum_{i,j=1}^N\exp\left(-\frac{\|\bm x_i-\bm x_j\|_2^2}{2h_n^2}\right).
\end{align*}
The constant term (first term) in the free energy is not relevant to the optimization with respect to the particles and thus they are not computed in the optimization. 
For the large two-sample problems, the cross term and square term (the second and third term) are computed using the mini-batch procedure to save computation. 
There are many methods available to solve the proximal point minimization in each iteration. 
We have chosen the L-BFGS method \citep{liu1989limited} and it performs adequately. 

\begin{algorithm}[htb]
\caption{The EVI-MMD Algorithm with Adaptive Bandwidth}\label{alg:EVI-MMD}
\begin{algorithmic}[1]
\Require 
\begin{itemize}
\item [] $\rho_0$: initial distribution of the particles. By default, we use the uniform distribution with a proper domain.
\item [] $\tau^*$: we set $\tau^*=d$, the dimension of the problem
\item [] $N$: total number of particles
\item [] $L$: the size of samples to generate $\bm \xi_l\sim \mathcal{N}({\bf 0}, \bm I_d)$ for $l=1,\ldots,L$ or the size of the mini-batch
\item [] \texttt{maxIter}: the maximum number of iterations
\item [] $b$, $c$: parameters for vanishing bandwidth for the Gaussian kernel
\end{itemize}
\State Generate initial particles $\{\bm x_i^{(1)}\}_{i=1}^N$ from a distribution $\rho_0$. 
\State $a = \text{median}\{\|\bm x_i - \bm x_j \|, i,j=1,\ldots,N\}$.
\State Generate i.i.d. samples $\bm \xi_l\sim \mathcal{N}({\bf 0}, \bm I_d)$ for $l=1,\ldots, L$.
\For{$n=1:\texttt{maxIter}$}
\State $h_n = a/n^c + b$	
\State $\{\bm x_i^{(n+1)}\}_{i=1}^N= \argmax_{\{\bm x_i\}^N} \frac{1}{2\tau^* N}\sum_{i=1}^{N}\|\bm x_i^{(n)} - \bm x_i\|^2 + \mathcal{F}_{h_n}^*(\{\bm x_i\}_{i=1}^N)$
\EndFor
\end{algorithmic}
\end{algorithm}

In the Supplement, we use the eight-component Gaussian mixture as the target distribution and show the performance of Algorithm \ref{alg:EVI-MMD}. 
Different combination of $a$, $c$, and $\tau^*$ are used. 
From this example and many other simulation studies we have done, we recommend the following settings for all the tuning parameters. 
\begin{itemize}
\item $\tau^*\approx d$
\item $a$ is the median of all the pairwise distances between the initial particles. 
\item $b$ is 0.01 or 1 depending on the dimension of the problem. 
\item $c$ is mostly selected by trial-and-error and the usual candidates we have tried are $c=0.2, 0.5$. 
\item $L$ is set based on the dimension of the problem as well as the constraint of the computation cost. In general, a bigger $L$ value leads to more accurate estimations of the free energy and its derivative. 
\item \texttt{maxIter} should be large enough to ensure the convergence of MMD$^2$ and the particles. 
\item $N$ is largely based on consideration of the computational cost. 
\end{itemize}
Although we recommend these rules-of-thumb on the tuning parameters, users should still run multiple trials to select the best possible combination of the tuning parameters for their problems in practice.

\section{Numerical Examples}\label{sec:num}

In this section, we demonstrate the performance of the proposed EVI-MMD algorithm through three types of examples. 
They cover two scenarios in which the target distribution is fully specified and the two-sample problems. 
In the latter case, the EVI-MMD is an effective generative model. 

In all examples, we set $a$ as the median of the pairwise distance of the initial particles and $\tau^* = d$. 
We set $b = 0.01$ for the examples in Section \ref{sub:toy} and \ref{sub:highdim} and $b = 1$ in Section \ref{sub:gen}. 
For the parameter $c$, through trial-and-error, we set $c=0.5$ for the examples in Section \ref{sub:toy}, $c=0.1$ in Section  \ref{sub:highdim} and $c=0.2$ in Section \ref{sub:gen}.
All algorithms and examples are implemented in Pytorch 1.10.1 \citep{NEURIPS2019_9015}. 
The computation was performed on the Open Science Grid \citep{osg07,osg09}. 

\subsection{Toy Examples}\label{sub:toy}

We test the Algorithm \ref{alg:EVI-MMD} in three toy examples where the target distributions are listed below. 
\begin{enumerate}
\item Star-shaped five-component Gaussian mixture distribution:
\[
\rho(\x)=\frac{1}{5}\sum_{i=1}^5 N(\bm x |\bm \mu_i,\bm \Sigma_i),
\]
where for $i=1,\ldots, 5$,
\[
\bm \mu_i= \left[
\begin{array}{rr}
\cos\left(\frac{2\pi}{5}\right), & -\sin\left(\frac{2\pi}{5}\right)\\
\sin\left(\frac{2\pi}{5}\right),& \cos\left(\frac{2\pi}{5}\right)
\end{array}
\right]^{i-1}
\left[
\begin{array}{c} 1.5 \\ 0 \end{array}
\right], \quad 
\bm \Sigma_i=\left[
\begin{array}{rr}
\cos\left(\frac{2\pi}{5}\right), & -\sin\left(\frac{2\pi}{5}\right)\\
\sin\left(\frac{2\pi}{5}\right),& \cos\left(\frac{2\pi}{5}\right)
\end{array}
\right]^{i-1}
\left[ 
\begin{array}{cc}
1, & 0 \\
0, & 0.01
\end{array}
\right].
\]

\item Eight-component Gaussian mixture distribution:  
\[
\rho(\x)=\frac{1}{8}\sum_{i=1}^8 N(\bm x|\bm \mu_i,\bm \Sigma),
\]
where $\bm \mu_1=(0,4)$, $\bm \mu_2=(2.8,2.8)$, $\bm \mu_3=(4,0)$, $\bm \mu_4=(-2.8,2.8)$, $\bm \mu_5=(-4,0)$, $\bm \mu_6=(-2.8,-2.8)$, $\bm \mu_7=(0,-4)$, $\bm \mu_8=(2.8,-2.8)$, and $\bm \Sigma=\left[\begin{array}{rr} 0.2, & 0 \\ 0, & 0.2 \end{array}\right]$. 

\item Wave-shaped distribution: 
\[
\rho(\x)=9.93^{-1}\exp\left(-0.1x_1^2-(x_2-\sin(\pi x_1))^2\right).
\]
\end{enumerate}

Although the first two distributions are both Gaussian mixture distributions, the eight-component Gaussian mixture distribution is more challenging since the effective support region for each Gaussian component is not connected, unlike the star-shaped distribution. 

\begin{figure}[htb]
\centering
\begin{subfigure}{\linewidth}
\includegraphics[width=0.32\linewidth]{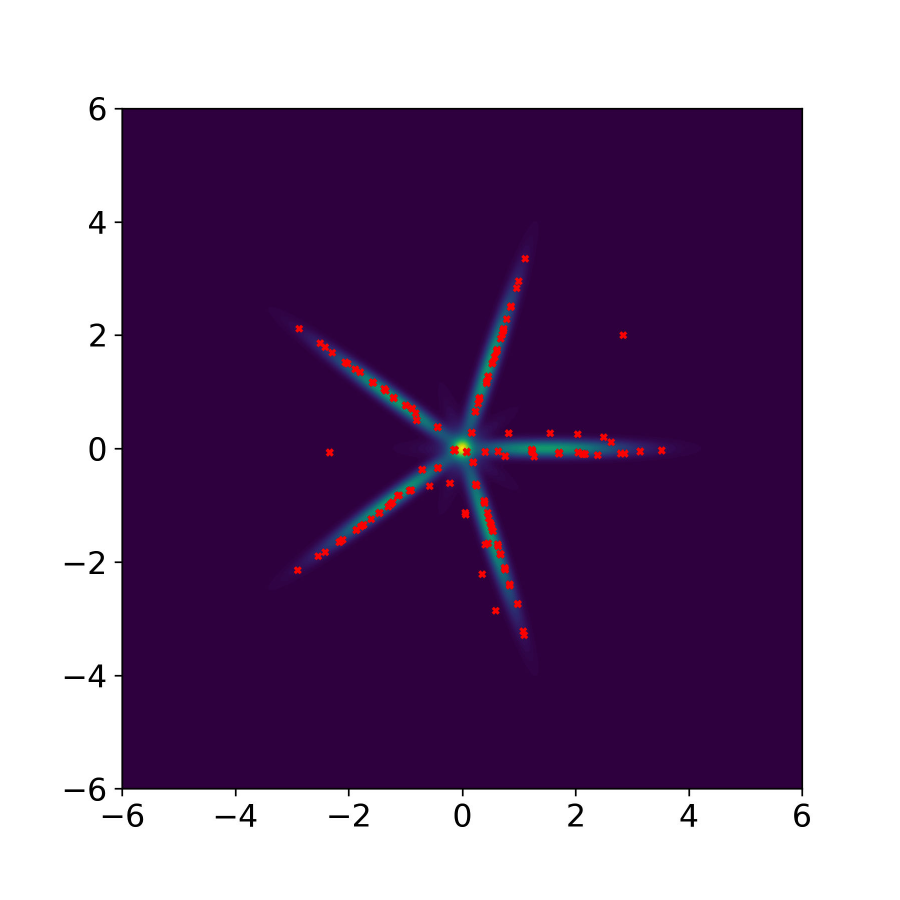}
\includegraphics[width=0.32\linewidth]{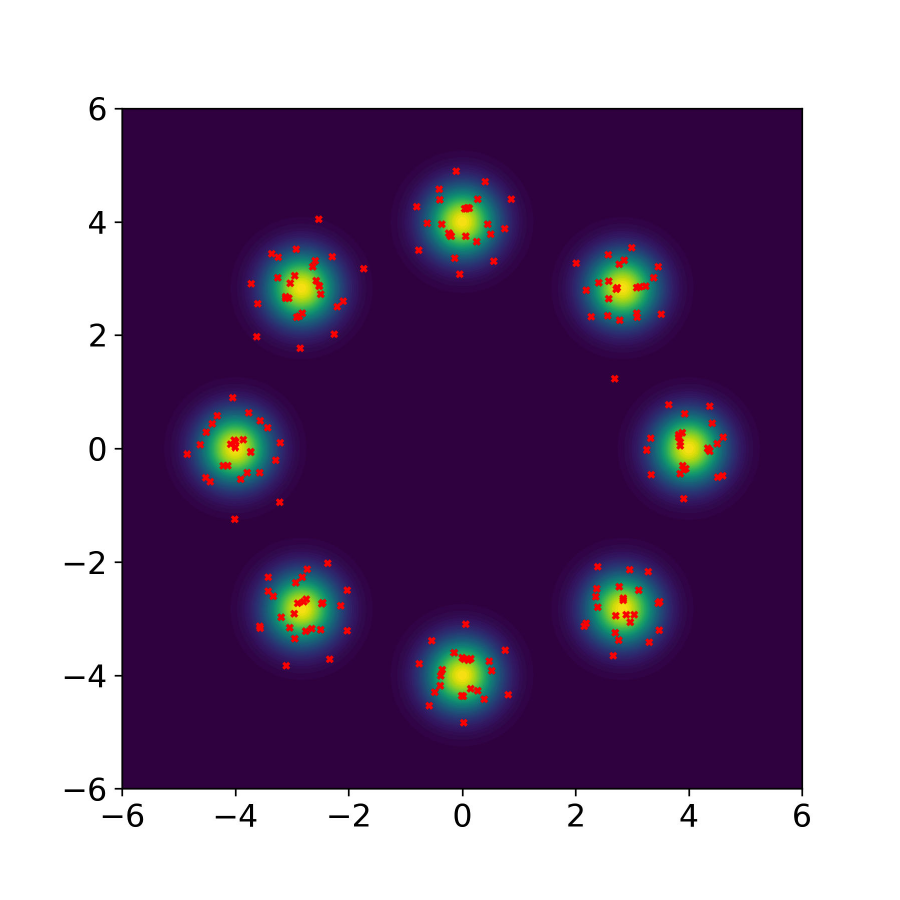}
\includegraphics[width=0.32\linewidth]{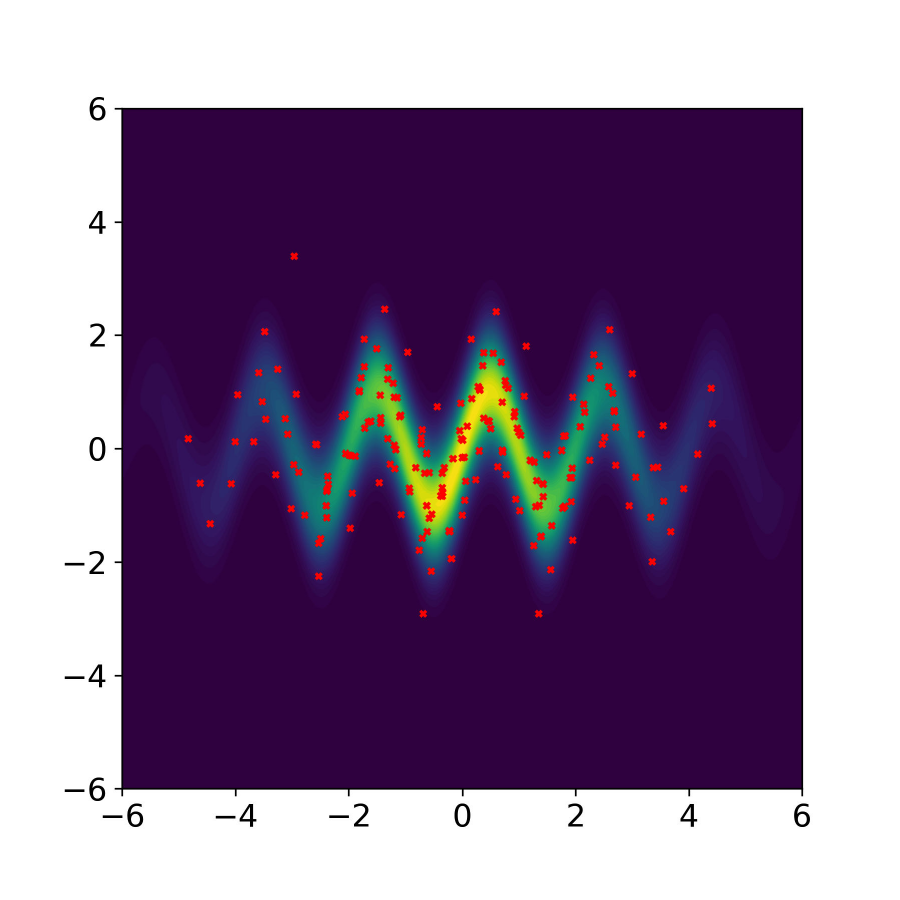}
\end{subfigure}
\begin{subfigure}{\linewidth}
\includegraphics[width=.32\linewidth]{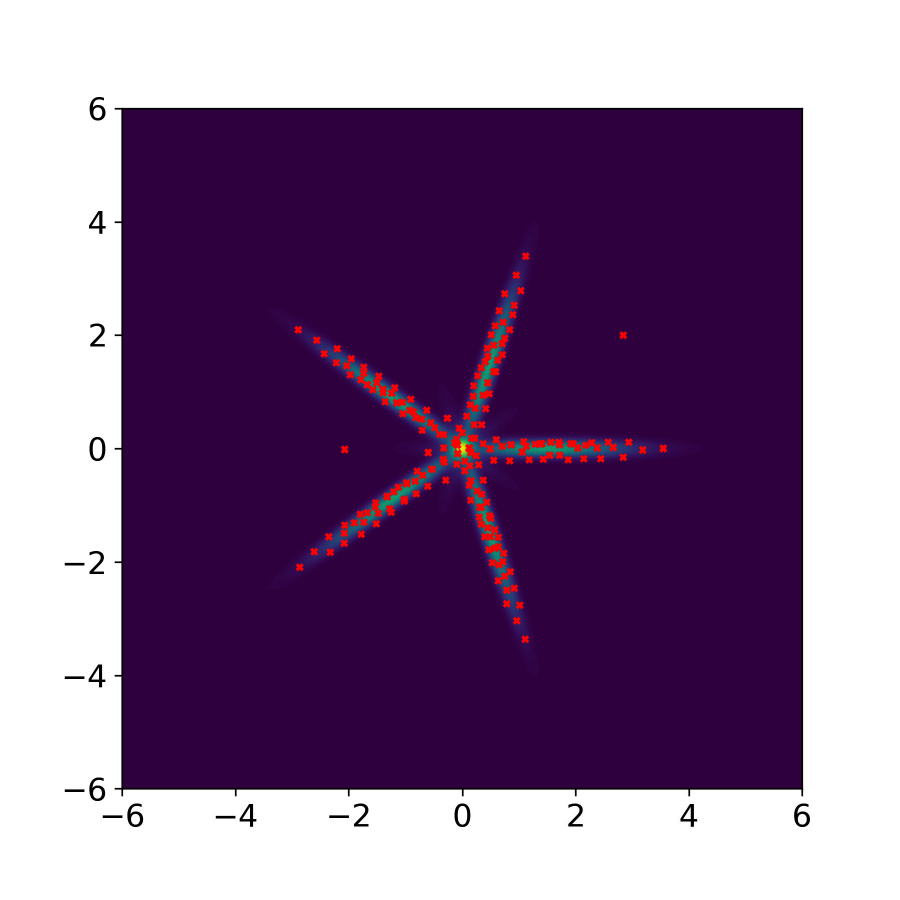}
\includegraphics[width=.32\linewidth]{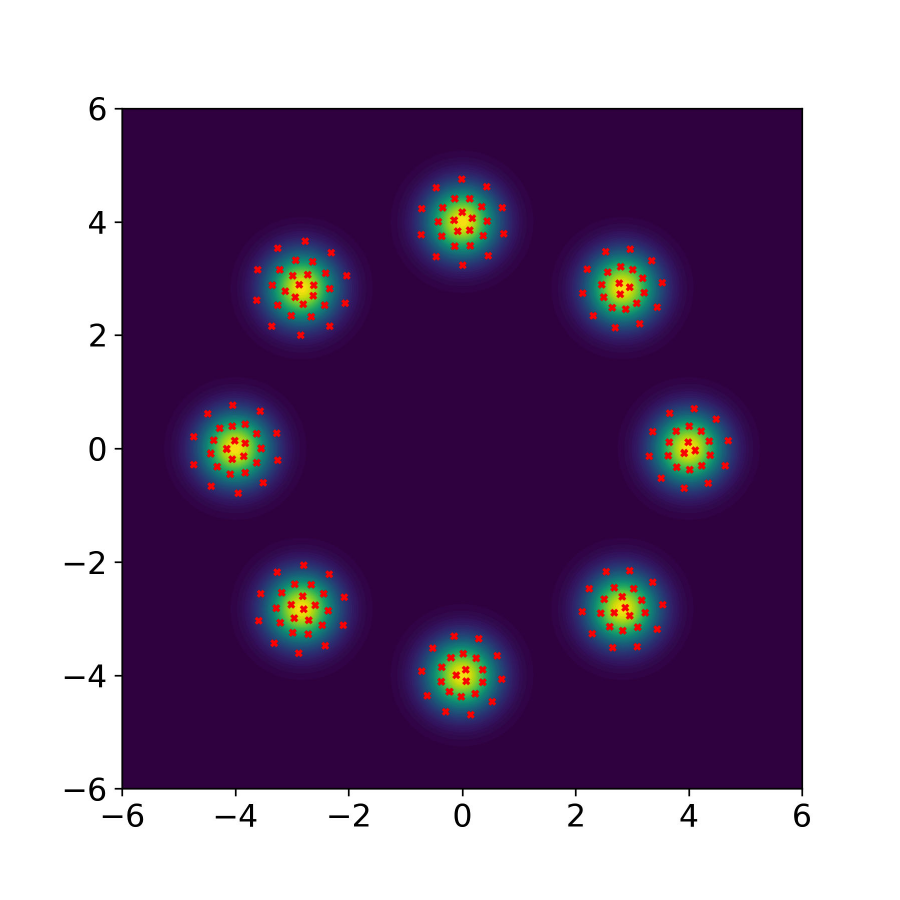}
\includegraphics[width=.32\linewidth]{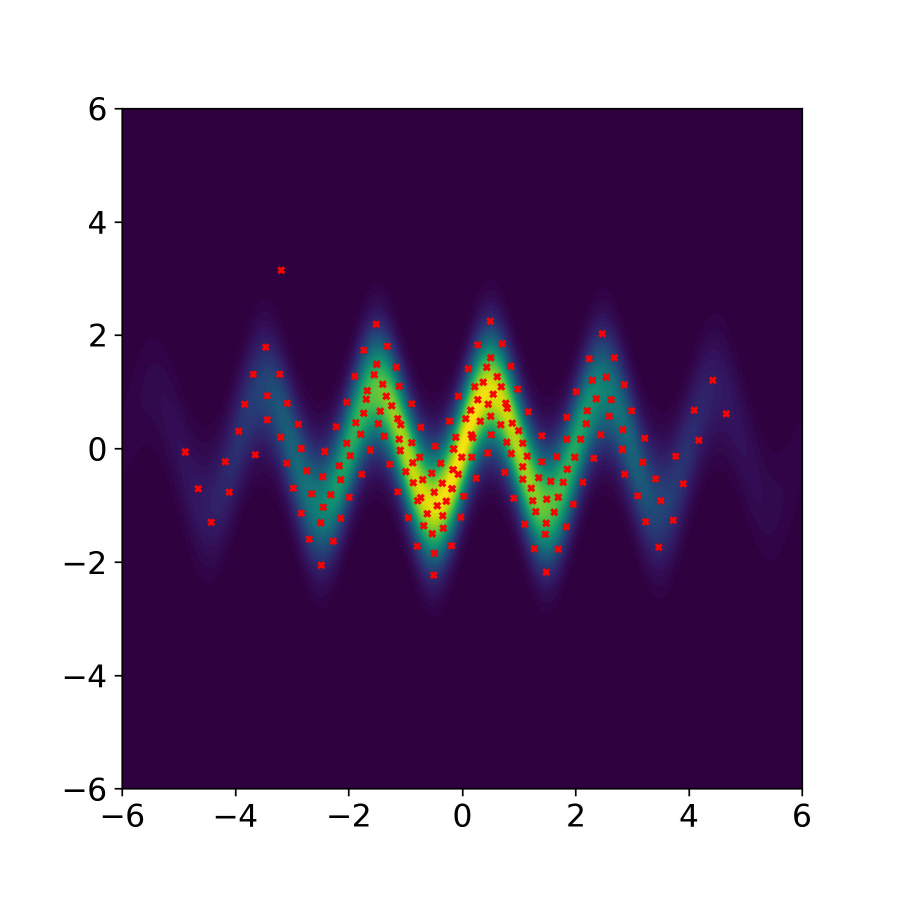}
\end{subfigure}
\begin{subfigure}{\linewidth}
\includegraphics[width=.32\linewidth]{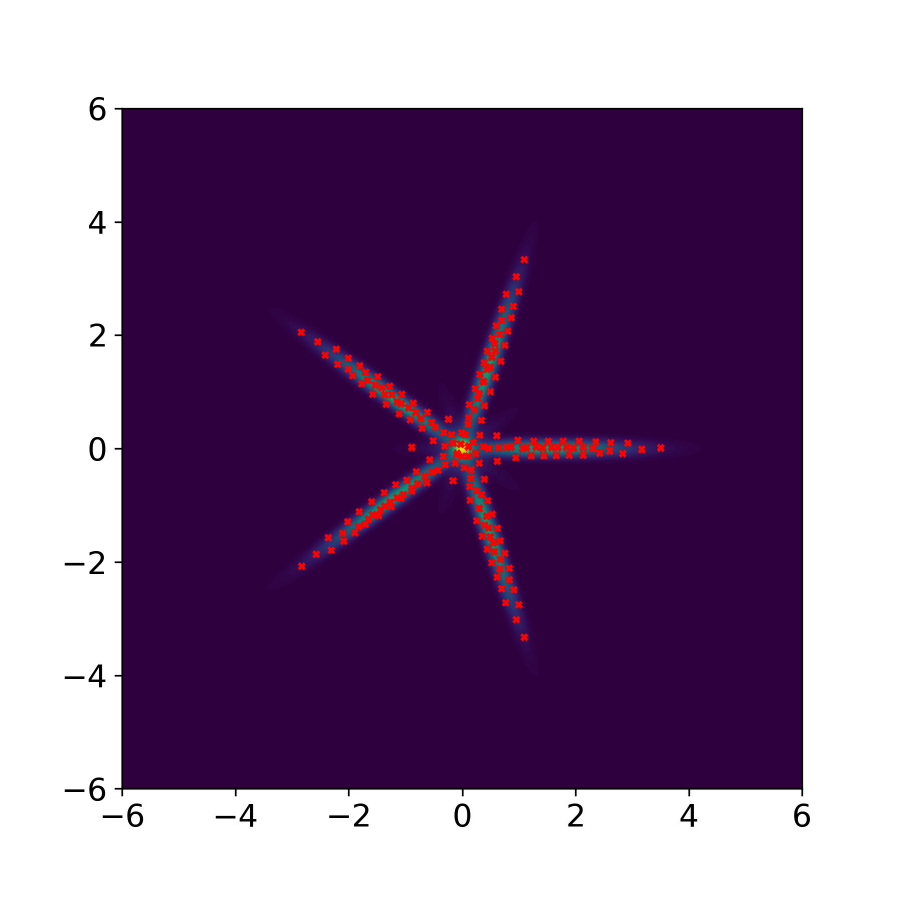}
\includegraphics[width=.32\linewidth]{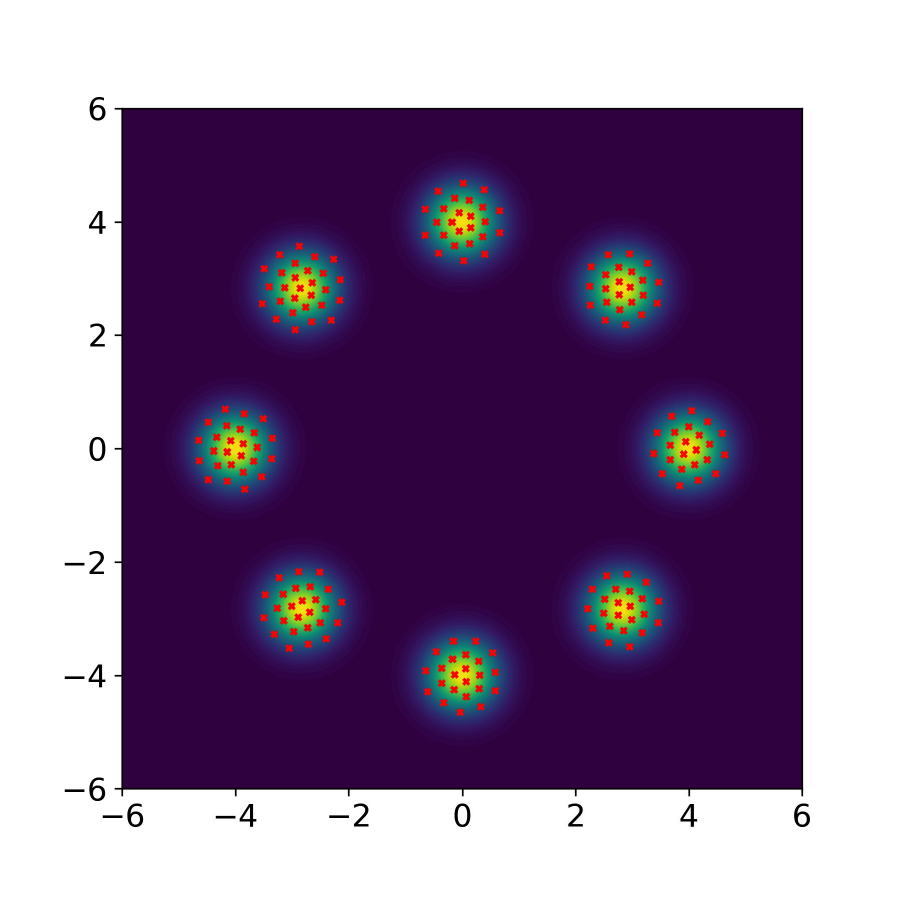}
\includegraphics[width=.32\linewidth]{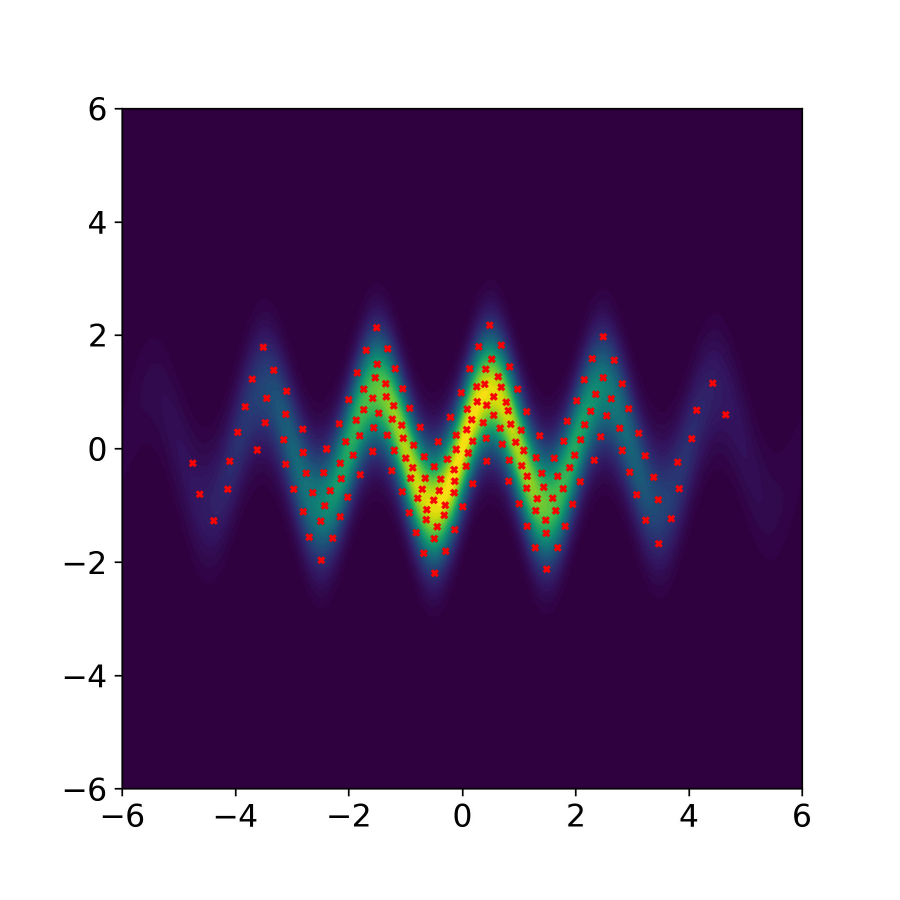}
\end{subfigure}
\caption{From top to bottom, each column of sub-figures show the particles by the EVI-MMD algorithm at $n=100$, $n=500$, and $n=1000$ iterations. The target density function is plotted as the contour in the background.}
\label{fig:toys}
\end{figure}

For all three examples, we set $N=200$ and \texttt{maxIter}=1000, with initial distributions of Uniform$[-2,2]$, Uniform$[-4,4]$, and Uniform$[-3,3]$, respectively. 
Figure \ref{fig:toys} displays the particles at the 100th, 500th, and 1000th iterations across three rows of sub-figures. 
By the 100th iteration, most particles have moved toward the high-density regions. 
At the 500th iteration, the particles are largely aligned, with only a few outliers remaining. 
By the 1000th iteration, even these outliers have converged to the target distribution. 
This behavior exemplifies the exploration-exploitation trade-off of the EVI-MMD method. 
In the early stages, a larger bandwidth encourages exploration, guiding particles toward high-density areas. 
In later stages, a smaller bandwidth facilitates exploitation, refining the alignment of particles to the target distribution.

We compare the proposed Algorithm \ref{alg:EVI-MMD} with other similar methods, including the EVI-Im by \cite{wang2021particle}, SVGD by \cite{liu2017stein}, and Langevin Monte Carlo (LMC) by \cite{rossky1978brownian}. 
For the EVI-Im and SVGD methods, we set the step size $\eta_0=0.1$ and a fixed bandwidth $h=0.1$ for the Gaussian kernel. 
The tuning parameters of LMC are $a=0.1$, $b=1$, and $c=0.55$, which decides the step size of LMC by the equation $\eta_0 = a (b+n)^{-c}$ (different from the proposed algorithm).

\begin{figure}[htb]
\centering
\begin{subfigure}{\linewidth}
\includegraphics[width=.33\linewidth]{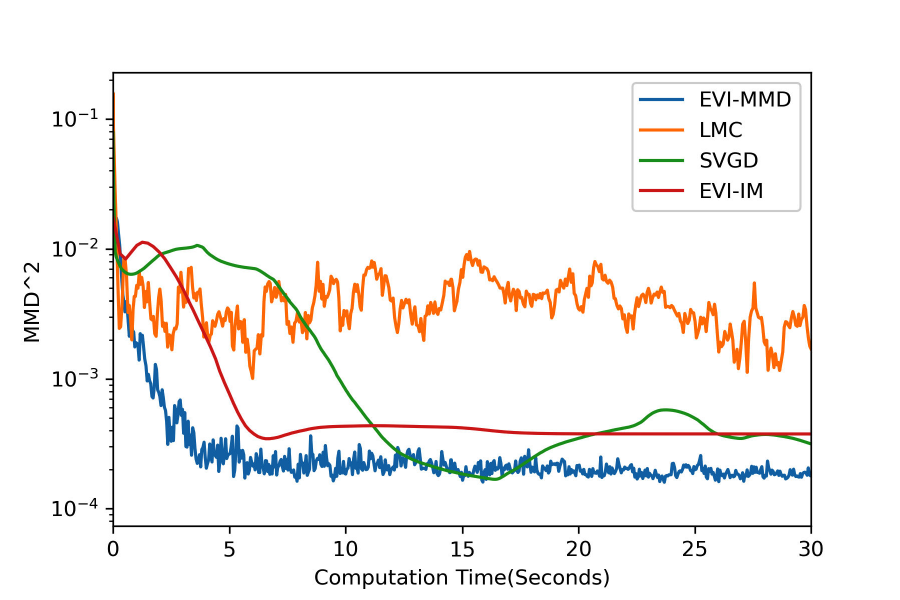}\hfill
\includegraphics[width=.33\linewidth]{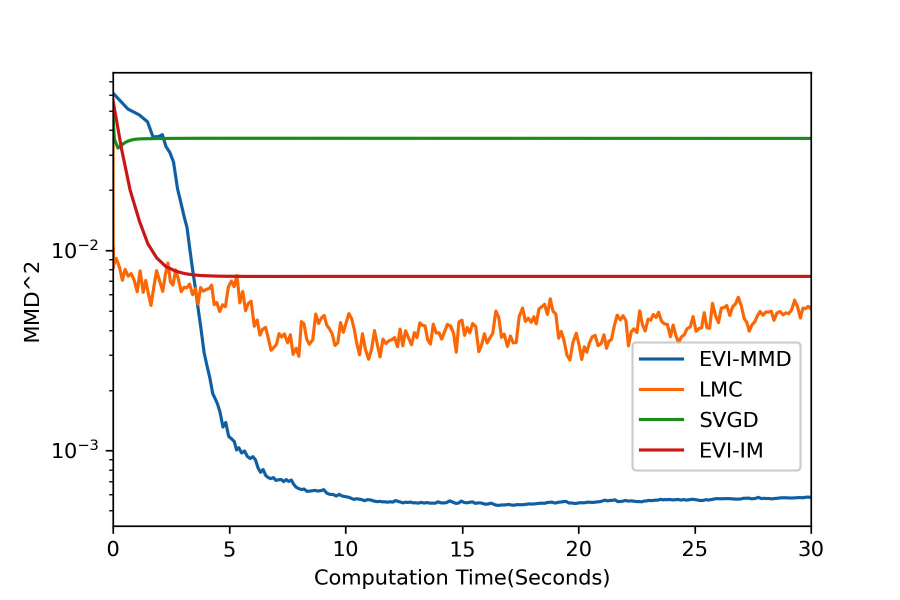} \hfill
\includegraphics[width=.33\linewidth]{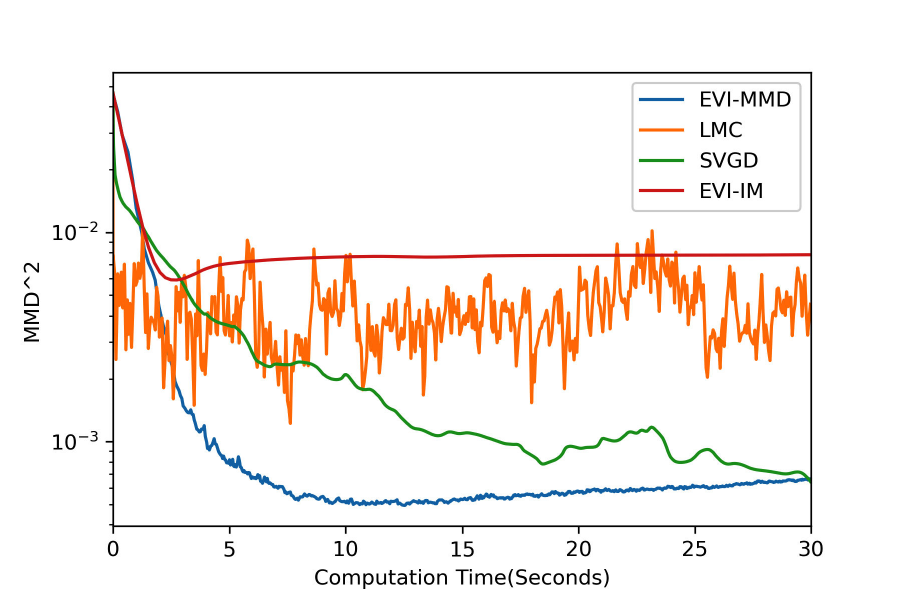}
\end{subfigure}
\caption{From left to right, the three sub-figures show the decreasing MMD$^2$ of four methods for the three toy examples with the target distributi:ons star-shaped five-component Gaussian mixture distribution, eight-component Gaussian mixture distribution, and wave-shaped distribution.\label{fig:toymmd}}
\end{figure}

To fairly compare the four algorithms, we compute the MMD$^2$ criterion using a fixed bandwidth $h=0.5$. 
So this MMD$^2$ is \emph{not} the objective function of any algorithms in this comparison. 
Figure \ref{fig:toymmd} shows the decreasing MMD$^2$ with respect to iterations of all the algorithms.
{}From Figure \ref{fig:toymmd} we can see that the EVI-MMD has the best performance for all three examples.

\subsection{Gaussian Distribution with Increasing Dimension}\label{sub:highdim}

In this example, we study the performance of the proposed algorithm with respect to the increasing dimension. 
We compare the EVI-MMD with the adaptive bandwidth for the Gaussian kernel with two alternative methods: 
\begin{itemize}
\item EVI-Energy-Distance: this method means that we set the free energy in Algorithm \ref{alg:EVI-MMD-Ordinary} to be the energy distance in \eqref{eq:energy} proposed in \cite{szekely2013energy}, i.e., $\mathcal{F}(\{\bm x_i\}_{i=1}^n)=E(F_n,F)$. 
\item Support-Points: this method is proposed by \cite{mak2018support}, which minimizes the same energy distance by a combination of the convex-concave procedure (CCP) with the resampling method. 
\end{itemize}
The EVI-Energy-Distance method and the Support-Points method both minimize the same objective function but use different minimization methods.
The EVI-MMD method and the EVI-Energy-Distance method minimize different objective functions but use the same implicit method. 

\begin{figure}[htb]
\centering
\begin{subfigure}{\linewidth}
\includegraphics[width = 0.5 \linewidth]{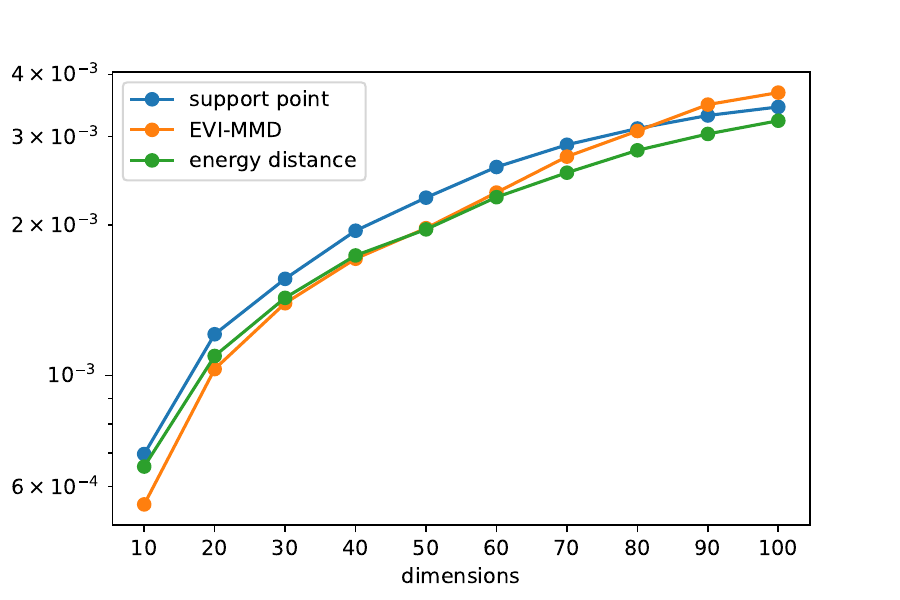}
\includegraphics[width = 0.5 \linewidth]{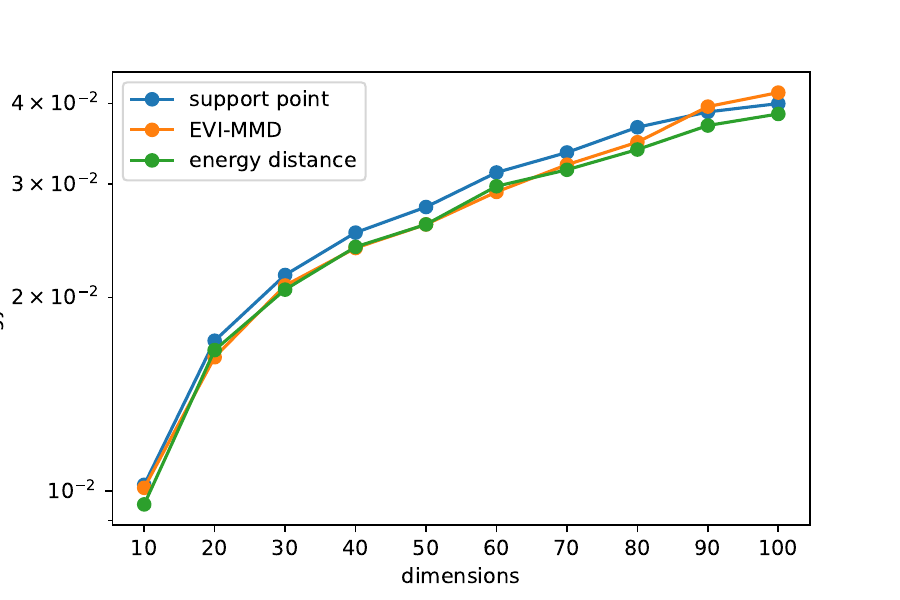}
\end{subfigure}
\caption{The MMD$^2$ criterion (Gaussian kernel with $h=5$) (left) and energy distance criterion (right) of the particles returned by three methods: blue curve for Support-Point, orange curve for EVI-MMD, and green curve for EVI-Energy Distance.}
\label{fig:highdimgaus}
\end{figure}

We sample training data of size $M=50,000$ from a standard Gaussian distribution with dimensions $d=10, 20,\ldots, 100$. 
The same training data is used to compare the three methods. 
The same training data is used to compare the three methods, along with identical particle sizes $N=500$ and initial samples generated from Uniform$[-2,2]^d$.
For the EVI-MMD method, the tuning parameters $(a, b, c)$ are set as described at the beginning of this section. Both EVI-MMD and EVI-Energy-Distance use $\texttt{maxIter} = 500$ for lower-dimensional examples ($d <= 60$) and $\texttt{maxIter} = 5000$ for higher-dimensional examples, with a mini-batch size of $L=5000$.
For Support-Points, we adopt the default settings recommended by \cite{mak2018support}. 
Figure \ref{fig:highdimgaus} compares the MMD$^2$ (using a Gaussian kernel with fixed bandwidth) and the energy distance $E(F_n,F)$ defined in \eqref{eq:energy} for the particles returned by the three methods after convergence, across dimensions $d=10,\ldots, 100$. 
Notably, the MMD with a Gaussian kernel can fail if the bandwidth is unsuitable. 
A reasonable choice is $h = \sqrt{d/2}$, and and we fix $h=5$ for our experiments. 

From the \ref{fig:highdimgaus}, we observe that the EVI-MMD method generally outperforms the Support-Points method, except in the $d=90$ and $d=100$ cases. 
Note that the $y$-axis is in log scale, so the differences between the methods are not substantial. 
This suggests that EVI-MMD delivers results comparable to those of Support-Points. 
However, in Section \ref{sub:gen}, the advantage of EVI-MMD becomes much more pronounced, as Support-Points fails to converge, highlighting the robustness and effectiveness of our proposed method.

\subsection{Generative Model}\label{sub:gen}

Generative learning models have been widely used in various machine learning applications. 
They can solve supervised and unsupervised learning problems. 
Simply put, the generative learning model generates new samples based on the training data. 
More advanced generative learning models are combined with deep neural networks \citep{jabbar2021survey}, Naive Bayes, Gaussian mixture model, hidden Markov model, etc. \citep{harshvardhan2020comprehensive}.
In this example, we use the simplest nonparametric generative learning setup and apply the EVI-MMD to three benchmark image datasets, MNIST \citep{deng2012mnist}, Fashion-MNIST \citep{xiao2017online}, and Cifar10 \citep{krizhevsky2009learning}.

For the MNIST and Fashion-MNIST datasets, each data point is a grey image of $d=28\times 28=784$ size of pixels. 
For the Cifar10 dataset, each data point is a full-color image of $d=32\times 32 \times 3=3072$ size of pixels.
One pixel value is in $[0,1]$. 
All three are extremely high-dimensional two-sample problems. 
For the MNIST and Fashion-MNIST datasets, we use the whole dataset of size $M=60,000$ as the training data and resample a mini-batch of $L=100$ samples in each iteration.
However, for the Cifar10 dataset, due to the extremely high dimension, we randomly choose a subset of $M=5000$ images as the training data and also use the mini-batch of the size of $L=100$. 
We choose $L=100$, which is relatively small considering the dimension, is mainly due to the limited computing resources, but it has been proved to be sufficient. 

\begin{figure}[!htb]
\centering
\begin{subfigure}{\linewidth}
\includegraphics[width=0.32\linewidth]{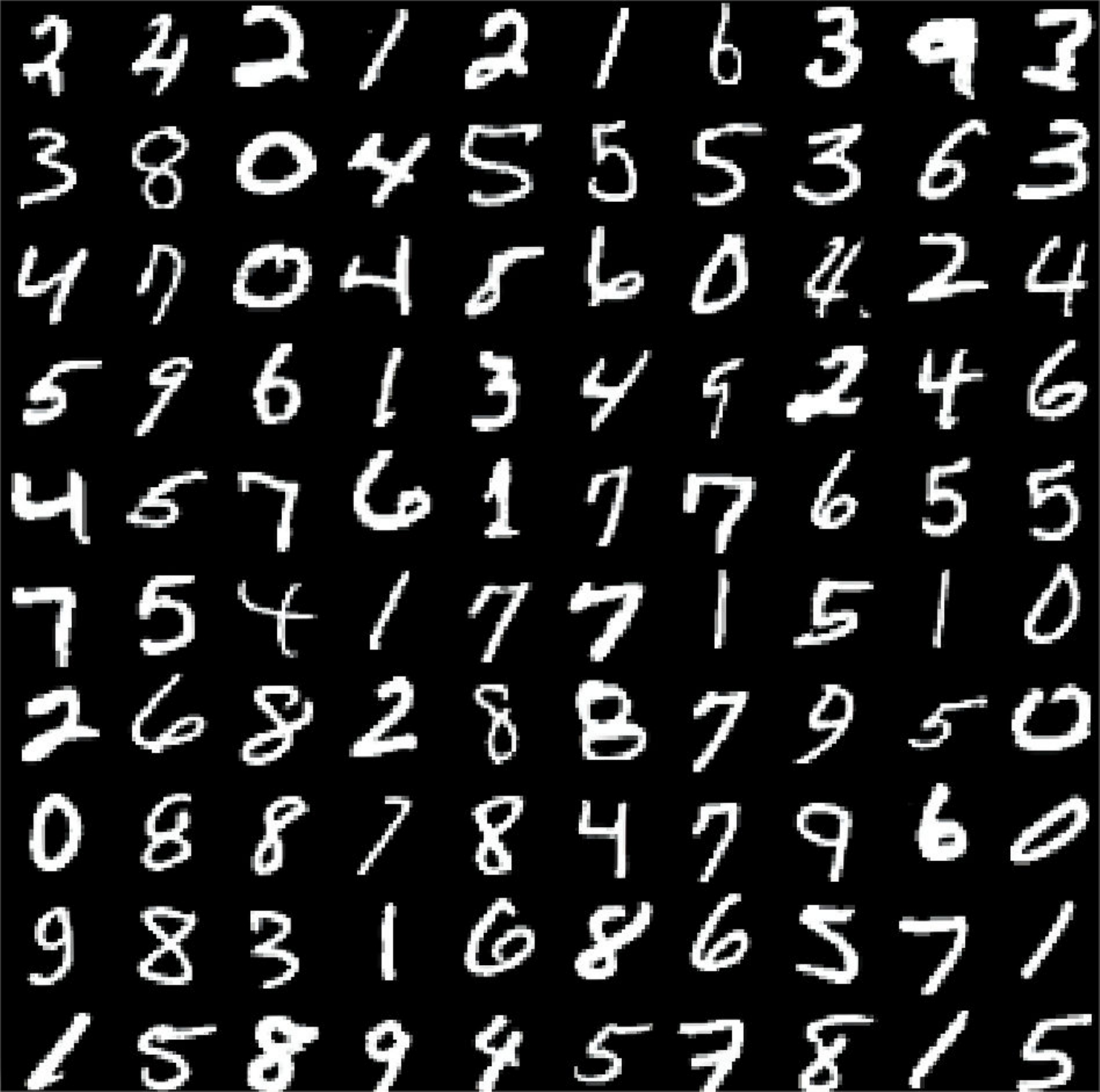}
\hfill
\includegraphics[width=0.32\linewidth]{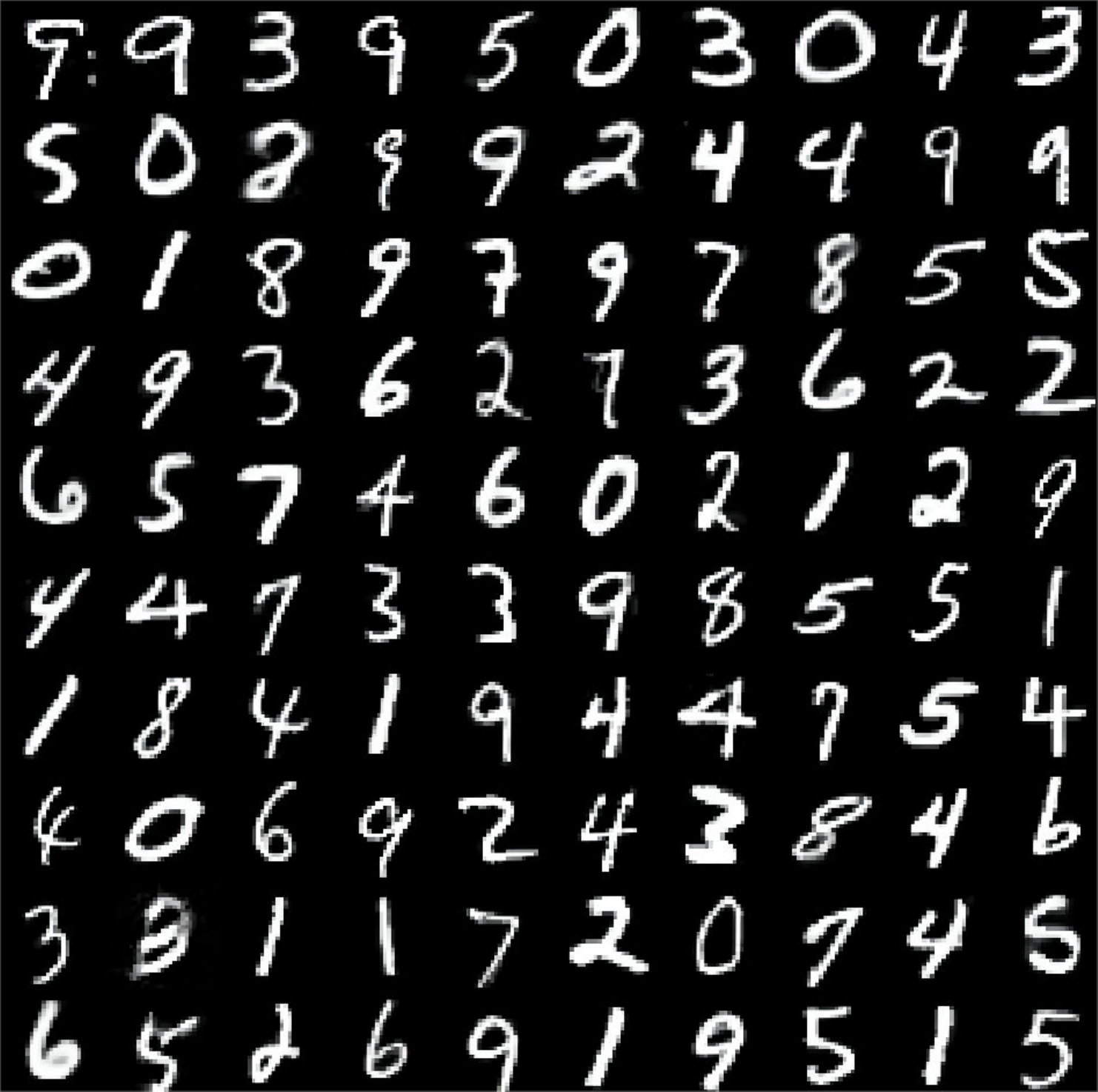}
\hfill
\includegraphics[width=0.32\linewidth]{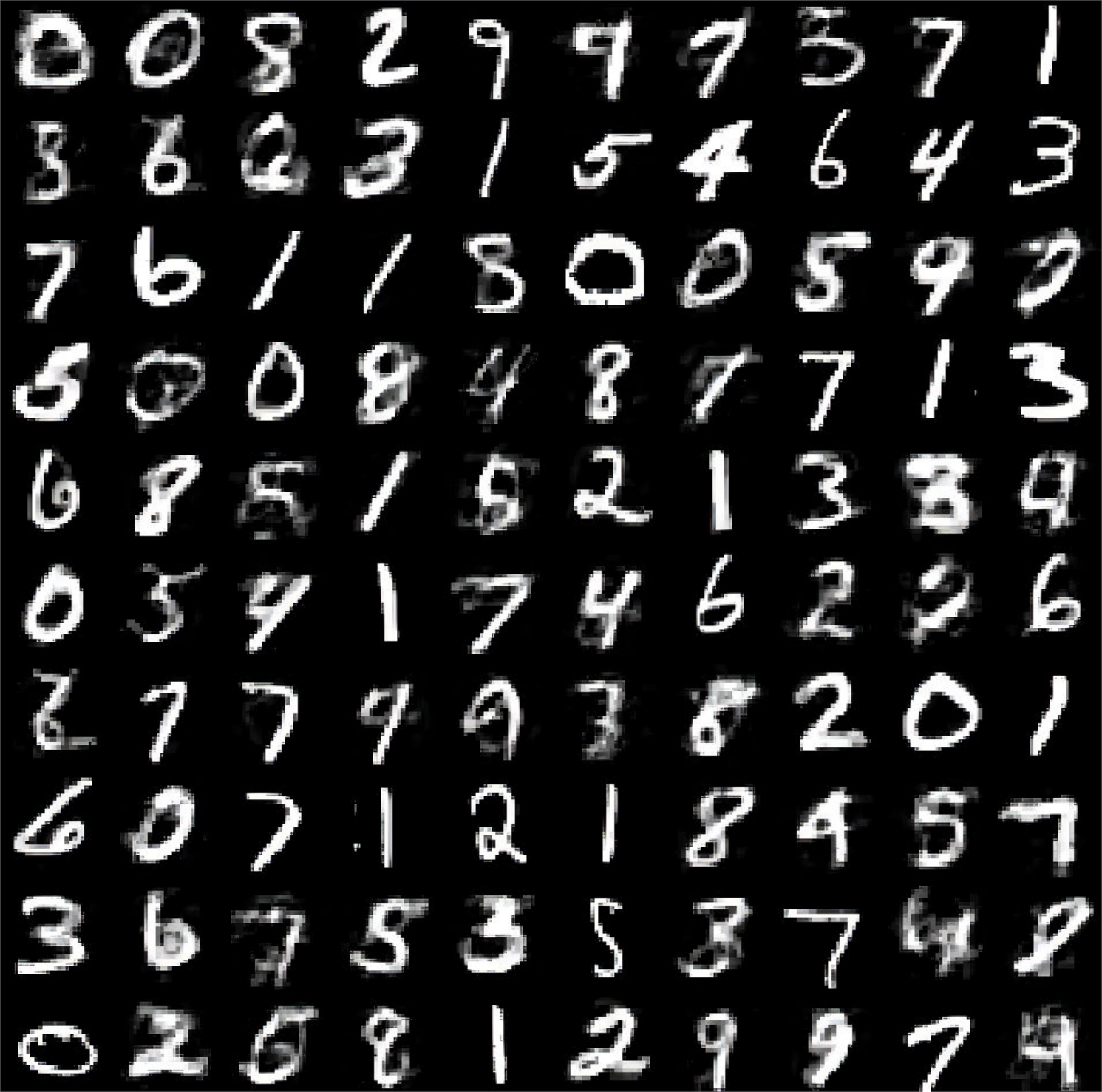}
\end{subfigure}	

\vspace{1em}
\begin{subfigure}{\linewidth}
\includegraphics[width=.32\linewidth]{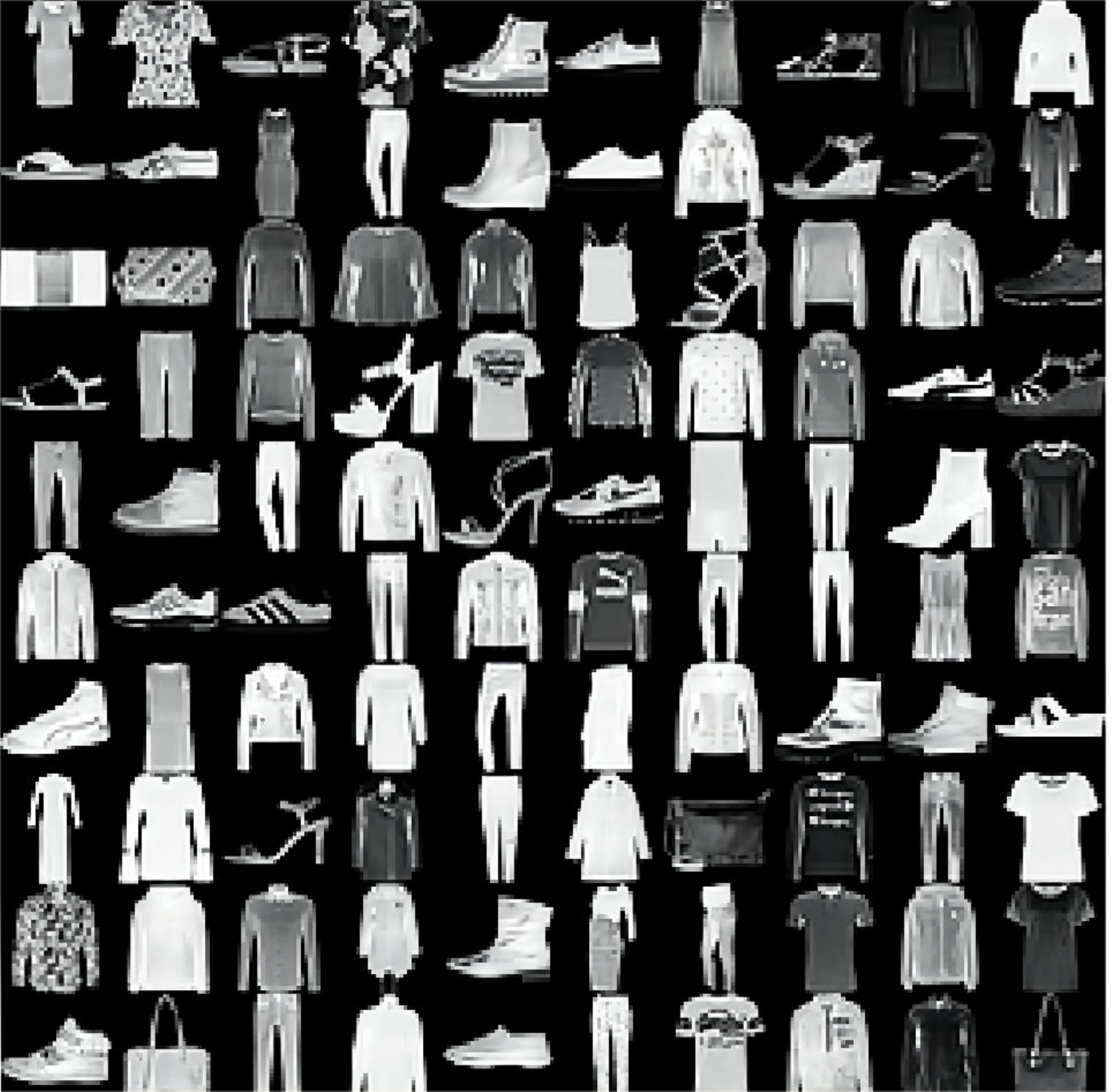}
\hfill
\includegraphics[width=.32\linewidth]{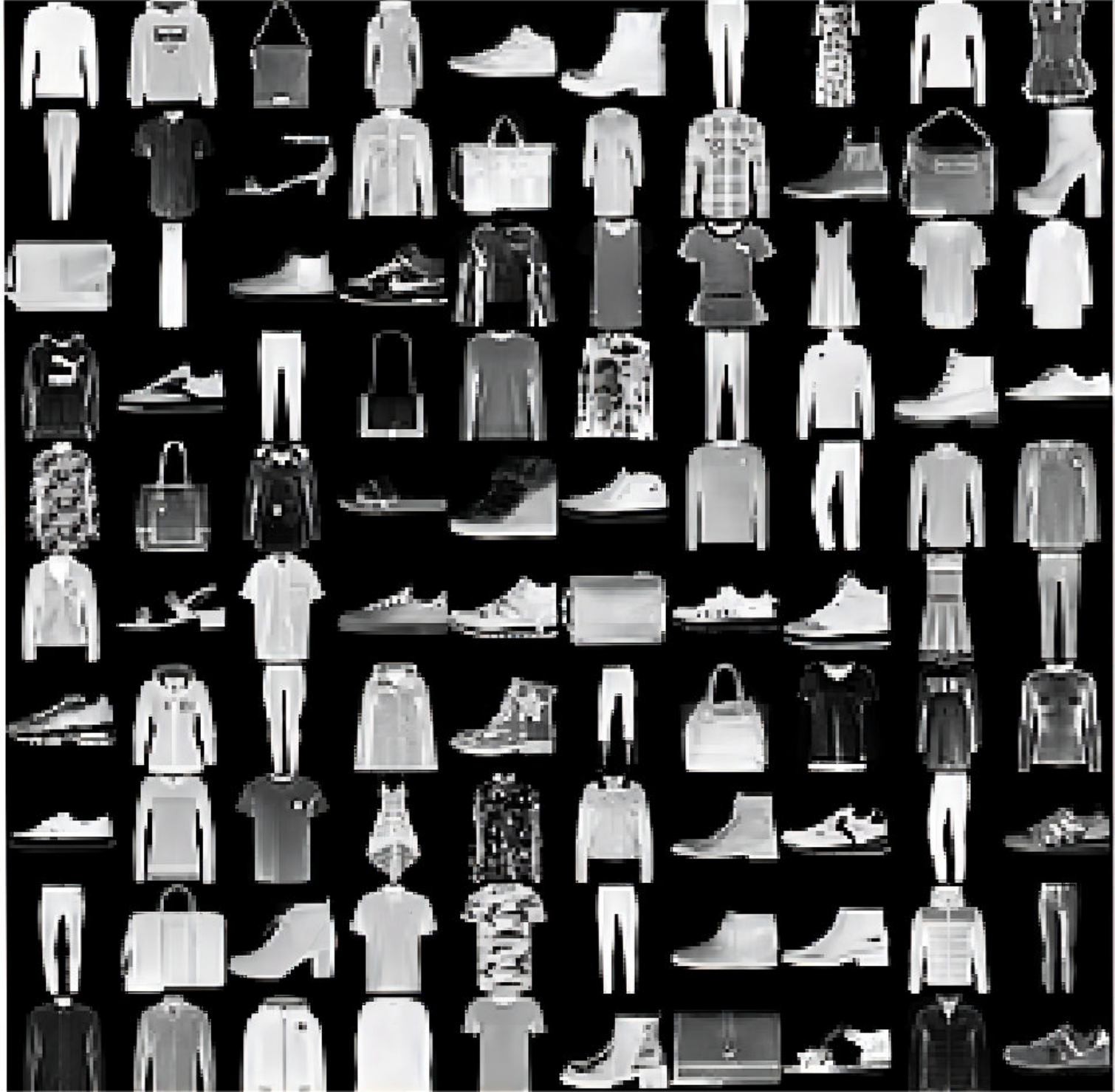}
\hfill
\includegraphics[width=0.32\linewidth]{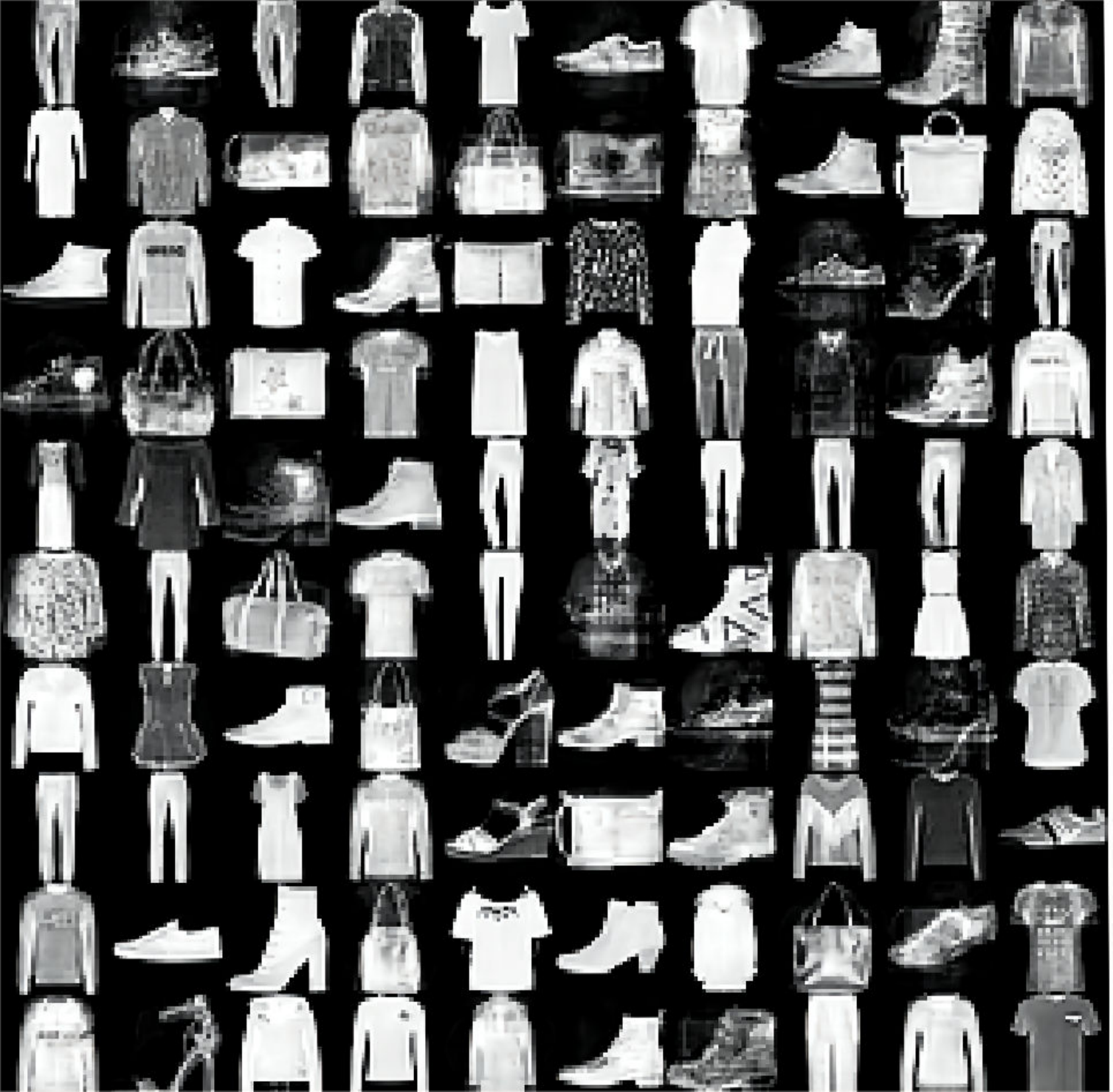}
\end{subfigure}	

\vspace{1em}
\begin{subfigure}{\linewidth}
\includegraphics[width=.32\linewidth]{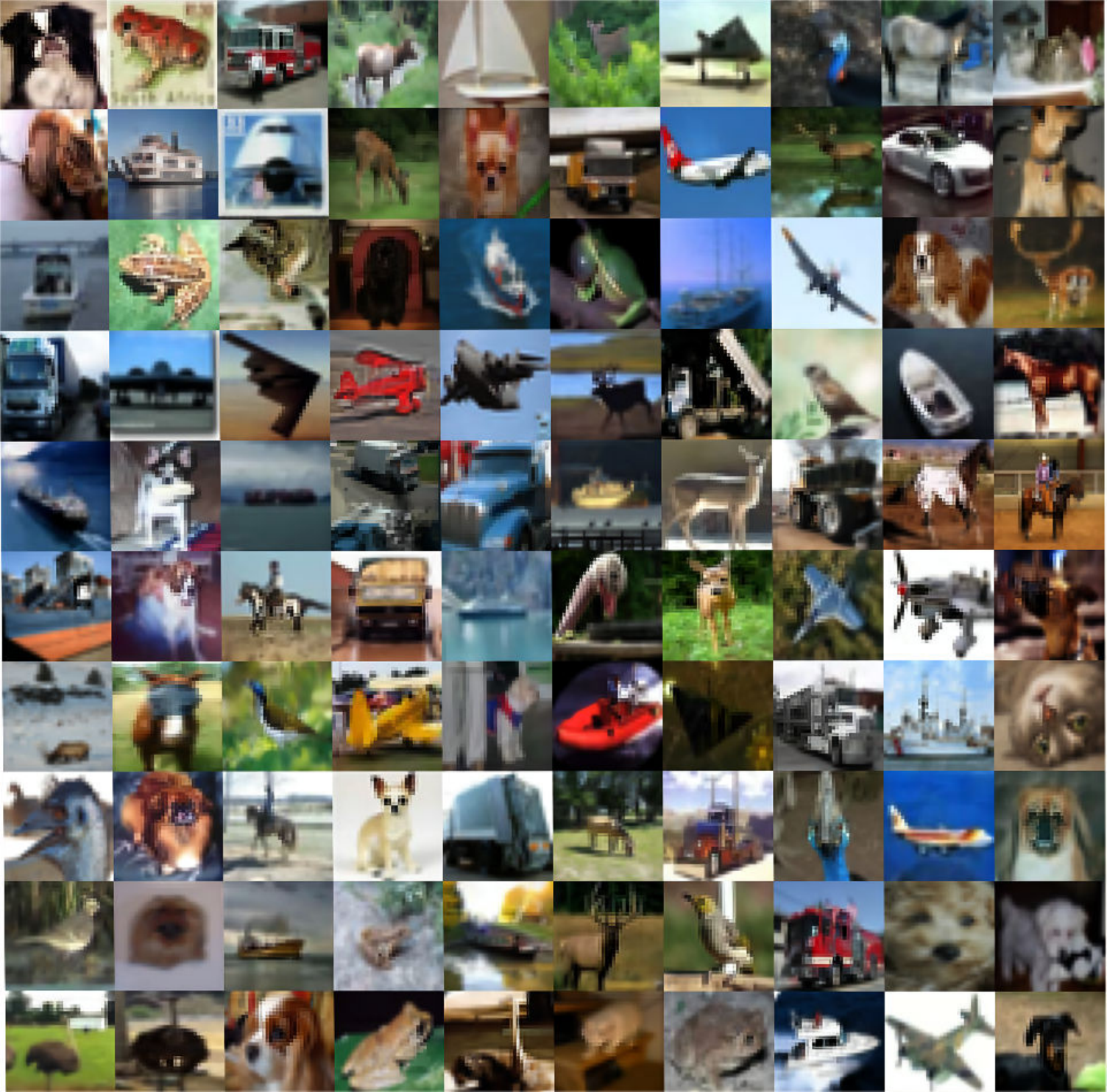}
\hfill
\includegraphics[width=.32\linewidth]{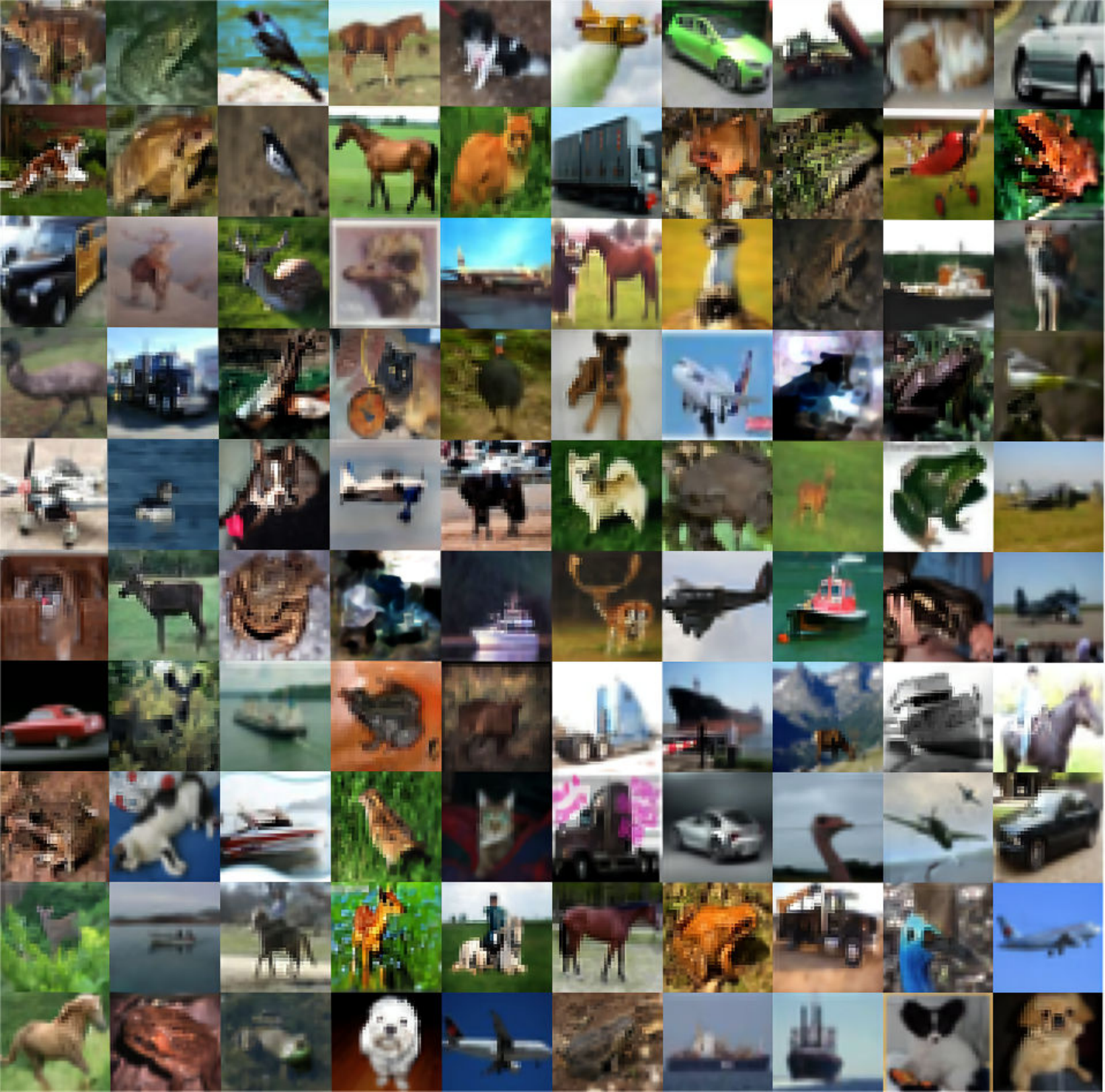}
\hfill
\includegraphics[width=0.32\linewidth]{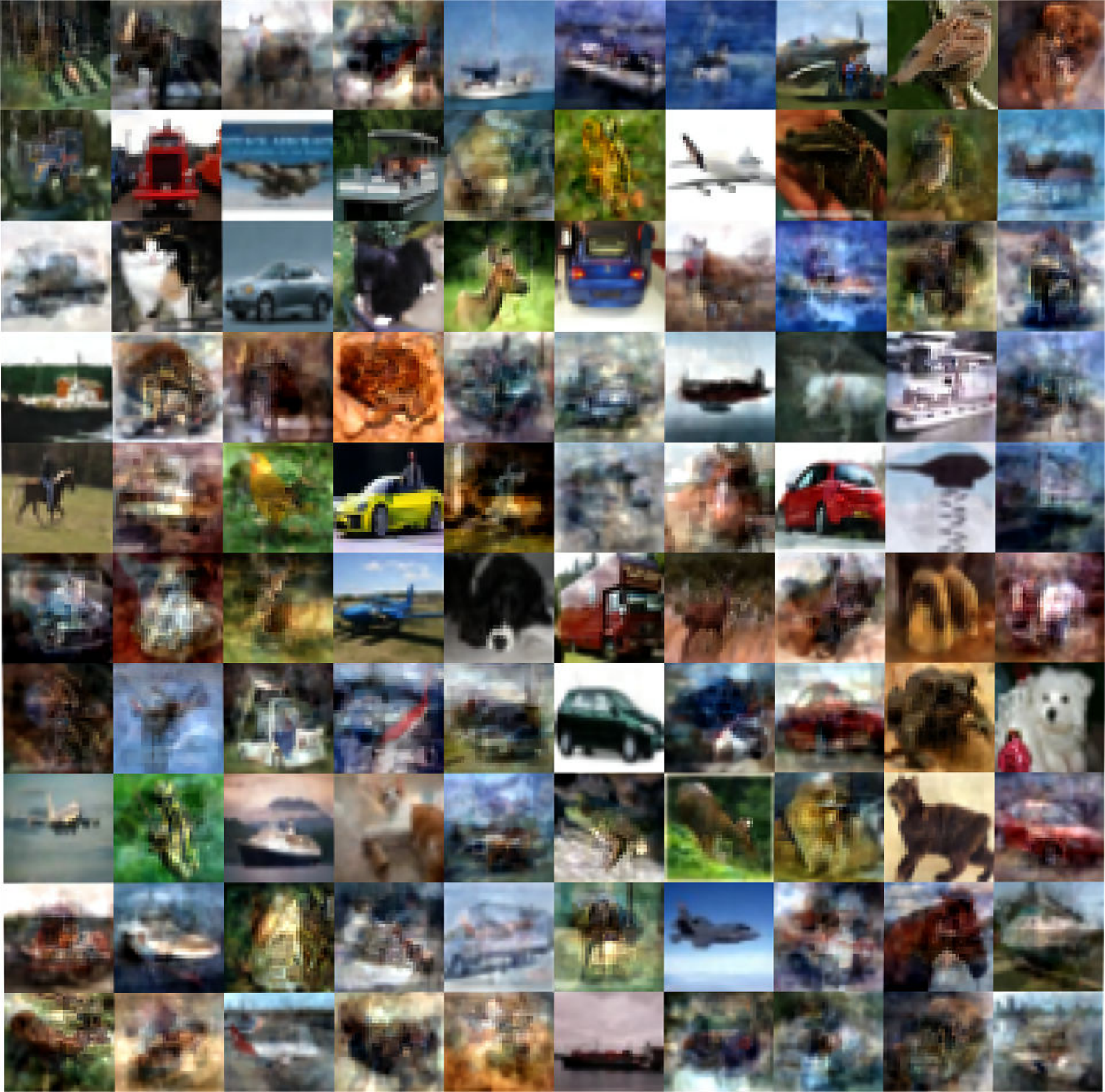}
\end{subfigure}
\caption{Visual comparison between 100 training data randomly chosen from the original training data (left column), new images generated by EVI-MMD (middle column), and EVI-Energy-Distance (right column).}
\label{fig:gen}
\end{figure}

We generate $N=100$ new imagines using the EVI-MMD method and the EVI-Energy-Distance method defined in Section \ref{sub:highdim}\footnote[1]{We do not include the Support-Points method in this comparison because some issues with the R codes by \cite{mak2018support}. We are not sure the reason but the returned results do not show any sign of convergence.}. 
The initial particles are sampled from a uniform distribution in $[0,1]^{d}$. 
We terminate both algorithms at $\texttt{maxIter}=500$ for the EVI-MMD method. 
Figure \ref{fig:gen} compares side-by-side 100 training data randomly chosen from the original training data (left column), new images generated by EVI-MMD (middle column), and EVI-Energy-Distance (right column). 
Both the training and new images are put into a $10\times 10$ panel. 
We can see the new images generated by both methods are very similar to the training data, and the EVI-MMD returns slightly sharper images than the EVI-Energy-Distance method. 

\begin{figure}[htb]
\centering
\begin{subfigure}{\linewidth}
\includegraphics[width=0.33\linewidth]{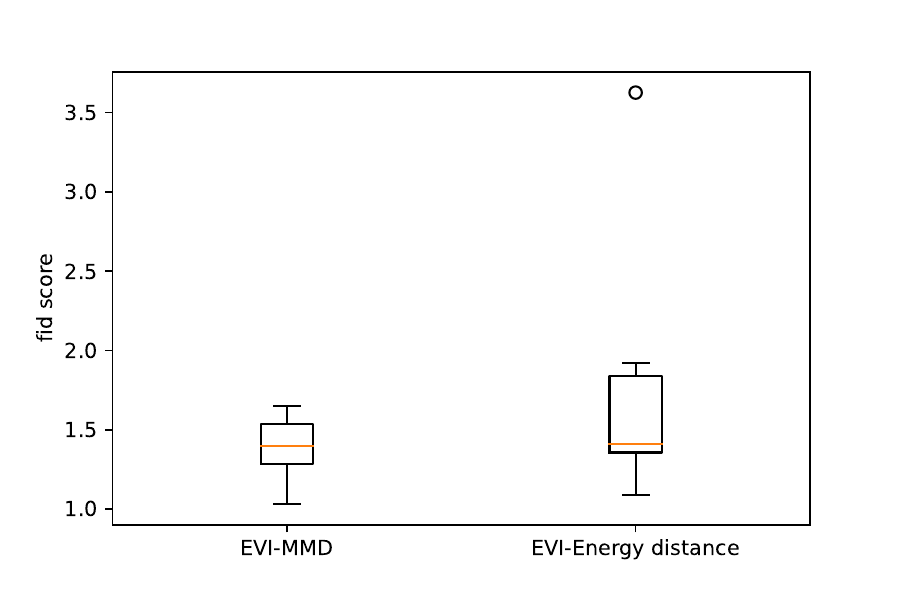}
\includegraphics[width=0.33\linewidth]{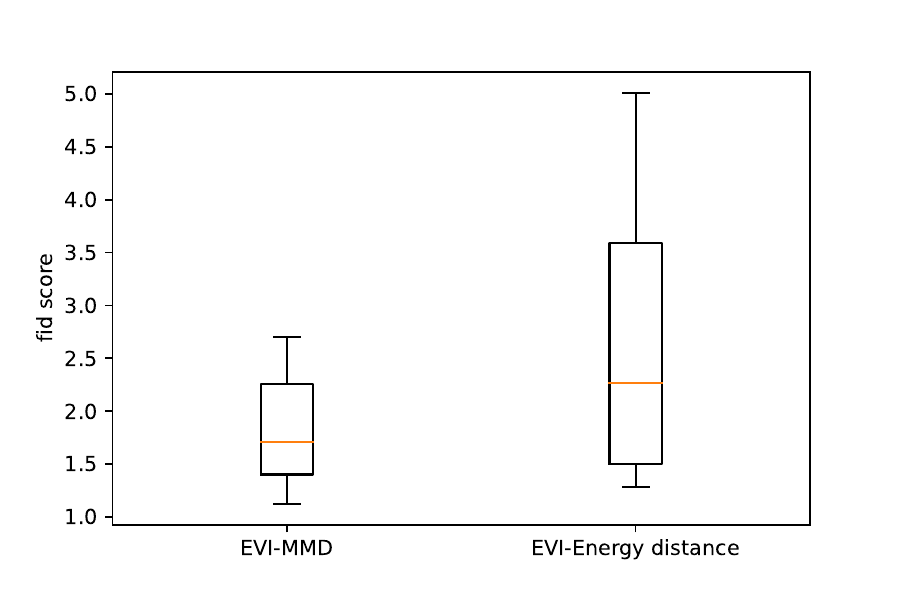}
\includegraphics[width=0.33\linewidth]{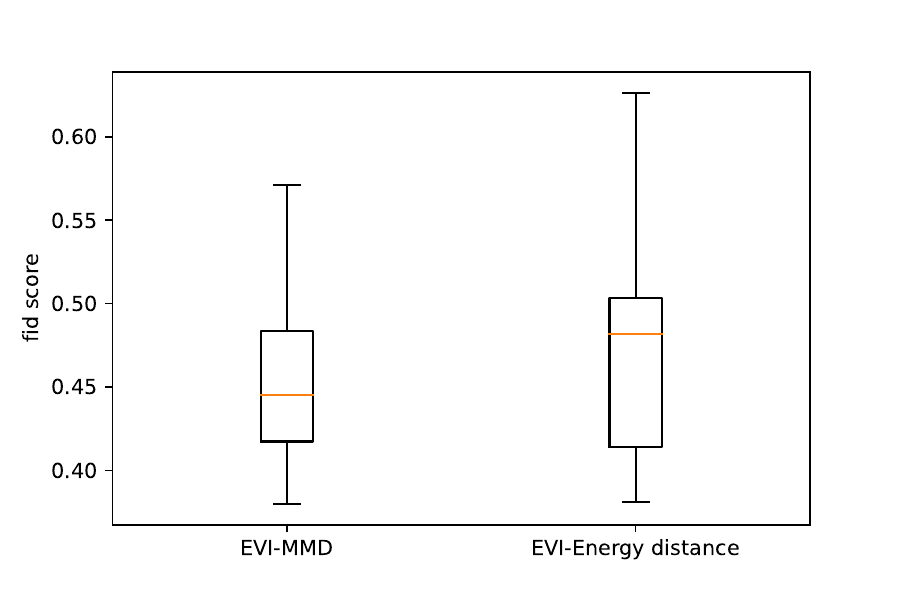}
\end{subfigure}
\caption{Boxplot of FID scores for the examples, MNIST, MNIST-Fashion and CIFAR10 returned by EVI-MMD (left box) and EVI-Energy-Distance (right box).}
\label{fig:higdim1}
\end{figure}

To provide a numerical comparison of the two methods, we calculate the FID score \citep{DBLP:journals/corr/HeuselRUNKH17} of the generated images. 
Due to the high computational cost, we randomly sample a subset of 500 images from the training data and calculate the FID score between the new images and the subset of training images. 
Repeating this 10 times we obtain 10 FID scores for each example. 
Figure \ref{fig:higdim1} shows the boxplots of all the FID scores for the two methods for all three examples. 
We can see that both methods have a comparable FID score and the EVI-MMD outperforms the EVI-Energy-Distance in the MNIST and Cifar10 examples, which is consistent with the visual comparison in Figure \ref{fig:gen}. 

\begin{figure}[htb]
\centering
\begin{subfigure}{\linewidth}
\includegraphics[width=.32\linewidth]{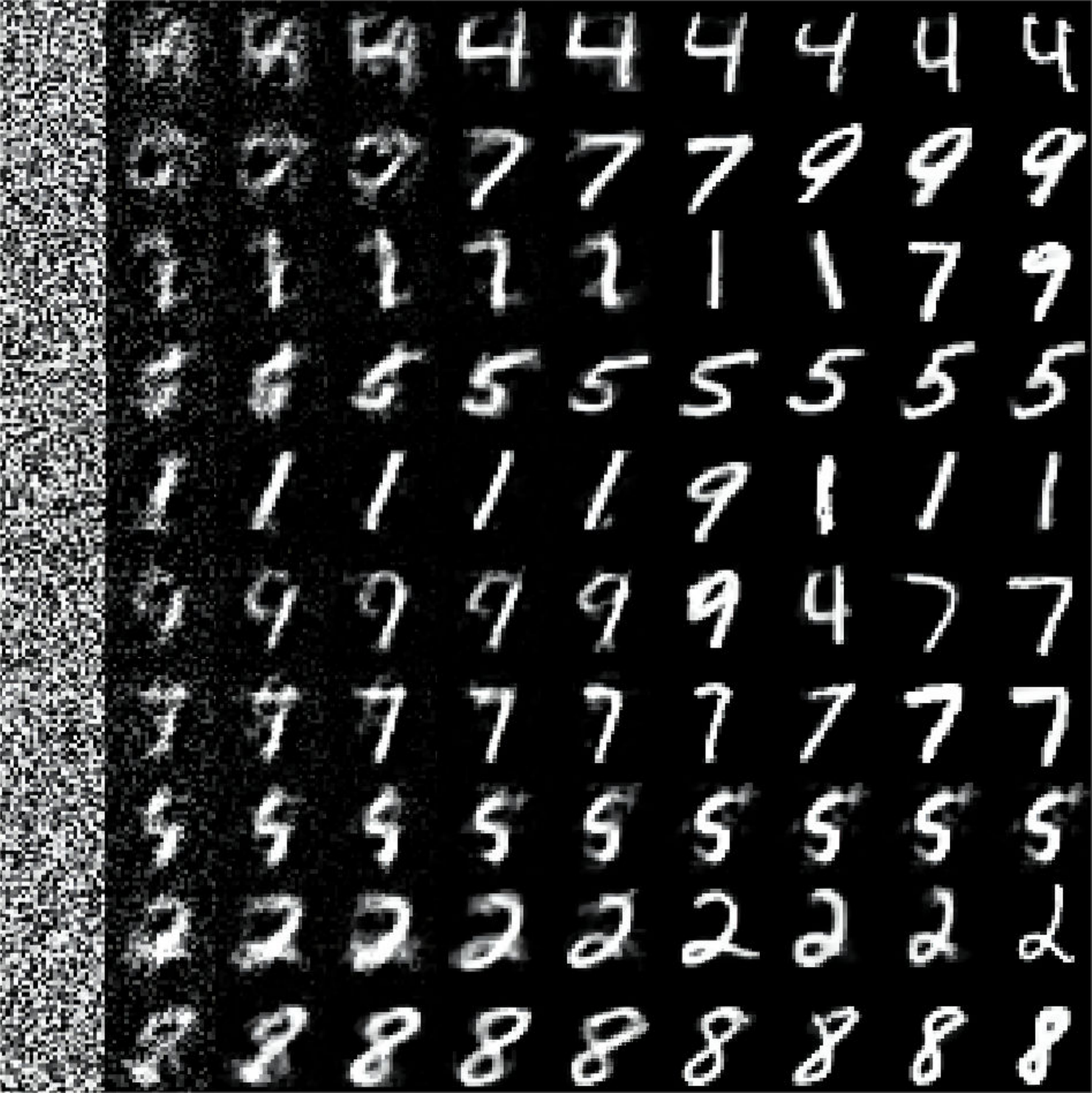}\hfill
\includegraphics[width=.33\linewidth]{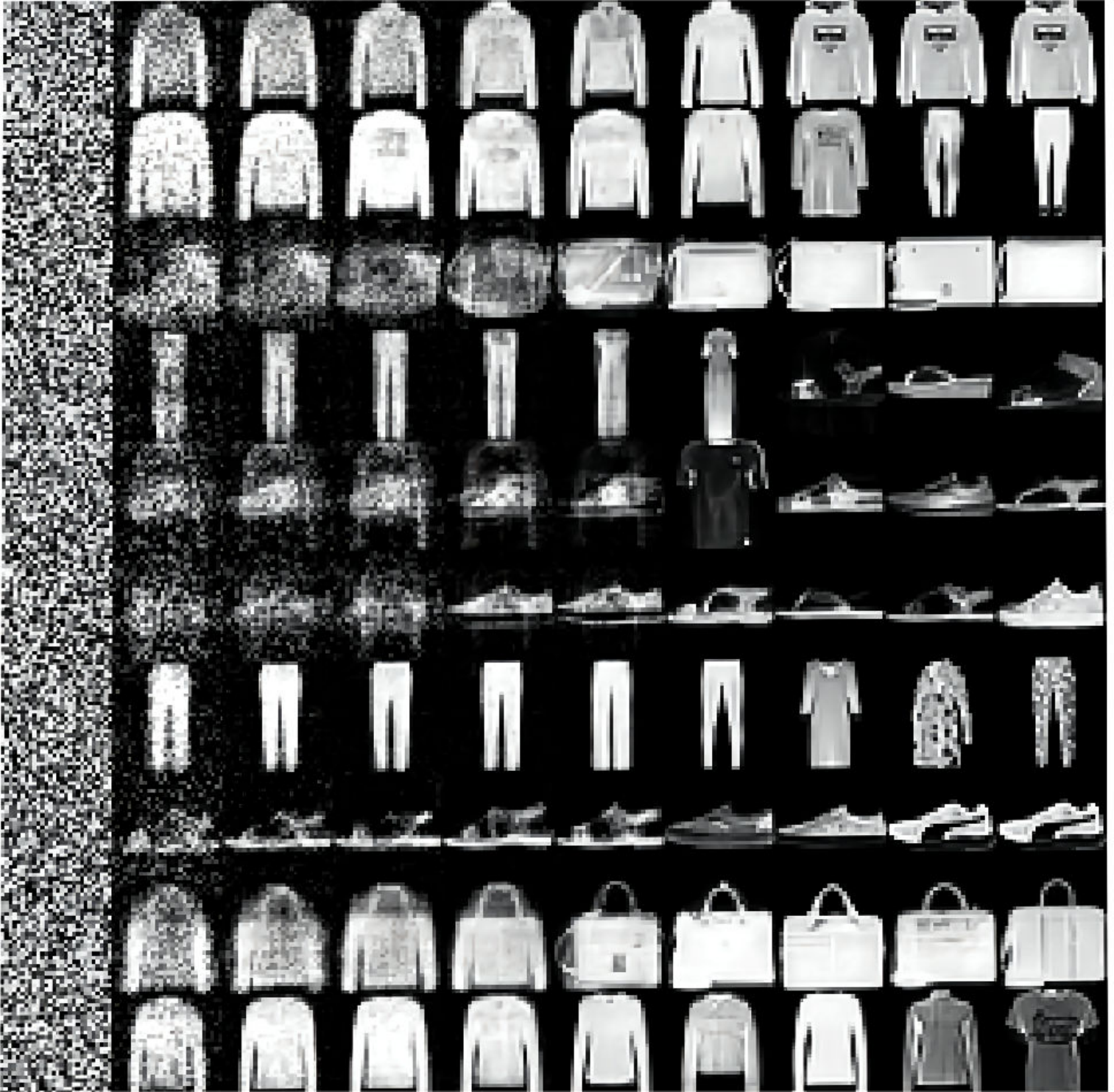}\hfill
\includegraphics[width=.33\linewidth]{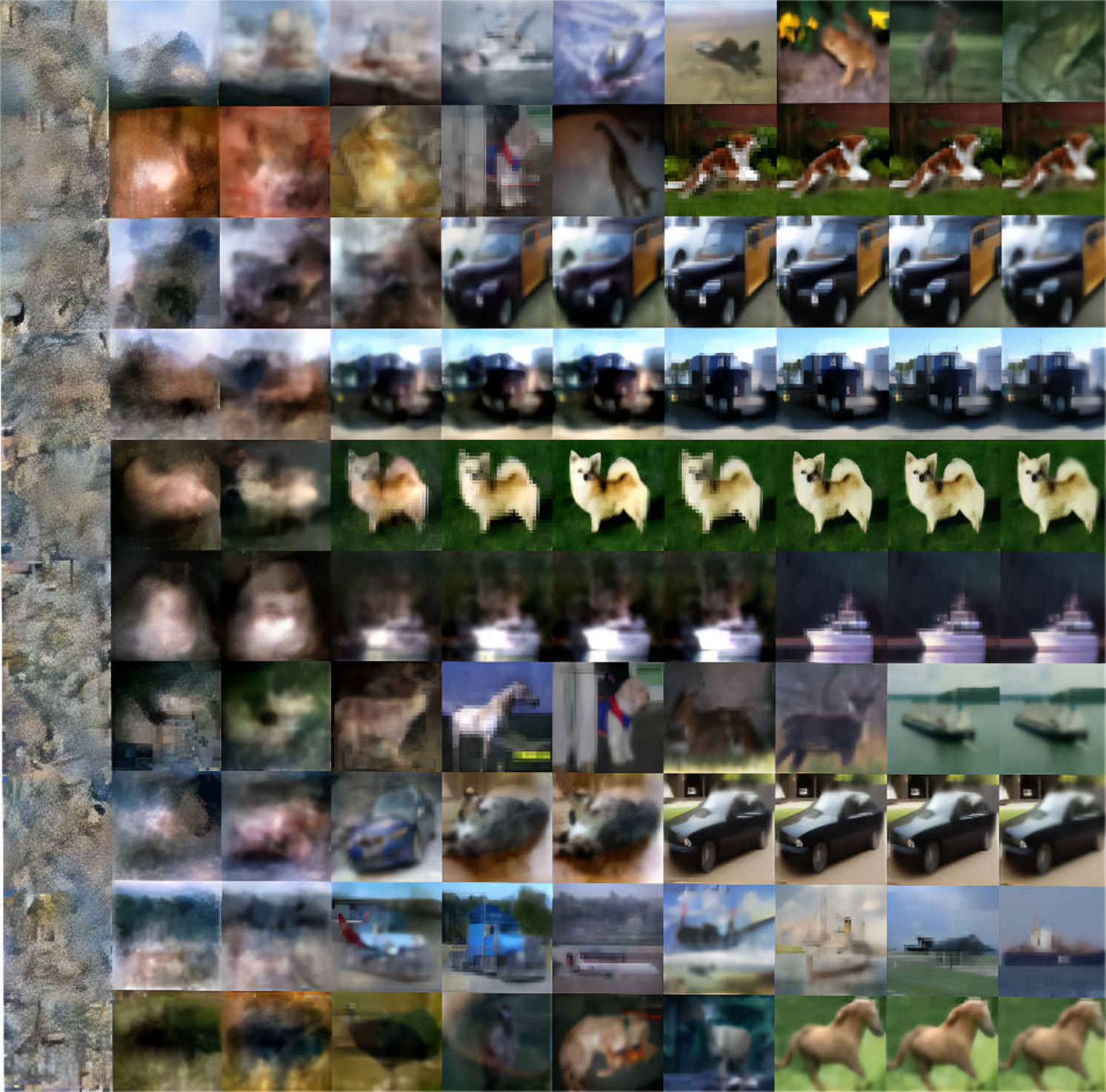}
\end{subfigure}
\caption{The trajectories of the movement of some randomly picked particles of the EVI-MMD method.}
\label{fig:gen2}
\end{figure}

We also provide the trajectory of the EVI-MMD method in Figure \ref{fig:gen2}. 
It shows the evolution of some randomly picked particles from static images to the final sharp images. 
Interestingly, some images are switched directions in the middle of their trajectories. 
For example, in the second row of the middle panel, the image suddenly switched from a top to a dress then to a pair of pants. 
This is because when we resample a mini-batch in each iteration. 
New images can appear in the mini-batch while some old images are not selected, causing the particles to constantly change to approximate the mini-batch data.
Note that the EVI-MMD is by no means the best approach for the MNIST benchmark example. 
Readers can find many more sophisticated generative learning approaches that return better results. 
However, the EVI-MMD is probably the simplest by comparison and its results are adequate.

\FloatBarrier

\section{Conclusion}\label{sec:con}

In this paper, we develop a new deterministic sampling method to approximate a target distribution by minimizing the kernel discrepancy, alternatively known as maximum mean discrepancy (MMD). 
The minimization of MMD is solved by the general energetic variational inference (EVI) framework first introduced by \cite{wang2021particle}. 
Specifically, we use the quadratic dissipation functional of the EVI and apply the particle approximation to the continuous energy-dissipation law, which is then followed by the variation procedure. 
This leads to a dynamic system that moves the particles from their initial positions to the target distribution.
Using the implicit Euler scheme to solve this dynamic system, we obtain a special algorithm based on the EVI framework to minimize MMD, which we call EVI-MMD algorithm.
In each iteration of updating the particles, we solve a proximal minimization problem using algorithms such as L-BFGS. 
To overcome the long-existing issue of bandwidth selection of the Gaussian kernel, we propose a novel way to specify the bandwidth dynamically. 
Other than approximating a fully specified distribution that is difficult to sample from, the EVI-MMD algorithm can also be used as a generative learning model for two-sample problems. 

The EVI-MMD method also has the following limitations. First, The objective function in EVI-MMD involves pairwise interactions between particles, resulting in a computational complexity of $O(N^2)$ per iteration. While we employ mini-batching techniques for the target data in two-sample problems, the scaling with respect to the number of generated particles ($N$) remains a challenge for very large-scale applications. Second, the proposed dynamic bandwidth selection strategy is currently heuristic. While our extensive ablation studies in the Supplementary Material confirm its robustness and effectiveness, a rigorous theoretical proof for the optimal decay rate ($c$) and its impact on the convergence rate of the MMD flow is still needed. Third, although the ``median trick'' simplifies the selection of the scale parameter $a$, the algorithm's performance can still be sensitive to the choice of the decay parameter $c$ and the time step $\tau$ in extremely high-dimensional or highly multi-modal spaces. Further research into automated tuning or line-search methods for these parameters could enhance the algorithm's ``plug-and-play'' capability.

The EVI framework is very general. 
Many new variational inference methods can be proposed if we choose different combinations of the key components of the EVI framework. 
Such key components include the choice of divergence functionals, order of variational and discretization of the continuous EVI scheme, using parametric or non-parametric model for the flow map, and various numerical schemes for solving the dynamic system, etc. 
In fact, some existing methods can be also included in the EVI framework. 
The unified EVI framework also paves the way for a unified theoretical foundation for similar algorithms. 
We will keep pursuing these directions and fully explore the potentials of the EVI framework. 

\begin{center}
{\Large\bf Acknowledgment}
\end{center}
This research was done using services provided by the OSG Consortium \citep{osg07,osg09}, which is supported by the National Science Foundation awards \# 2030508 and \# 1836650.
L. Kang's work is partially supported by the National Science Foundation Grants DMS-1916467 and DMS-2153029.
Y. Wang and C. Liu are partially supported by the National Science Foundation Grant DMS-1759536, DMS-1950868, and DMS-2153029.

\bibliographystyle{asa}
\bibliography{MMD}

\newpage

\setcounter{page}{1}
\setcounter{figure}{0}
\setcounter{table}{0}
\setcounter{lem}{0}
\setcounter{theorem}{0}
\setcounter{proposition}{0}

\makeatletter 
\renewcommand{\thefigure}{S\@arabic\c@figure}
\renewcommand{\thetable}{S\@arabic\c@table}
\renewcommand{\thelem}{S\@arabic\c@lem}
\renewcommand{\theproposition}{S\@arabic\c@proposition}
\renewcommand{\thetheorem}{S\@arabic\c@theorem}
\renewcommand{\thesection}{A\@arabic\c@section}
\makeatother

\begin{center}
{\Large\bf Supplement: Tuning The Bandwidth Parameters of the Kernel Function}
\end{center}

Choosing the tuning parameters $a$, $c$, and $\tau$ based on sound theoretical results is challenging. In Section \ref{sub:finalalg}, we provided a rough rule-of-thumb for their selection. Here, we conduct a more detailed study of these parameters using the eight-component Gaussian mixture distribution.

\begin{figure}[!htb]
	\centering
	\begin{subfigure}{\linewidth}
		\makebox[0pt][r]{\makebox[30pt]{\raisebox{40pt}{\rotatebox[origin=c]{90}{$c=0.5$}}}}%
		\includegraphics[width=0.24\linewidth]{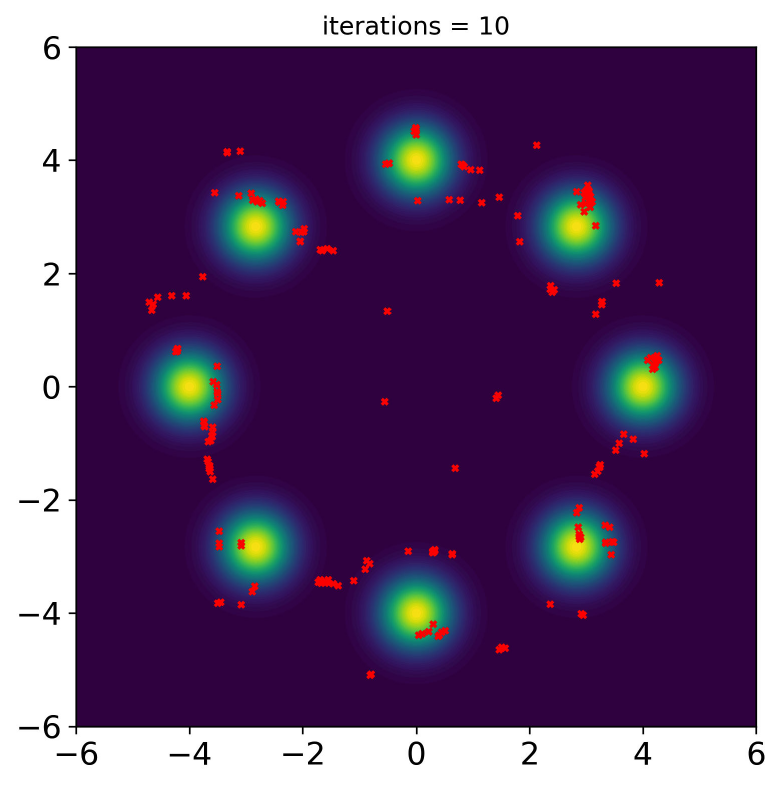}
		\includegraphics[width=0.24\linewidth]{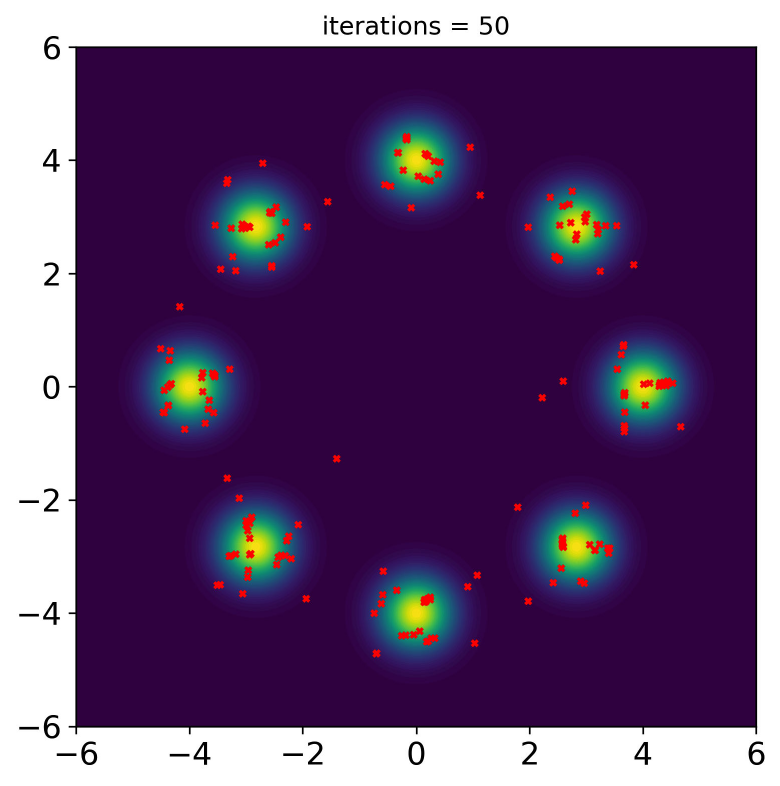}
		\includegraphics[width=0.24\linewidth]{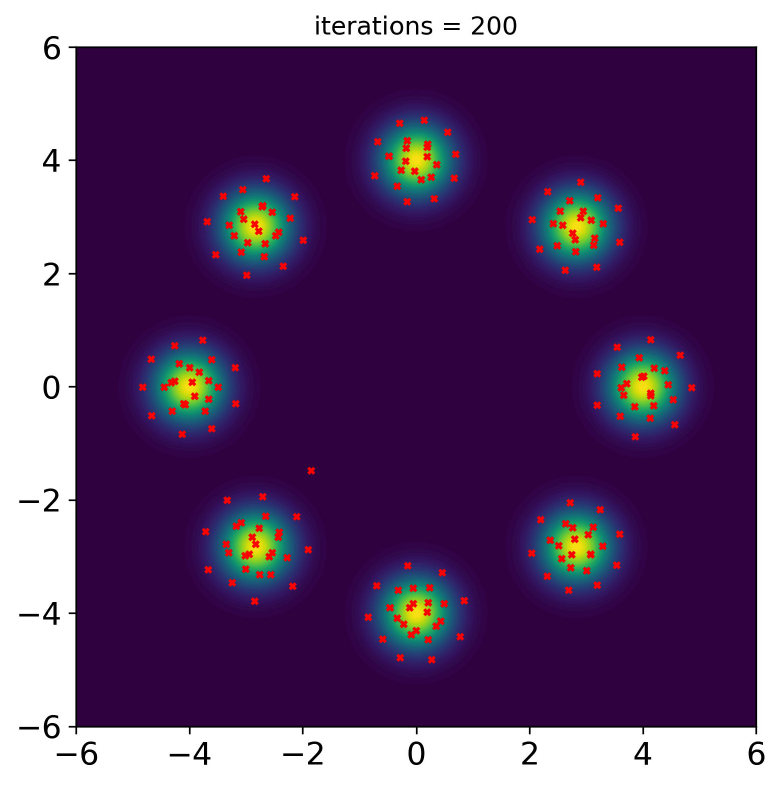}
		\includegraphics[width=0.24\linewidth]{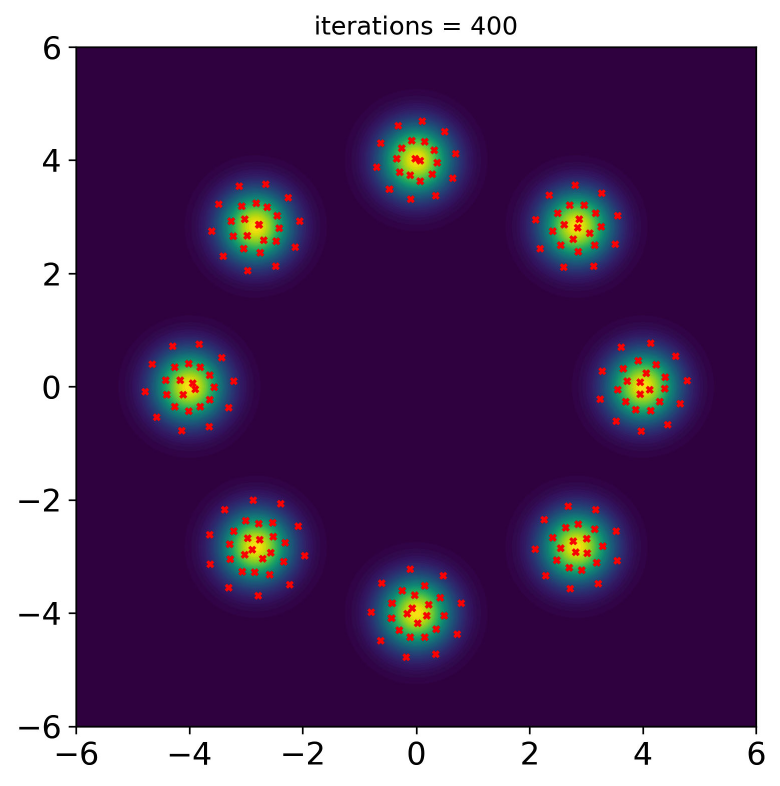}
	\end{subfigure}	
	\begin{subfigure}{\linewidth}
		\makebox[0pt][r]{\makebox[30pt]{\raisebox{40pt}{\rotatebox[origin=c]{90}{$c=0.4$}}}}%
		\includegraphics[width=.24\linewidth]{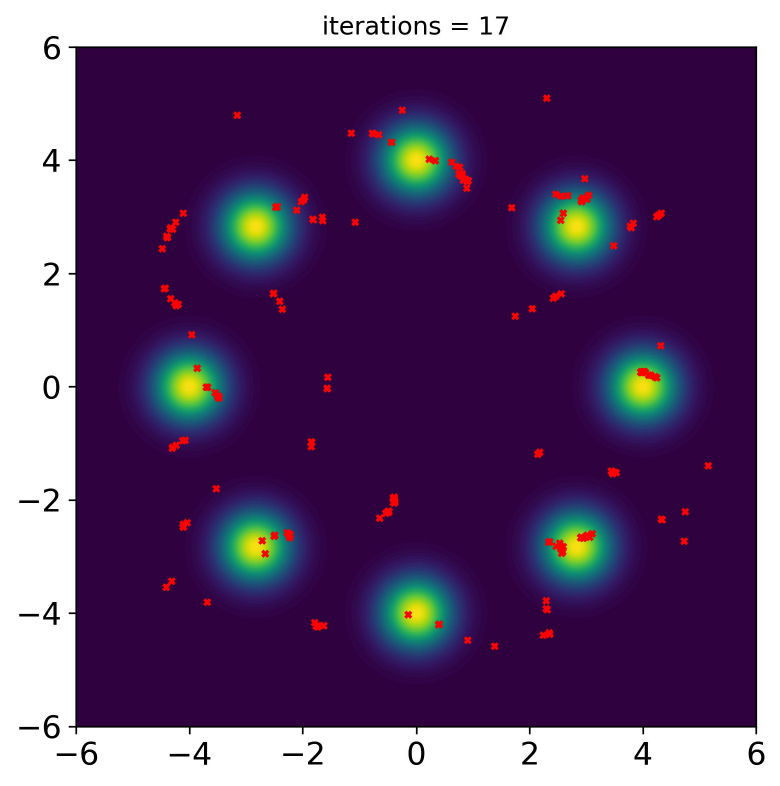}
		\includegraphics[width=.24\linewidth]{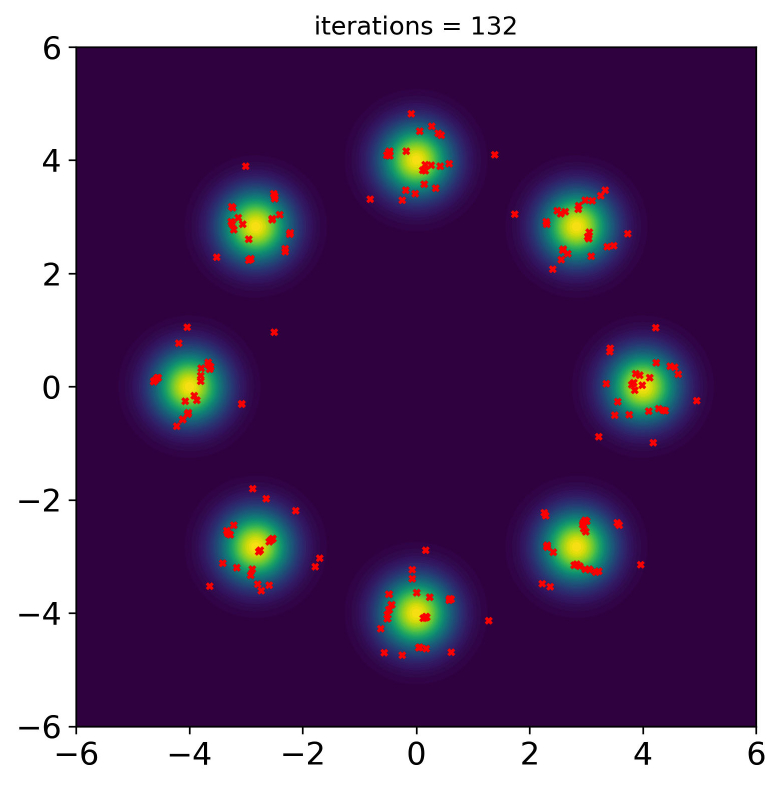}
		\includegraphics[width=0.24\linewidth]{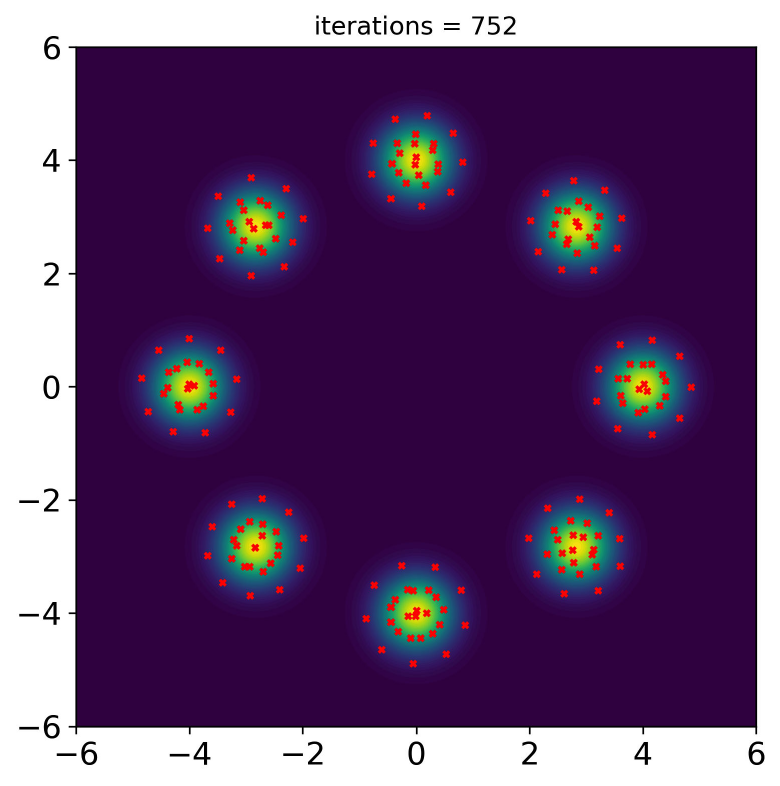}
		\includegraphics[width=0.24\linewidth]{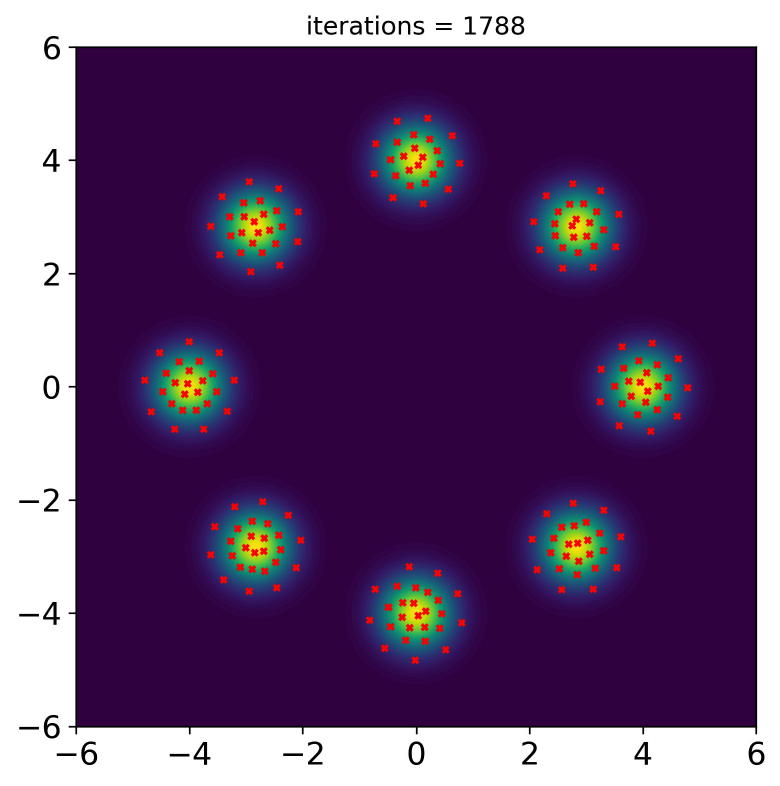}
	\end{subfigure}		
	\begin{subfigure}{\linewidth}
		\makebox[0pt][r]{\makebox[30pt]{\raisebox{40pt}{\rotatebox[origin=c]{90}{$c=0.6$}}}}%
		\includegraphics[width=.24\linewidth]{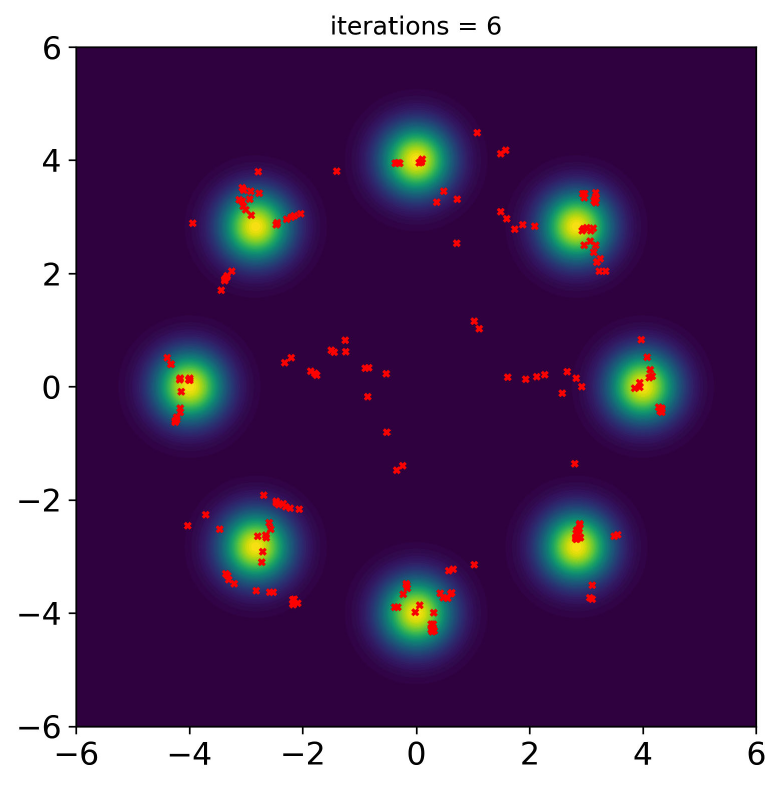}
		\includegraphics[width=.24\linewidth]{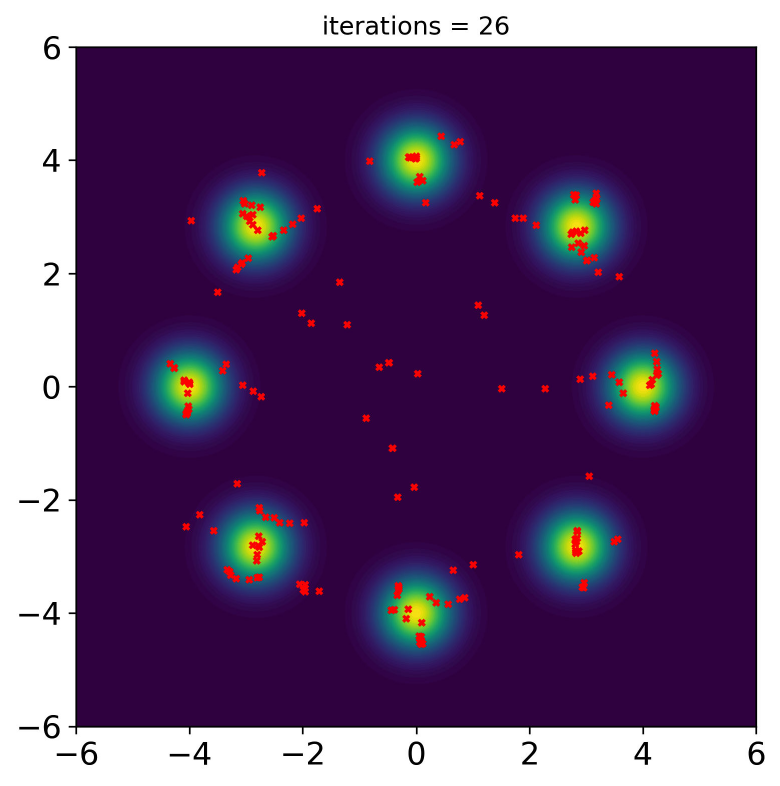}
		\includegraphics[width=0.24\linewidth]{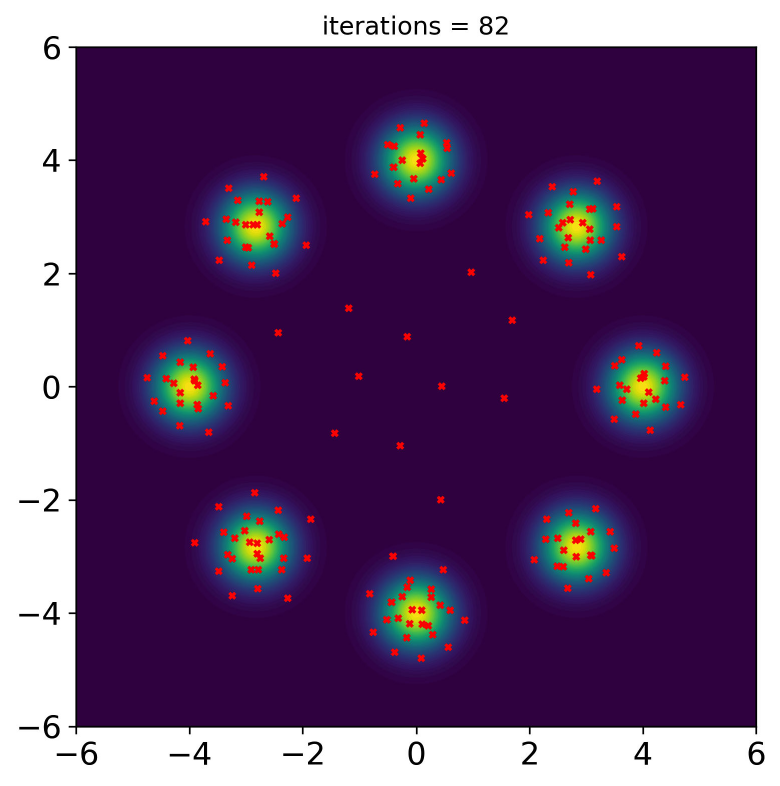}
		\includegraphics[width=0.24\linewidth]{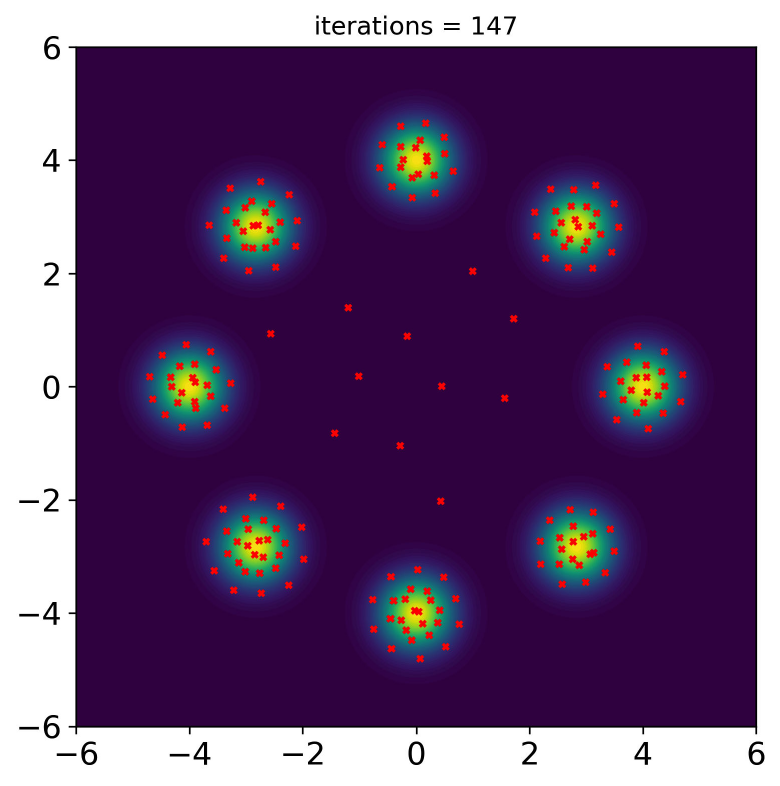}
	\end{subfigure}
	\caption{Particle trajectories of different choice of $c$. In the first row, $c = 0.5$ as in the toy example; in the second row: $c = 0.4$; in the third row: $c = 0.6$. For each figure of the same column, the value of $h_n$ is similar.}
	\label{fig:tune1}
\end{figure}

\begin{figure}[!htb]
	\centering
	\begin{subfigure}{\linewidth}
		\makebox[0pt][r]{\makebox[30pt]{\raisebox{40pt}{\rotatebox[origin=c]{90}{$a\approx4$}}}}%
		\includegraphics[width=0.24\linewidth]{base_11.pdf}
		\includegraphics[width=0.24\linewidth]{base_15.pdf}
		\includegraphics[width=0.24\linewidth]{base_30.pdf}
		\includegraphics[width=0.24\linewidth]{base_50.pdf}
	\end{subfigure}	
	\begin{subfigure}{\linewidth}
		\makebox[0pt][r]{\makebox[30pt]{\raisebox{40pt}{\rotatebox[origin=c]{90}{$a=2$}}}}%
		\includegraphics[width=.24\linewidth]{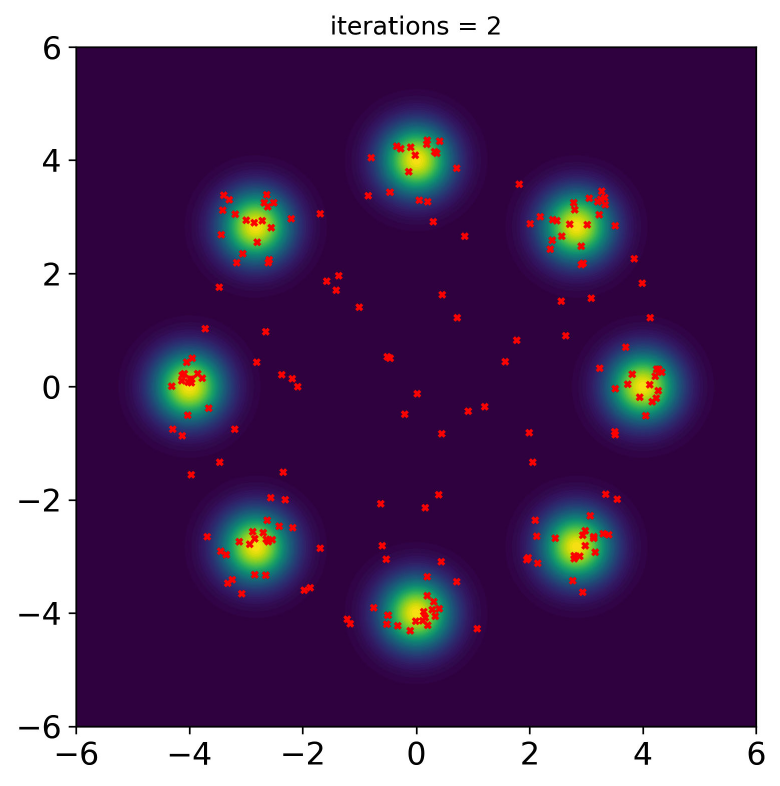}
		\includegraphics[width=.24\linewidth]{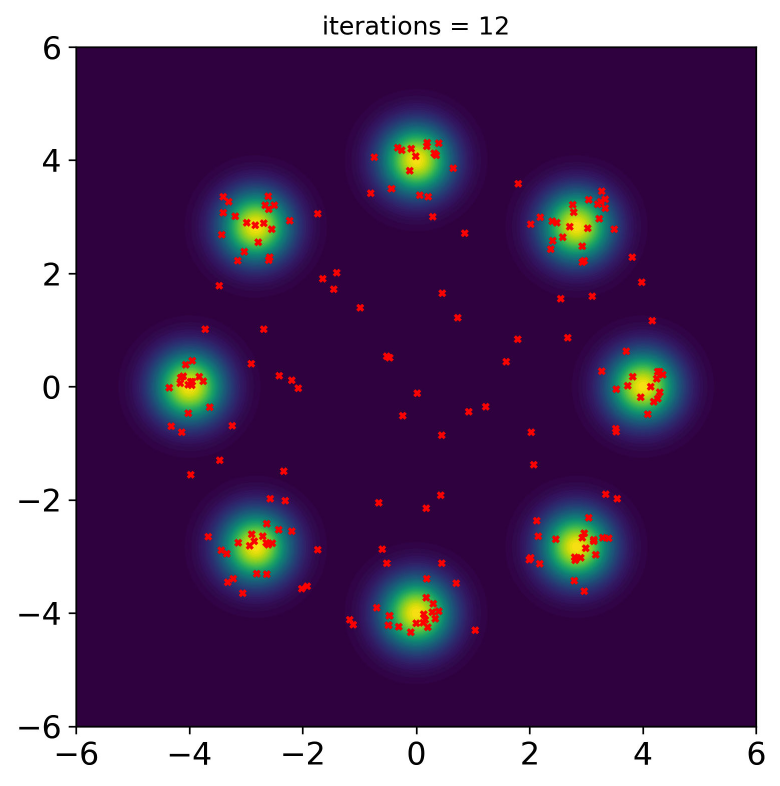}
		\includegraphics[width=0.24\linewidth]{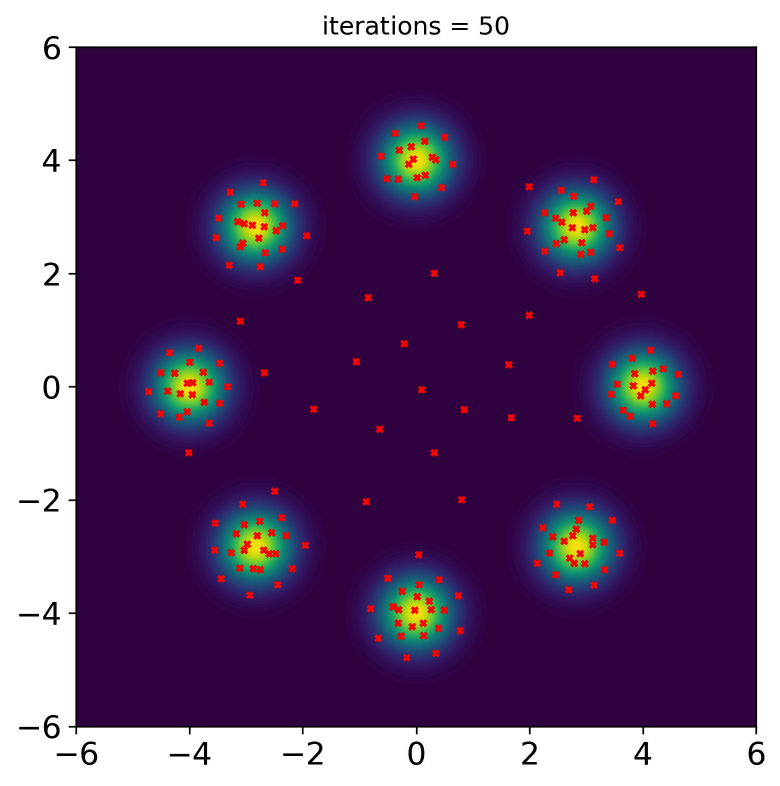}
		\includegraphics[width=0.24\linewidth]{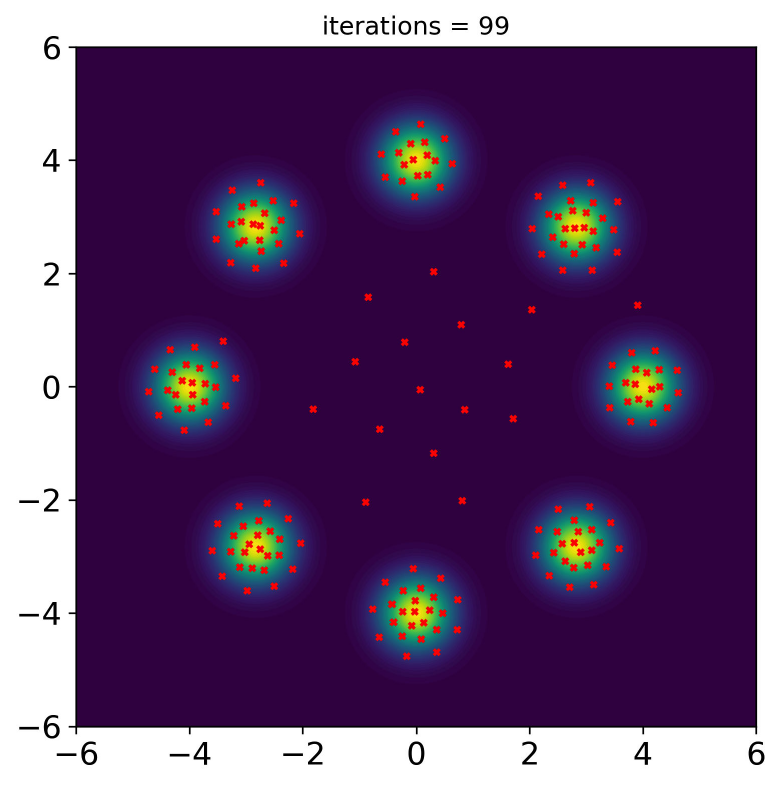}
	\end{subfigure}
		\begin{subfigure}{\linewidth}
		\makebox[0pt][r]{\makebox[30pt]{\raisebox{40pt}{\rotatebox[origin=c]{90}{$a=3$}}}}%
		\includegraphics[width=.24\linewidth]{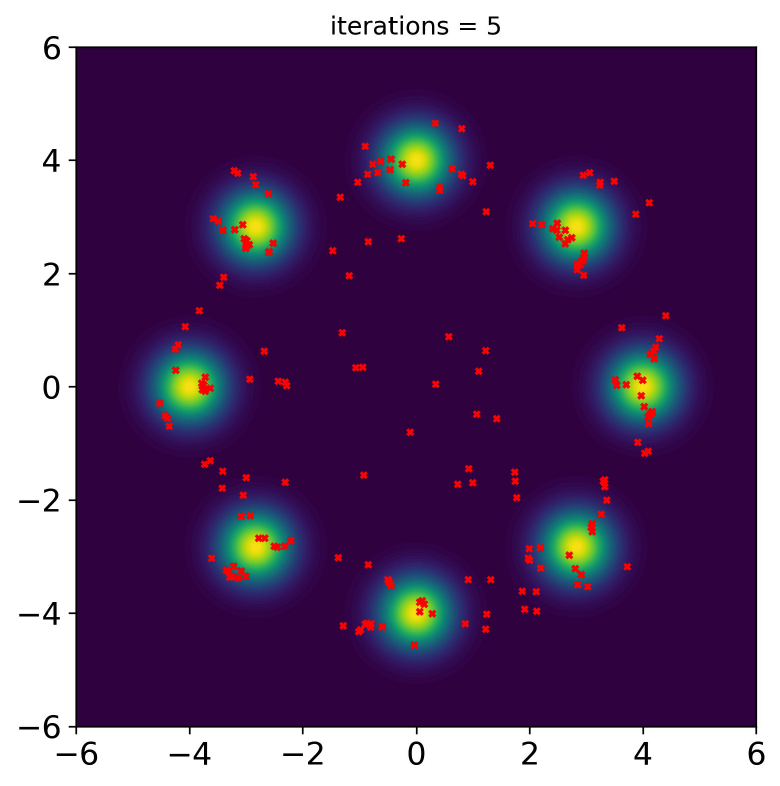}
		\includegraphics[width=.24\linewidth]{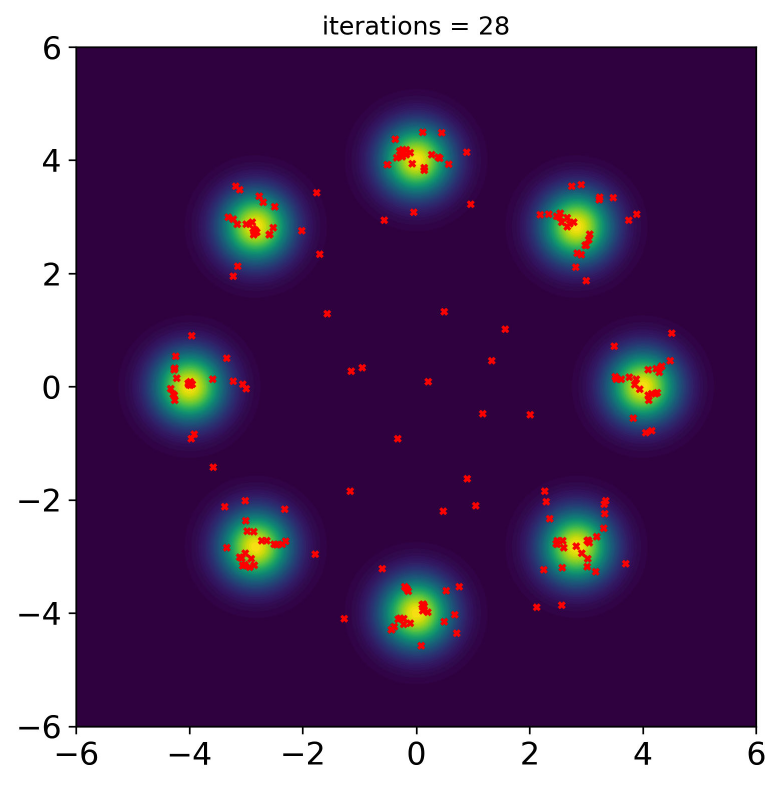}
		\includegraphics[width=0.24\linewidth]{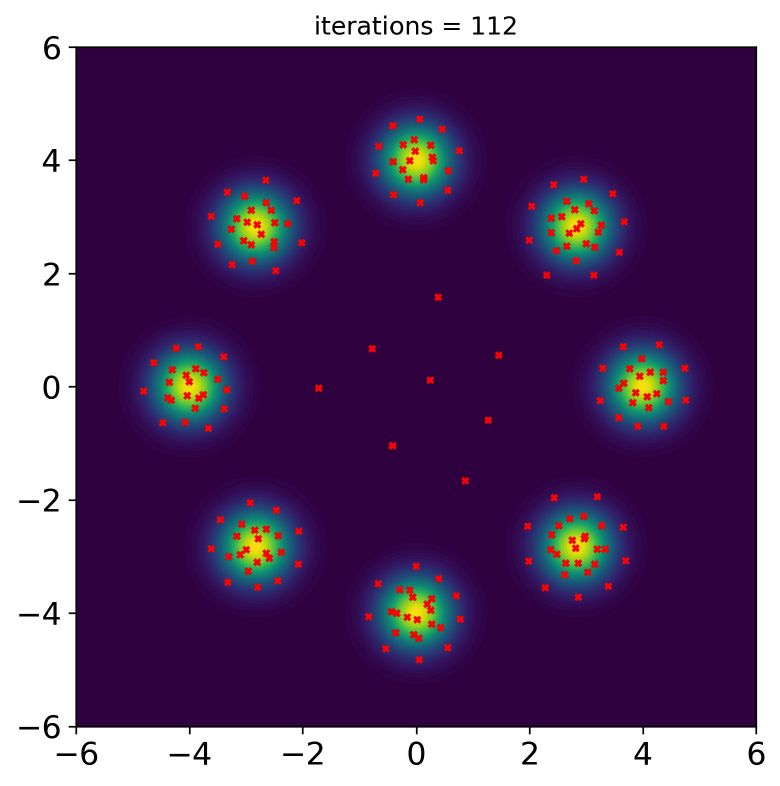}
		\includegraphics[width=0.24\linewidth]{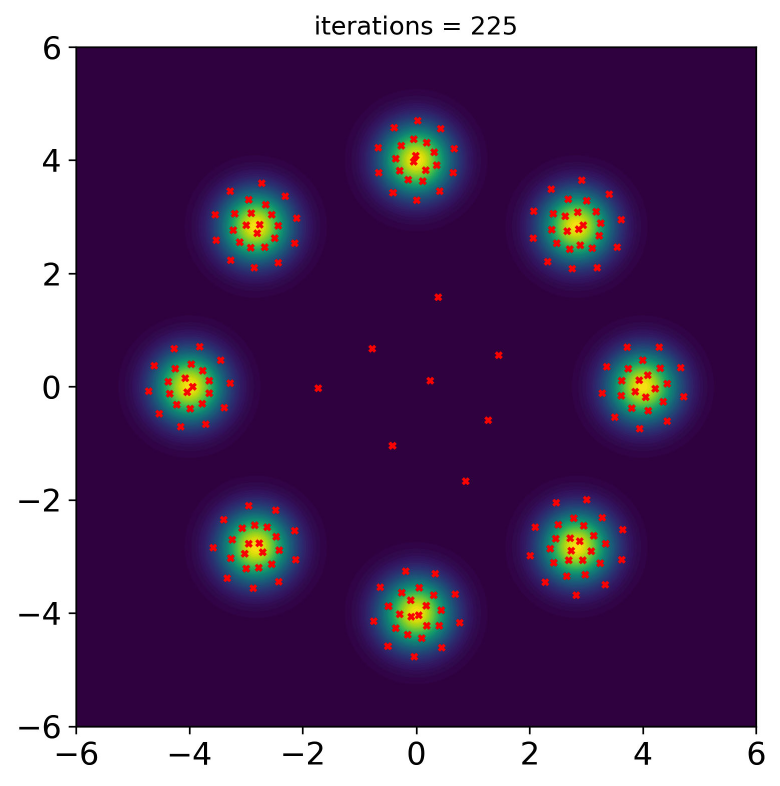}
	\end{subfigure}	
	\begin{subfigure}{\linewidth}
		\makebox[0pt][r]{\makebox[30pt]{\raisebox{40pt}{\rotatebox[origin=c]{90}{$a=5$}}}}%
		\includegraphics[width=.24\linewidth]{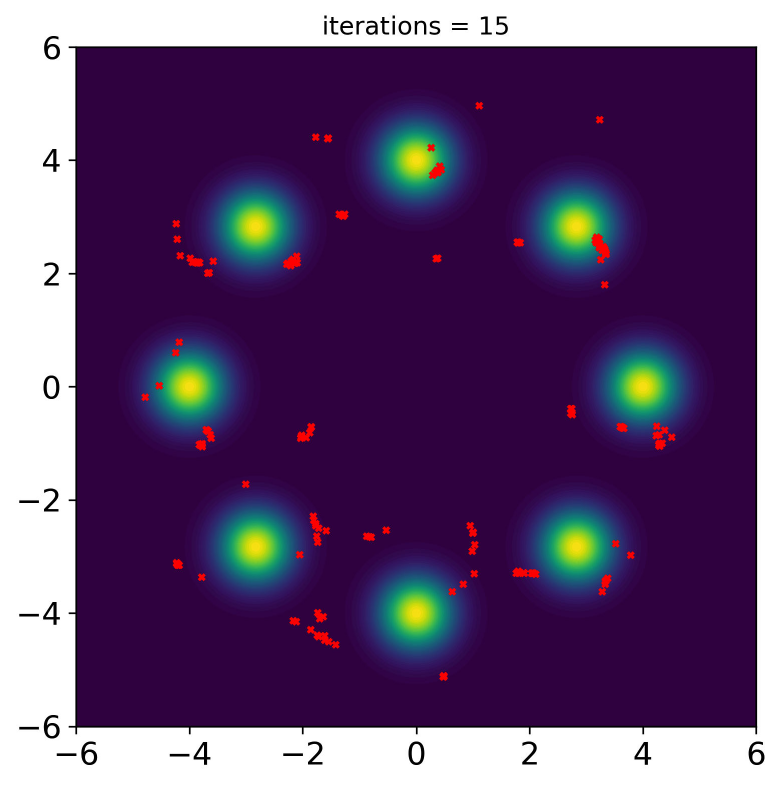}
		\includegraphics[width=.24\linewidth]{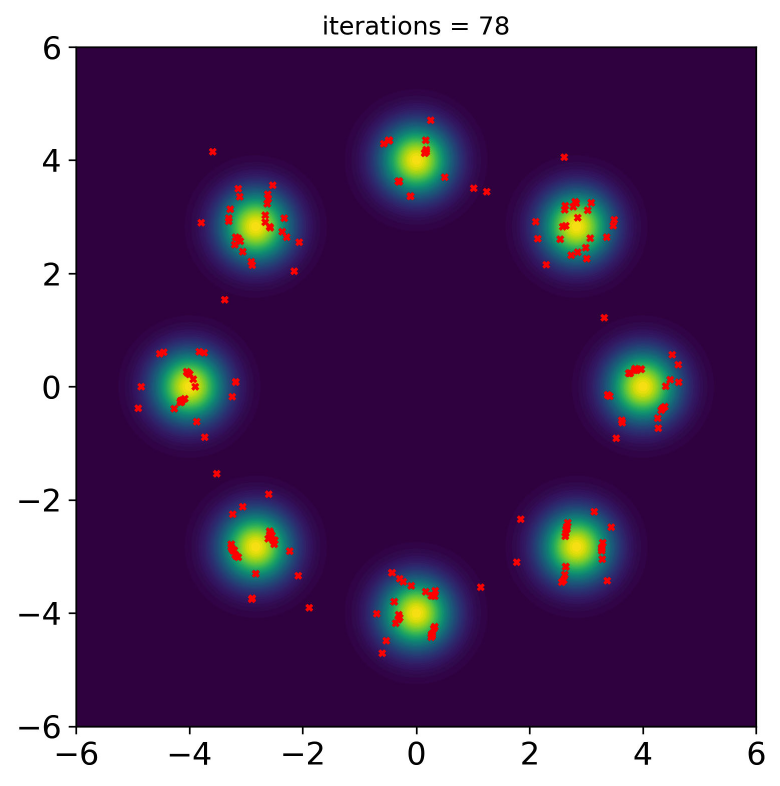}
		\includegraphics[width=0.24\linewidth]{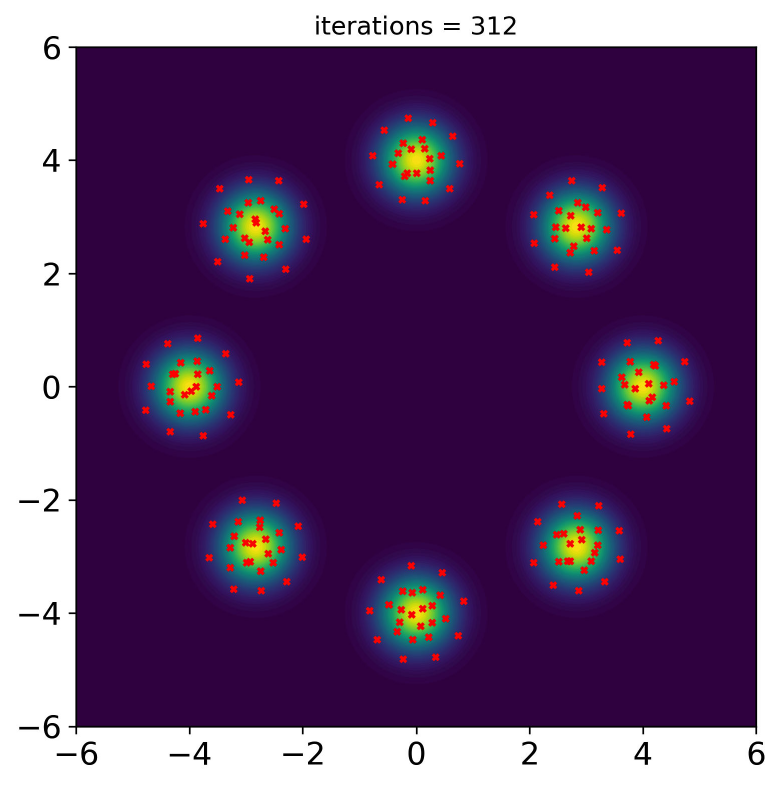}
		\includegraphics[width=0.24\linewidth]{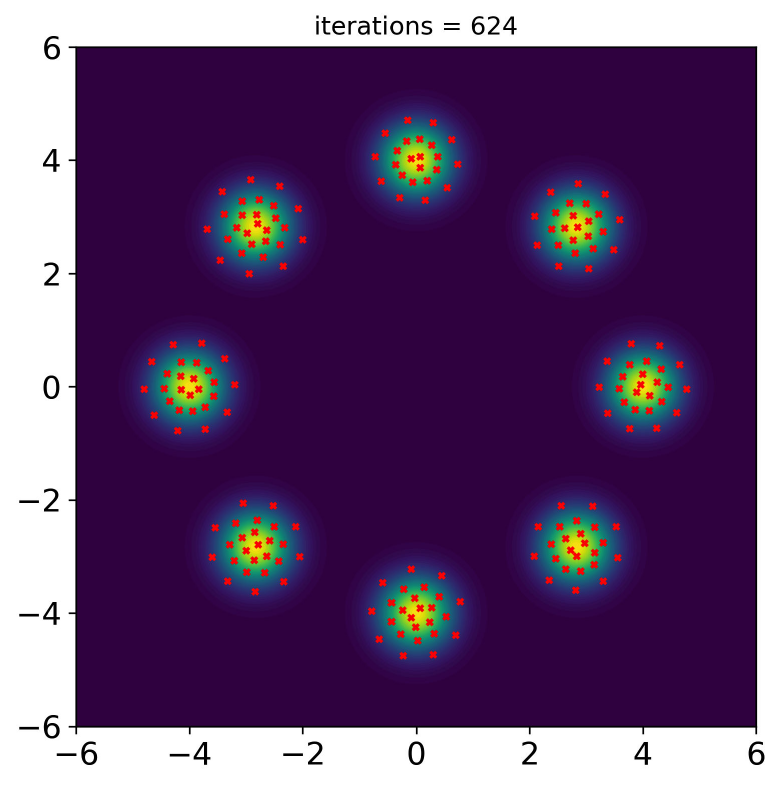}
	\end{subfigure}
		\begin{subfigure}{\linewidth}
		\makebox[0pt][r]{\makebox[30pt]{\raisebox{40pt}{\rotatebox[origin=c]{90}{$a=6$}}}}%
		\includegraphics[width=.24\linewidth]{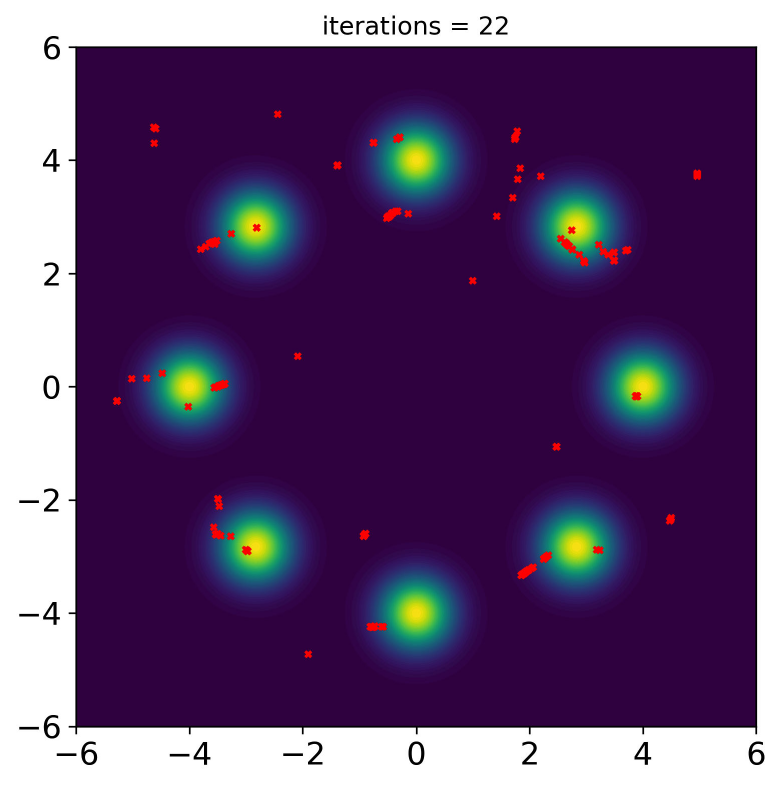}
		\includegraphics[width=.24\linewidth]{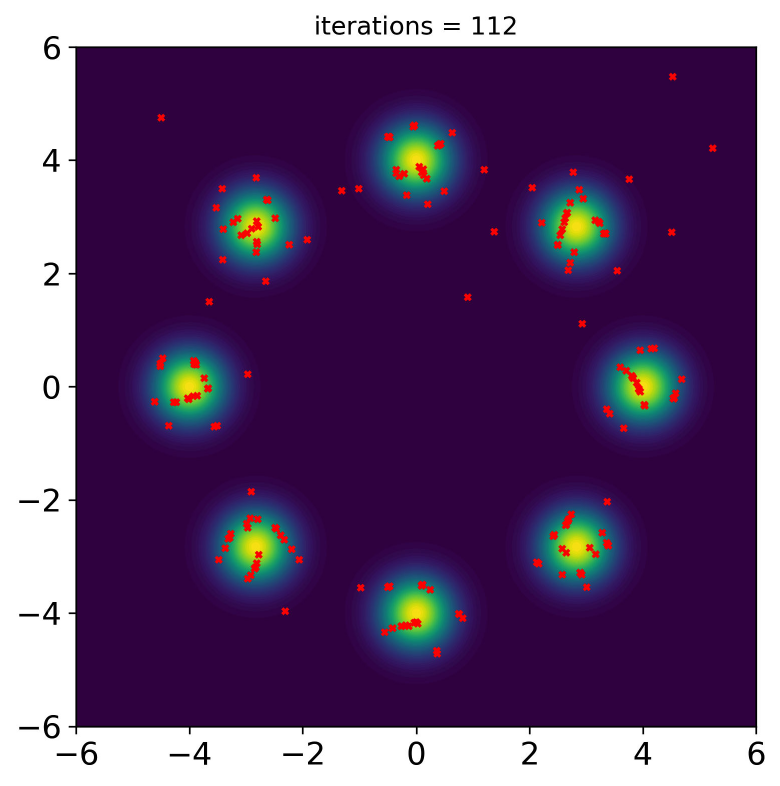}
		\includegraphics[width=0.24\linewidth]{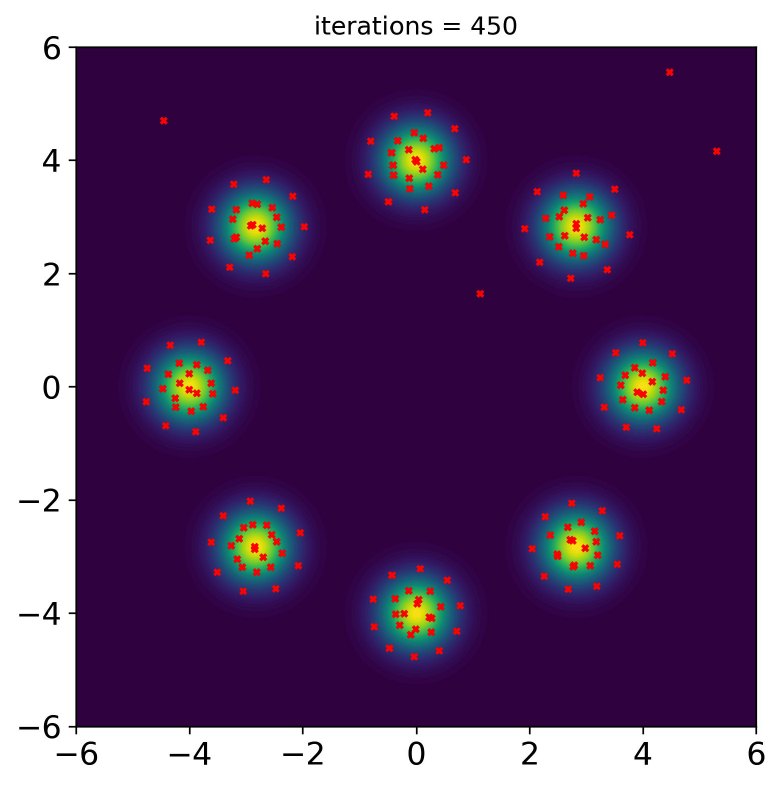}
		\includegraphics[width=0.24\linewidth]{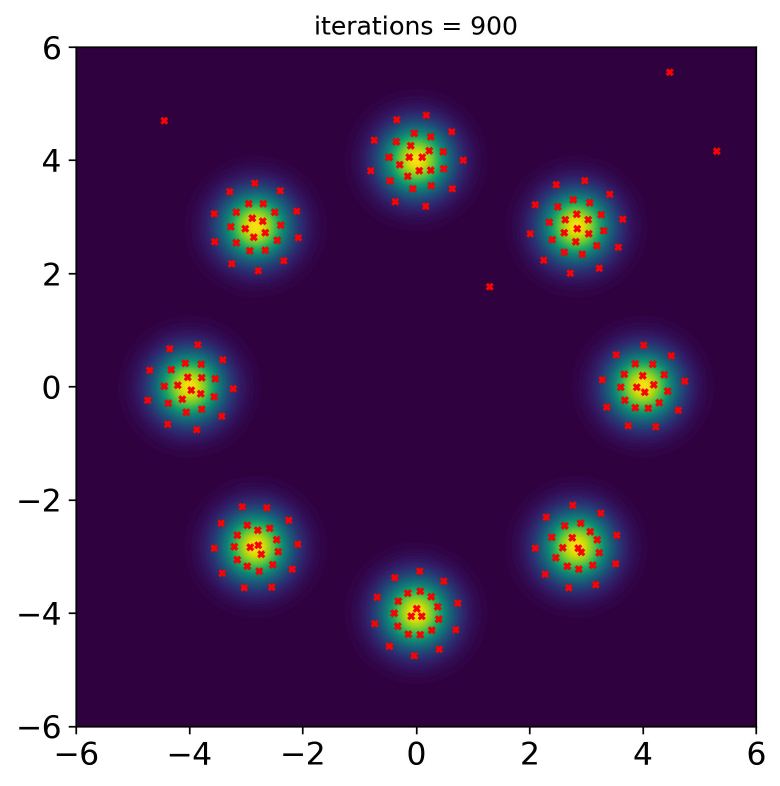}
	\end{subfigure}	
	\caption{Particle trajectories of different choice of $a$. In the first row, $a \approx 4$, the median of the pairwise distance of the initial particles as in the toy example; in the remaining rows, $a = 2, 3$, and $5$, respectively. For each sub-figure of the same column, the value of $h_n$ is similar.}\label{fig:tune2}
\end{figure}

The first row of Figure \ref{fig:tune1} (from left to right) shows the distribution of particles at iterations $\text{iter} = 10, 50, 200, 400$ using the same parameters as in Section \ref{sub:toy}. The remaining rows in Figure \ref{fig:tune1} illustrate the effects of increasing or decreasing the value of $c$, while keeping other parameters fixed. For each sub-figure, iterations are chosen such that the bandwidth $h_n$ is numerically similar column-wise. The title of each sub-figure indicates the corresponding iterations. Thus, the first row of Figure \ref{fig:tune1} serves as a reference, while the other rows demonstrate the impact of varying $c$. Similarly, Figure \ref{fig:tune2} explores the role of the parameter $a$.

Two key observations emerge. First, in both Figures \ref{fig:tune1} and \ref{fig:tune2}, particles in high-density regions exhibit similar alignment for comparable bandwidths $h_n$. Second, particles initially located in low-density areas tend to remain as outliers even after 200 iterations, reflecting the trade-off between exploration and exploitation.

For the case $c=0.6$ and $a=5$, the overall performance aligns with the proposed parameter settings in the toy example (Section \ref{sub:toy}). However, achieving the same bandwidth requires more iterations, as $h_n$ decreases more slowly than in Section \ref{sub:toy}. The exception is the case $a=2$, where the initial bandwidth is too large, causing some particles to be pushed away and fail to converge before the bandwidth decreases to the exploitation stage. This issue is resolved by using the median trick, as shown in the first row of Figure \ref{fig:tune2}.

\begin{figure}[!htb]
	\centering
	\begin{subfigure}{\linewidth}
		\makebox[0pt][r]{\makebox[30pt]{\raisebox{40pt}{\rotatebox[origin=c]{90}{$\tau=2$}}}}%
		\includegraphics[width=0.24\linewidth]{base_11.pdf}
		\includegraphics[width=0.24\linewidth]{base_15.pdf}
		\includegraphics[width=0.24\linewidth]{base_30.pdf}
		\includegraphics[width=0.24\linewidth]{base_50.pdf}
	\end{subfigure}	
	\begin{subfigure}{\linewidth}
		\makebox[0pt][r]{\makebox[30pt]{\raisebox{40pt}{\rotatebox[origin=c]{90}{$\tau=0.1$}}}}%
		\includegraphics[width=.24\linewidth]{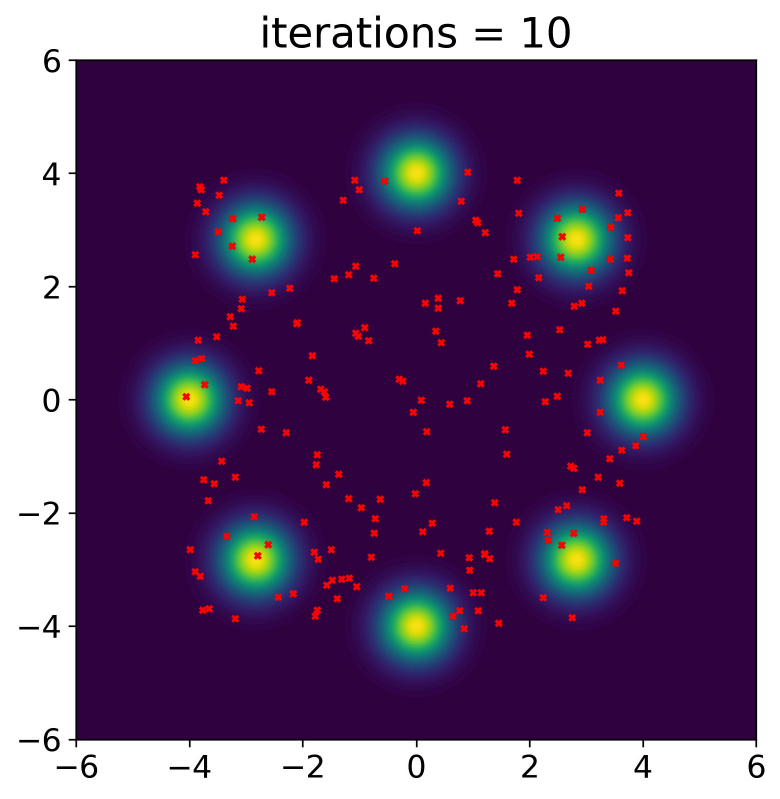}
		\includegraphics[width=.24\linewidth]{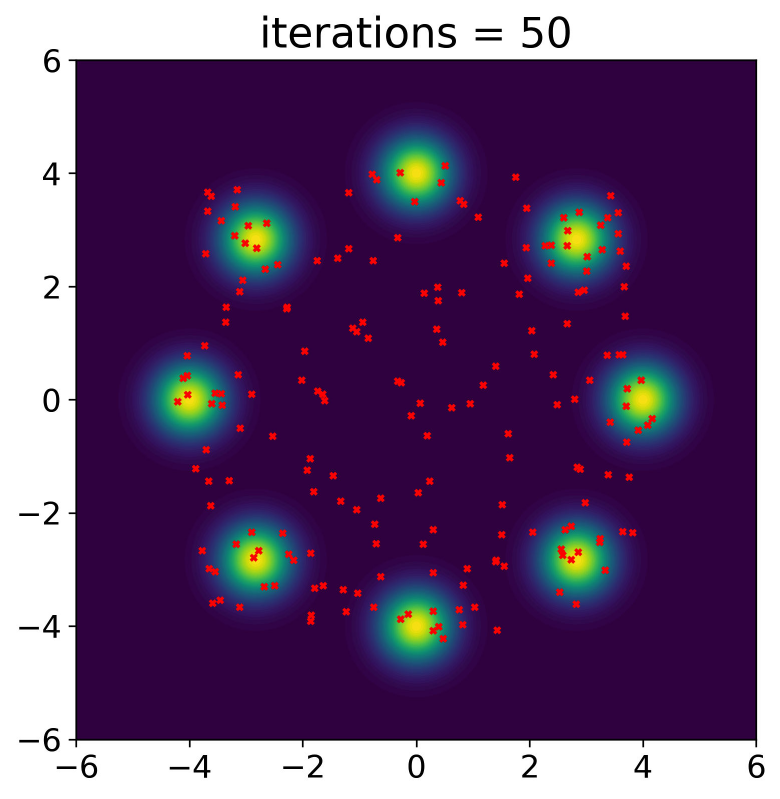}
		\includegraphics[width=0.24\linewidth]{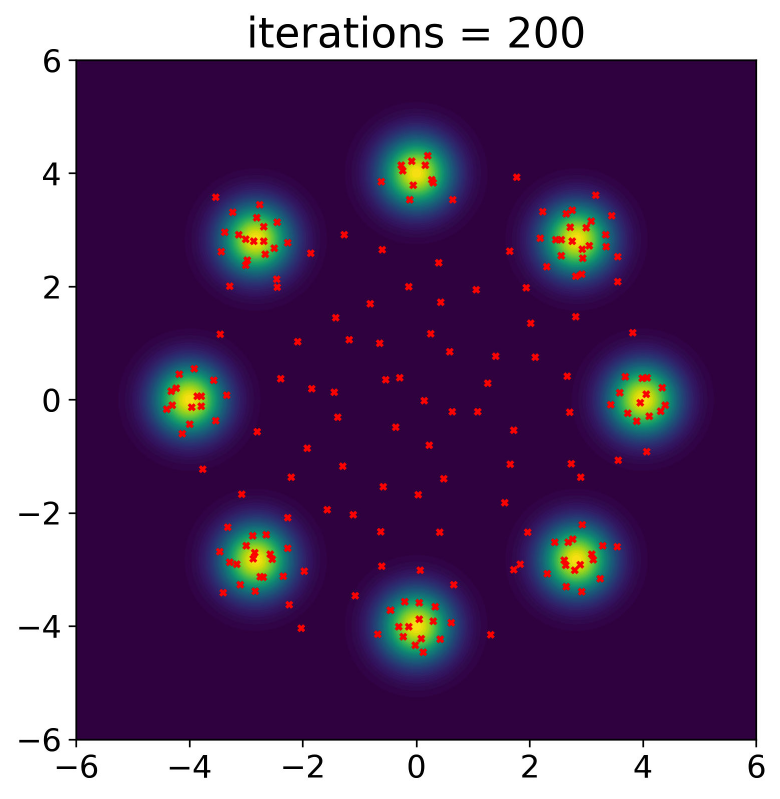}
		\includegraphics[width=0.24\linewidth]{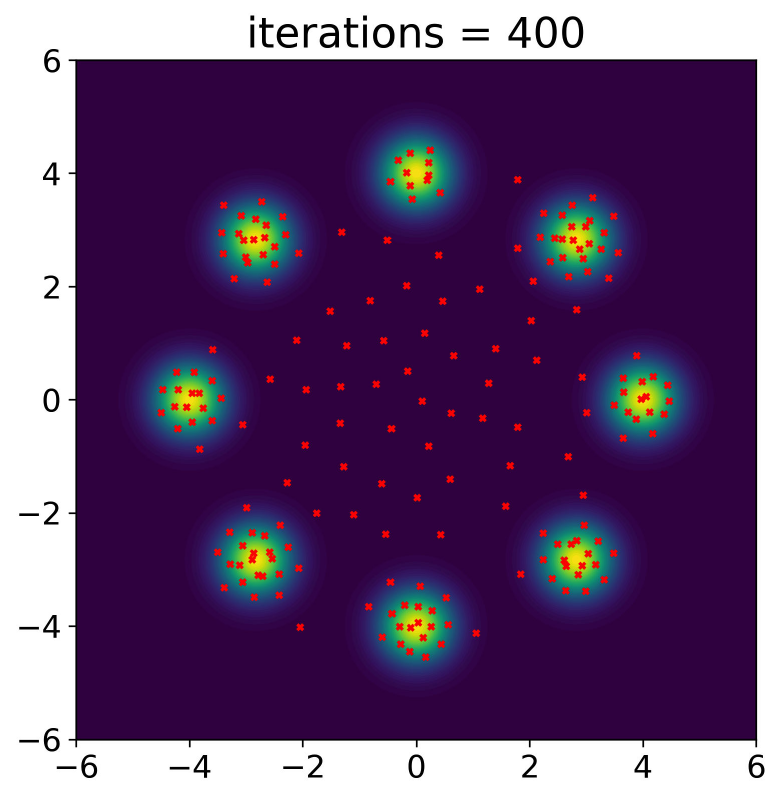}
	\end{subfigure}
	\begin{subfigure}{\linewidth}
		\makebox[0pt][r]{\makebox[30pt]{\raisebox{40pt}{\rotatebox[origin=c]{90}{$\tau=1$}}}}%
		\includegraphics[width=.24\linewidth]{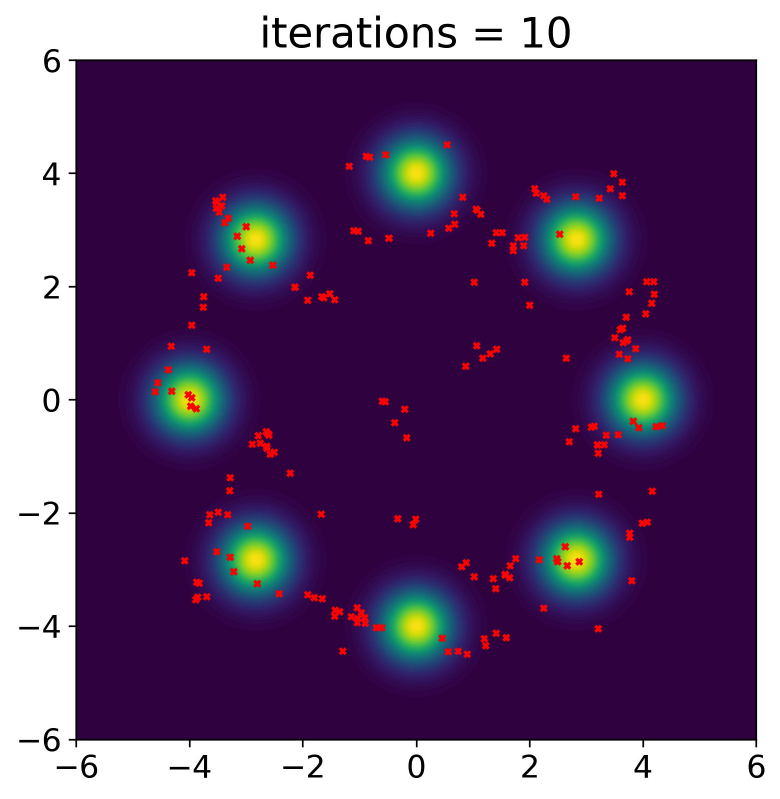}
		\includegraphics[width=.24\linewidth]{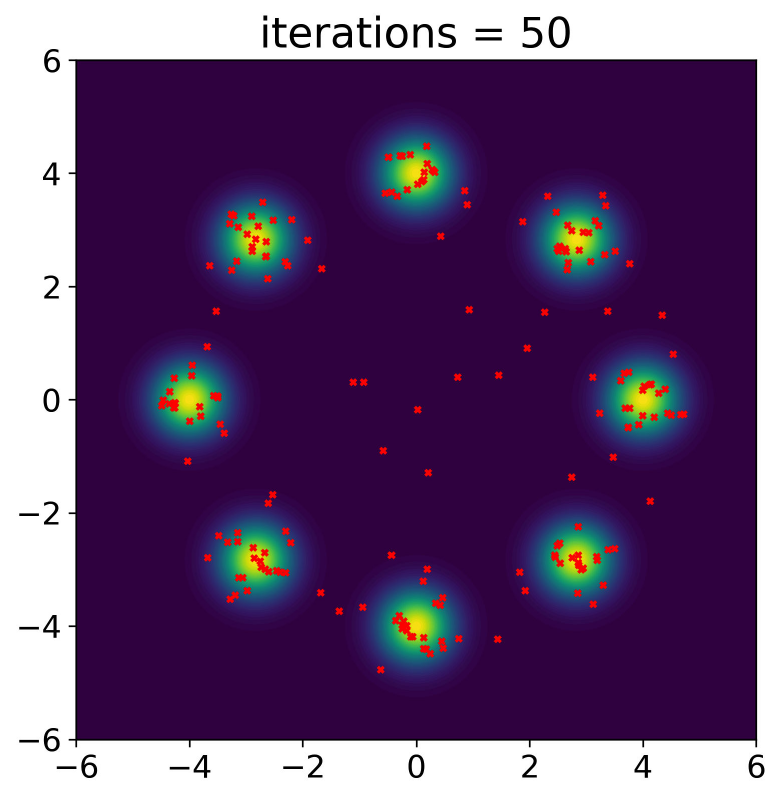}
		\includegraphics[width=0.24\linewidth]{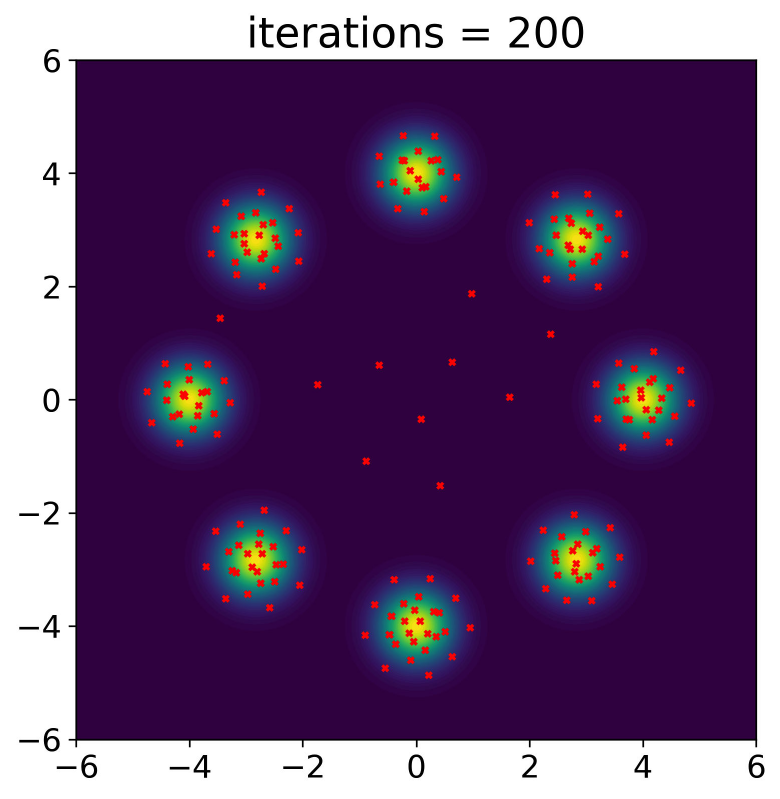}
		\includegraphics[width=0.24\linewidth]{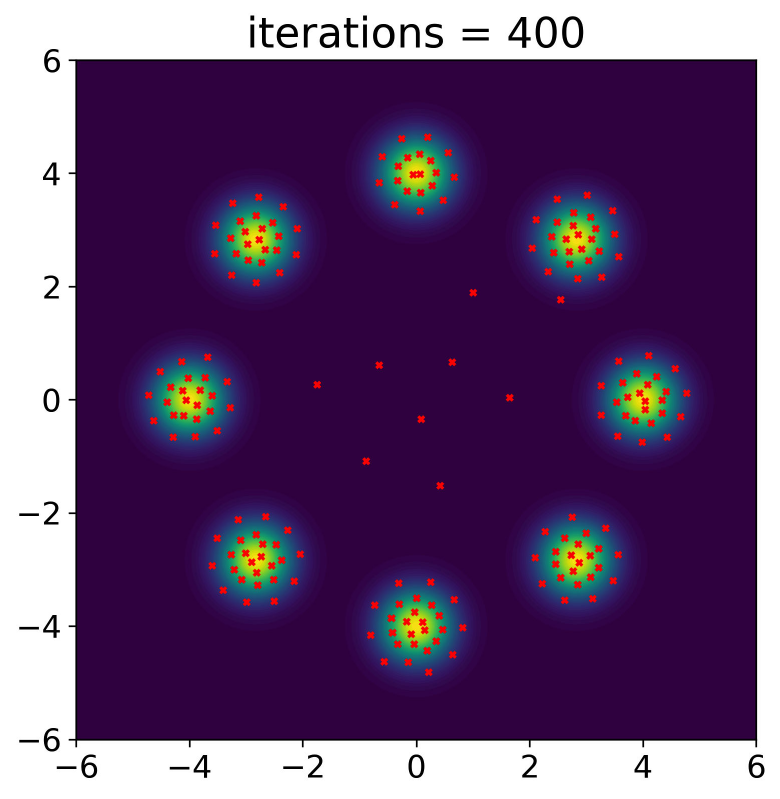}
	\end{subfigure}	
	\begin{subfigure}{\linewidth}
		\makebox[0pt][r]{\makebox[30pt]{\raisebox{40pt}{\rotatebox[origin=c]{90}{$\tau=4$}}}}%
		\includegraphics[width=.24\linewidth]{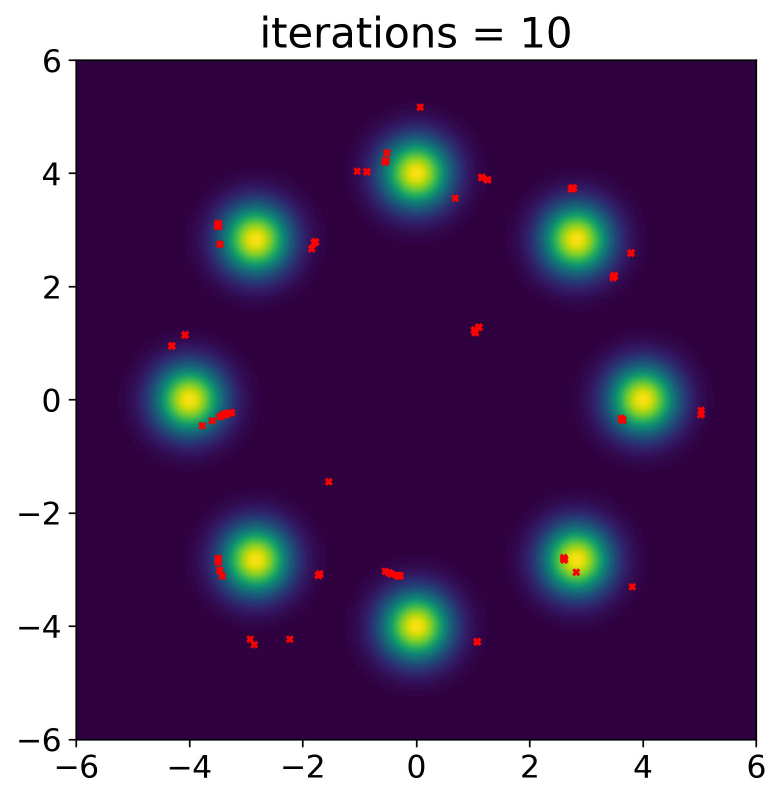}
		\includegraphics[width=.24\linewidth]{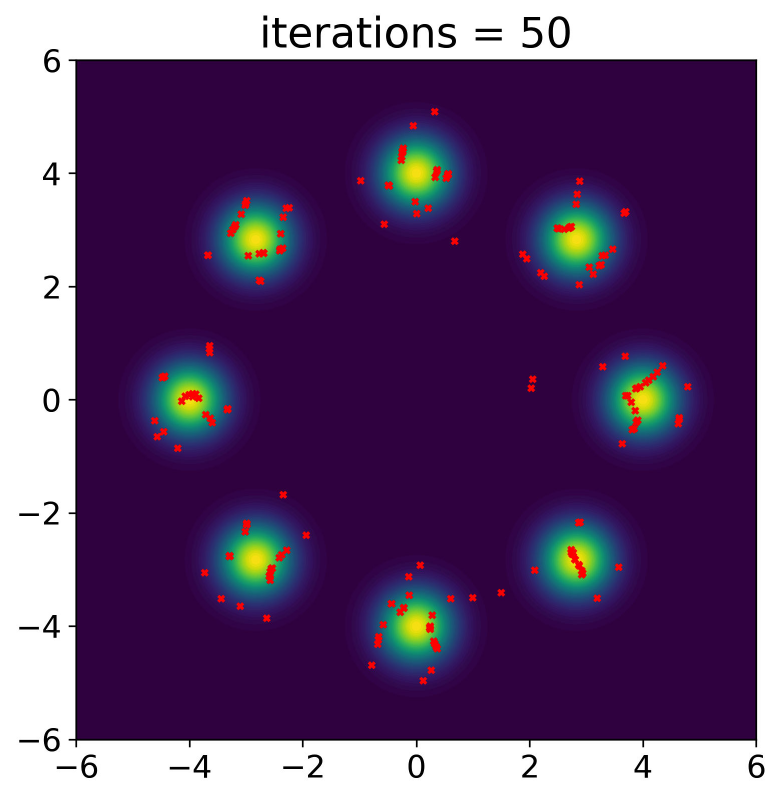}
		\includegraphics[width=0.24\linewidth]{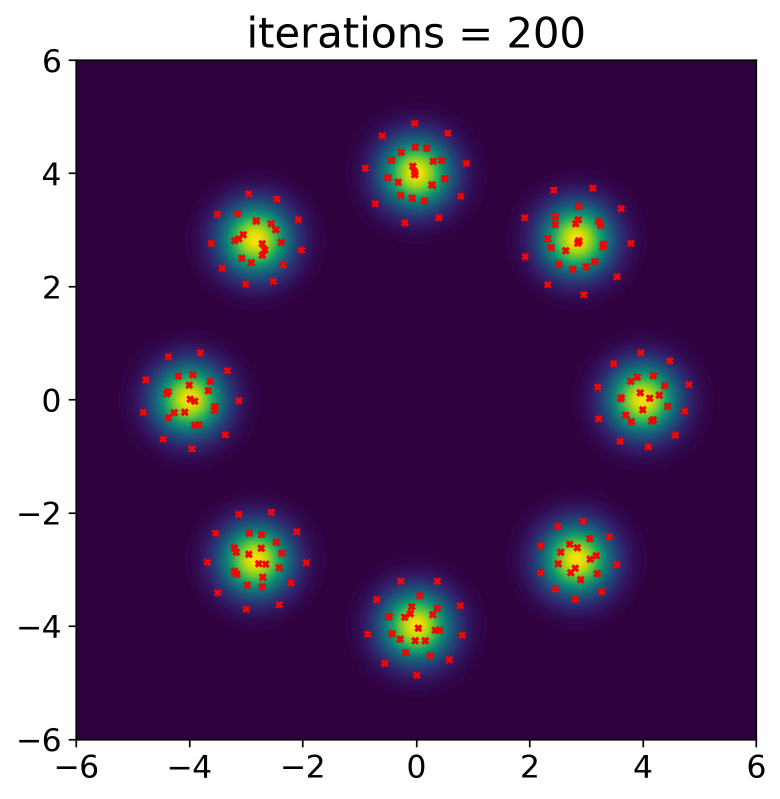}
		\includegraphics[width=0.24\linewidth]{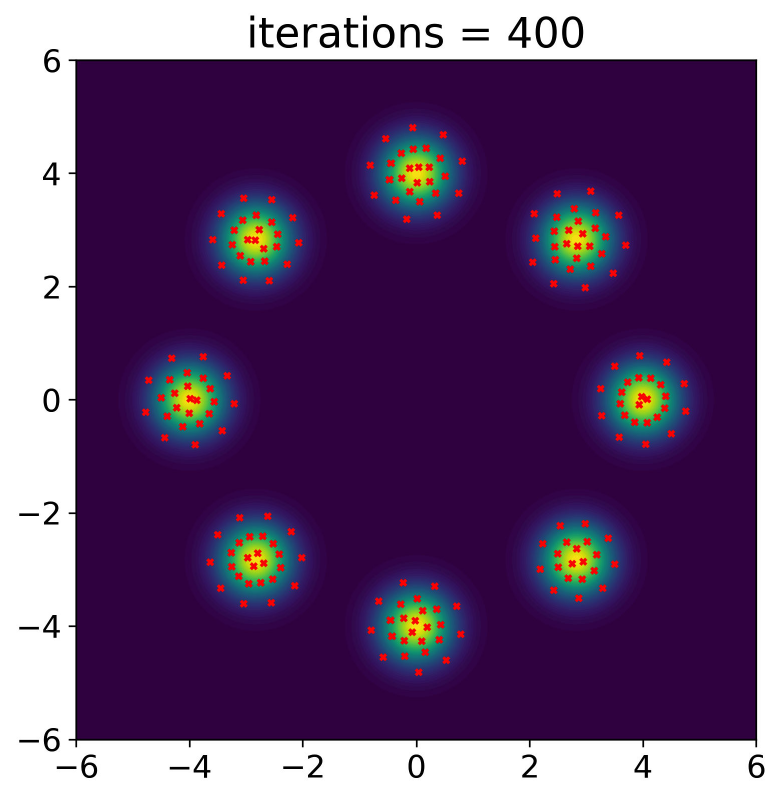}
	\end{subfigure}
	\begin{subfigure}{\linewidth}
		\makebox[0pt][r]{\makebox[30pt]{\raisebox{40pt}{\rotatebox[origin=c]{90}{$\tau=\infty$}}}}%
		\includegraphics[width=.24\linewidth]{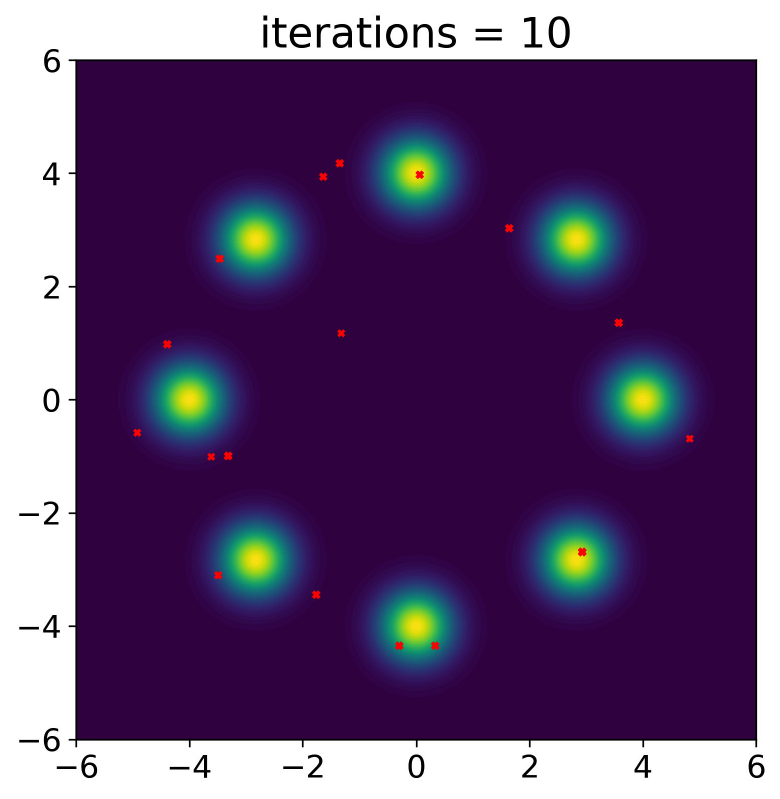}
		\includegraphics[width=.24\linewidth]{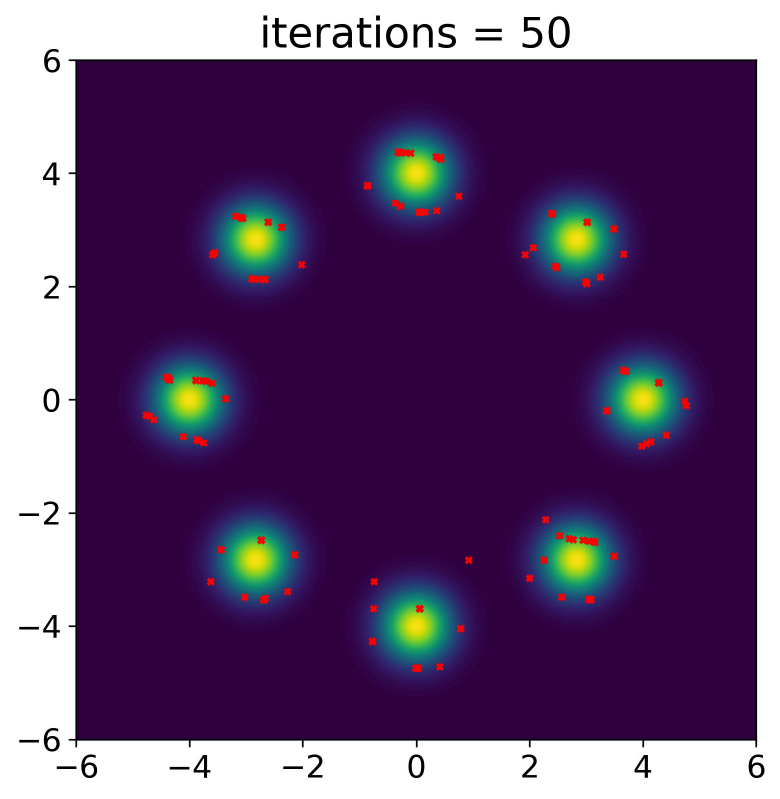}
		\includegraphics[width=0.24\linewidth]{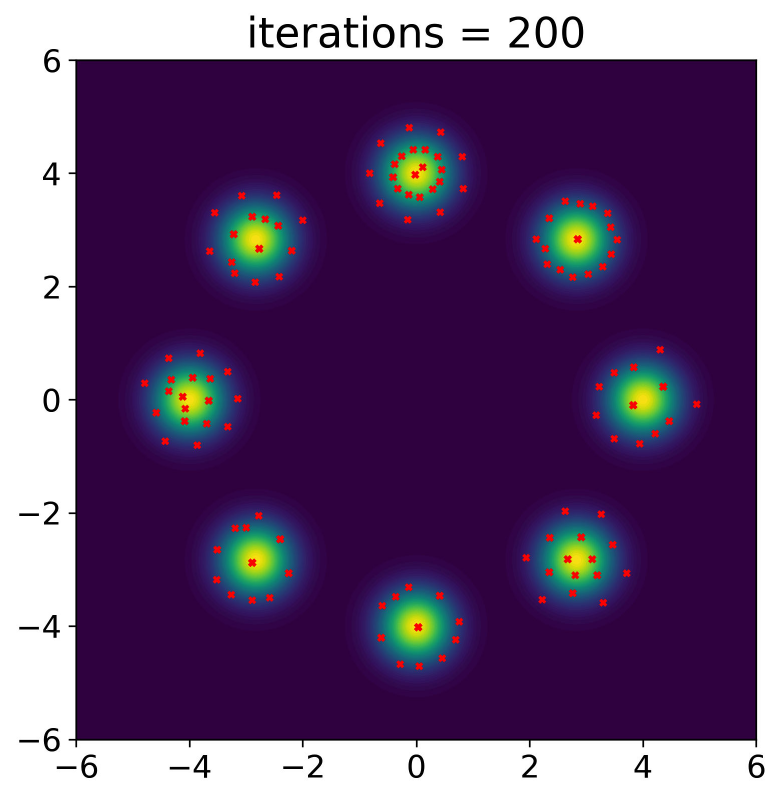}
		\includegraphics[width=0.24\linewidth]{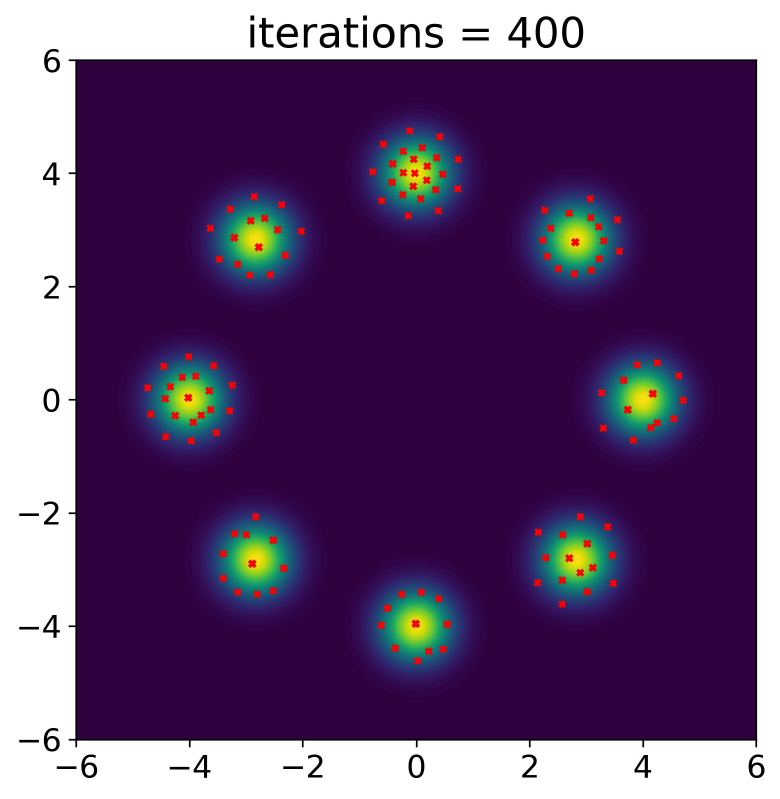}
	\end{subfigure}	
	\caption{Particle trajectories of different choice of $\tau$. In the first row, $\tau = 2$ as in the toy example; in the remaining rows, $\tau = 0.1, 1, 4 $ and $\infty$, respectively. For each sub-figure of the same column, the value of $h_n$ is similar.}\label{fig:tune3}
\end{figure}

Next, we examine the choice of $\tau$. As shown in Figure \ref{fig:tune3}, when $\tau$ is too small, particles cannot converge quickly enough before the bandwidth becomes small enough for the exploitation stage. A higher $\tau$ yields results similar to the proposed settings in Section \ref{sub:toy}. However, when $\tau=\infty$ (equivalent to solving the problem explicitly), some particles collapse into a single point. This motivates the use of the implicit scheme over the explicit scheme.

In conclusion, the convergence speed primarily depends on the bandwidth $h_n$. Selecting $a$, $c$, and $\tau$ involves balancing exploration (moving particles to high-density regions) and exploitation (local refinement). If outliers are prevalent, a slightly smaller $c$ or a larger initial bandwidth $h_0=a$ is recommended. To accelerate convergence for most particles, increasing $c$ or reducing the initial bandwidth $h_0=a$ can be effective.

\end{document}